%% file: main.tex
\PassOptionsToPackage{table,xcdraw,dvipsnames}{xcolor}
\documentclass{article}

\usepackage{microtype}
\usepackage{graphicx}
\usepackage{subfigure}
\usepackage{booktabs} 
\usepackage{multirow}
\usepackage{hyperref}

\usepackage[preprint]{icml2026}

\usepackage{amsmath}
\usepackage{amssymb}
\usepackage{mathtools}
\usepackage{amsthm}

\usepackage[capitalize,noabbrev]{cleveref}

\definecolor{graybg}{rgb}{0.95, 0.95, 0.95}

\theoremstyle{plain}

\theoremstyle{definition}

\theoremstyle{remark}

\usepackage[textsize=tiny]{todonotes}

\usepackage{xspace}
\newcommand{\llmname}[1]{{\fontfamily{pcr}\selectfont {#1}}\xspace}

\definecolor{ForestGreen}{RGB}{34,139,34}
\definecolor{myyellow}{RGB}{181, 181, 27}

\usepackage{makecell}
\usepackage{enumerate}
\usepackage{enumitem}
\usepackage{pifont}
\usepackage{fontawesome5}  

\definecolor{mygrey}{gray}{0.4}

\usepackage[ruled,algo2e,vlined]{algorithm2e}
\renewcommand{\algorithmicrequire}{\textbf{Input:}}  
\renewcommand{\algorithmicensure}{\textbf{Output:}} 
\SetKwInOut{Input}{Input}\SetKwInOut{Output}{Output}
\SetKwComment{Comment}{$\triangleright$\ }{}

\usepackage{listings}

\usepackage[most]{tcolorbox}
\tcbuselibrary{breakable,skins}
\tcbuselibrary{breakable, listings}

\usepackage{float}  

\definecolor{DeepGreen}{HTML}{5F8643}      
\definecolor{LightGreen}{HTML}{EEF5EA}     
\definecolor{MidGreen}{HTML}{DCEBD2}       
\definecolor{CaseGrey}{HTML}{B0B0B0}       

\definecolor{FactGreen}{HTML}{C8E6C9}      
\definecolor{VPCRed}{HTML}{FFCDD2}         
\definecolor{VTCYellow}{HTML}{FFF9C4}      
\definecolor{PTCGrey}{HTML}{E0E0E0}        

\definecolor{EarthTerracotta}{HTML}{c7522a}   
\definecolor{EarthSand}{HTML}{e5c185}         
\definecolor{EarthTeal}{HTML}{008585}         

\usepackage{soul}
\sethlcolor{FactGreen}
\newcommand{\hlfact}[1]{\sethlcolor{FactGreen}\hl{#1}}       
\newcommand{\hlvpc}[1]{\sethlcolor{VPCRed}\hl{#1}}           
\newcommand{\hlvtc}[1]{\sethlcolor{VTCYellow}\hl{#1}}        
\newcommand{\hlptc}[1]{\sethlcolor{PTCGrey}\hl{#1}}          

\newcommand{\stepbox}[2]{\colorbox{#1!20}{\textbf{[Step #2]}}}

\newtcolorbox{dashanalysis}[1][]{%
  breakable,
  colback=white,
  colframe=DeepGreen,
  boxrule=1.1pt,
  arc=2mm,
  left=2mm,right=2mm,top=2mm,bottom=2mm,
  enhanced,
  borderline={1.1pt}{0pt}{DeepGreen,dashed},
  #1
}

\usepackage{array}      
\usepackage{colortbl}   

\newcommand{\good}[1]{\textcolor{black}{#1}}
\newcommand{\bad}[1]{\textcolor{black}{#1}}

\definecolor{rowbase}{RGB}{245,245,245}  
\definecolor{rowhi}{RGB}{242,246,255}    
\definecolor{sagegreen}{HTML}{e5c185}

\newcommand{\dpos}[1]{\textsuperscript{\scriptsize\good{#1}}}
\newcommand{\dneg}[1]{\textsuperscript{\scriptsize\bad{#1}}}

\newcolumntype{W}{!{\color{black!40}\vrule width 0.55pt}} 
\newcolumntype{V}{!{\color{black!40}\vrule width 0.35pt}} 

\icmltitlerunning{Diagnosing Knowledge Conflict in Multimodal Long-Chain Reasoning}

\begin{document}

\twocolumn[
\icmltitle{Diagnosing Knowledge Conflict in Multimodal Long-Chain Reasoning}



\icmlsetsymbol{equal}{*}

\begin{icmlauthorlist}
\icmlauthor{Jing Tang}{hust,equal}
\icmlauthor{Kun Wang}{ntu,equal}
\icmlauthor{Haolang Lu}{bupt,equal} 
\icmlauthor{Hongjin Chen}{bupt}
\icmlauthor{KaiTao Chen}{bupt}
\icmlauthor{Zhongxiang Sun}{ruc}
\icmlauthor{Qiankun Li}{ntu}
\icmlauthor{Lingjuan Lyu}{sony}
\icmlauthor{Guoshun Nan}{bupt}
\icmlauthor{Zhigang Zeng}{hust}
\end{icmlauthorlist}

\vspace{-0.2cm}
\begin{center}
{\small 
\texttt{jingtang@hust.edu.cn} \quad
\texttt{wang.kun@ntu.edu.sg} \quad
\texttt{luhaolang@bupt.edu.cn}

}
\end{center}
\vspace{-0.2cm}

\icmlaffiliation{hust}{Huazhong University of Science and Technology}
\icmlaffiliation{ntu}{Nanyang Technological University}
\icmlaffiliation{bupt}{Beijing University of Posts and Telecommunications}
\icmlaffiliation{ruc}{Renmin University of China}
\icmlaffiliation{sony}{Sony AI, Zurich, Switzerland}

\icmlcorrespondingauthor{Guoshun Nan}{nanguoshun@gmail.com}
\icmlcorrespondingauthor{Zhigang Zeng}{zgzeng@hust.edu.cn}

\icmlkeywords{Machine Learning, ICML}
\vskip 0.3in

]




\printAffiliationsAndNotice{} 

\begin{abstract}
\input{secs/abstract}
\end{abstract}

\input{secs/1.introduction}
\input{secs/2.related_work}
\input{secs/3.knowledge_conflict}

\input{secs/4.conflict_detect}
\input{secs/5.intervention}

\input{secs/6.conclusion}

\input{secs/impact_statement}

\bibliography{ref}
\bibliographystyle{icml2026}

\clearpage
\input{secs/appendix}

\end{document}

%% file: secs/abstract.tex
Multimodal large language models in long chain-of-thought reasoning often fail when different knowledge sources provide conflicting signals. We formalize these failures under a unified notion of knowledge conflict, distinguishing input-level objective conflict from process-level effective conflict. Through probing internal representations, we reveal that: (\textbf{I}) \textbf{Linear Separability}: different conflict types are explicitly encoded as linearly separable features rather than entangled; (\textbf{II}) \textbf{Depth Localization}: conflict signals concentrate in mid-to-late layers, indicating a distinct processing stage for conflict encoding; (\textbf{III}) \textbf{Hierarchical Consistency}: aggregating noisy token-level signals along trajectories robustly recovers input-level conflict types; and (\textbf{IV}) \textbf{Directional Asymmetry}: reinforcing the model’s implicit source preference under conflict is far easier than enforcing the opposite source. Our findings provide a mechanism-level view of multimodal reasoning under knowledge conflict and enable principled diagnosis and control of long-CoT failures. Code is available at \href{https://anonymous.4open.science/r/Diagnosing-Knowledge-Conflict-C78D/}{anonymous link}.
    

%% file: secs/1.introduction.tex
\vspace{-1.5em}
\section{Introduction}
\label{sec:intro}
\vspace{-0.5em}

Multimodal large language models (MLLMs)~\cite{jin2025efficient,caffagni2024revolution,zhang2024mm} have made substantial progress in visual understanding~\cite{tong2024cambrian,ghatkesar2025looking,ma2025deepperception}, textual reasoning~\cite{wang2024mmlu,du2025virgo,mirzadeh2025gsmsymbolic}, and cross-modal alignment~\cite{yu2024rlhf,yan2025task,yu2025rlaif}, enabling complex perception–reasoning–decision workflows. A defining capability is long-form reasoning: beyond producing answers, these models can generate extended chains-of-thought (CoT)~\cite{wang2025multimodal,yue2025mmmu} that support challenging multi-step tasks. However, recent work increasingly documents failures under mutually contradictory evidence or constraints: models may ignore explicit instructions~\cite{wang2025comprehensive,zhao2025jailbreaking}, privilege the wrong evidential source~\cite{guan2024hallusionbench,liu-etal-2025-insight}, or yield plausible yet goal-inconsistent conclusions~\cite{fanous2025syceval}. These observations suggest that a \textbf{key bottleneck in multimodal reasoning is not always missing information, but reliable decision-making under conflicting signals}.

Building on these observations, prior work~\cite{zhang2024multitrust,lu2024wildvision} has characterized abnormal behavior under conflicting signals from several largely independent angles. In retrieval-augmented generation, a central question is whether models remain faithful to retrieved evidence or drift toward parametric priors~\cite{wu2024clasheval}. In vision settings with counterfactual or commonsense-violating inputs, MLLMs are often found to underweight visual evidence and default to “reasonable” answers that match world knowledge~\cite{tong2024eyes,liu2025unveiling}. In high-stakes domains, studies further report over-accommodation to user assertions, which can pull predictions away from the underlying evidence~\cite{sharma2024towards}. Although these lines of work differ in tasks, datasets, and evaluation criteria, their failure modes are strikingly similar: when information sources disagree, models do not reliably follow the appropriate basis for a decision, and instead exhibit unstable, hard-to-control trade-offs across sources. 


In this paper, we take a unified view that these phenomena arise from \emph{knowledge conflict} in multimodal reasoning. When generating tokens, MLLMs jointly rely on multiple knowledge sources, including visual evidence, textual instructions and contextual constraints, and parametric priors stored in the model weights~\cite{han2025learning,liu2024survey,karamcheti2024prismatic}. When these sources provide inconsistent signals for the same goal, the model must resolve which source to follow. Importantly, the resulting failures are not fabrications from missing knowledge, but incorrect source selection under conflict: \textbf{the model may have access to competing plausible cues yet follow the wrong basis}. Accordingly, our focus is not the act of answer generation itself, but whether conflict-induced failures can be \emph{localized, measured, and mechanistically tested}.

Multimodal long-CoT reasoning~\cite{ni2025visualo} makes this problem sharper by unfolding decisions over many steps, with the internal reasoning state evolving over time. Under this setting, knowledge conflict can be triggered at any point and modality along the trajectory rather than only at the final answer. Once a step commits to the wrong basis, subsequent steps may continue from that premise in a locally coherent manner, eventually producing a globally incorrect conclusion~\cite{zhang2024how}. More challenging, such deviations are often masked by fluent rationales~\cite{turpin2023language}, making it difficult to infer when the conflict emerged, what triggered it, and how it propagated from the final output alone. Understanding and correcting failures in long-CoT therefore requires step-level tools that can expose the underlying conflict dynamics.

In this work, 
\ding{68} We diagnose knowledge conflict dynamics on \textbf{7,500+} long-CoT trajectories from an \emph{objective conflict} benchmark, where \emph{effective conflicts} are activated in \textbf{78–90\%} of samples.
\ding{68} Through layer-wise analysis of three models, we identify a depth-dependent \emph{conflict encoding stage}. Using streaming probes to detect token-level conflict states, we find they exhibit high linear separability (\textbf{93.2$\sim$98.8\% AUC}, \textbf{76.9$\sim$97.8\% Recall@0.1}), revealing them as explicit, decodable features.
\ding{68} We employ three \emph{pluggable} methods for intervention. These methods can either steer model outputs toward selected directions, reducing conflict frequency by \textbf{up to 80\%}, or suppress high-confidence errors by up to \textbf{55\%}.

%% file: secs/2.related_work.tex
\section{Related Work}
\vspace{-0.3em}
\paragraph{Knowledge Conflict.}
Research on knowledge conflicts has identified three primary sources: conflicts between internal priors and visual information~\citep{liu-etal-2025-insight, du.2025mmkebench} or textual inputs~\citep{zhang-etal-2025-faithfulrag, su2024conflictbank}, and conflicts between visual and textual modalities~\citep{deng2025words}.
Building on these findings, significant efforts have been made to mitigate such conflicts through advanced strategies~\citep{xie2024adaptive, guo2023unkvqa}, including knowledge editing~\citep{tan2024massive, zhang2025conflictaware, cheng2024editing, chen2025knowledge} and retrieval augmentation~\citep{huo-etal-2025-micro, zhang2025knowpo, li2025juice}.
These approaches have demonstrated potential in enhancing model faithfulness and reliability~\citep{huang2025to, an2025boosting, NEURIPS2024_08a9e28c, zhang2024trustworthy, lu2025kedkg}.
Although the above evidence suggests that conflicts are coupled and multi-source, existing solutions remain fragmented across modalities and fail to model conflicts holistically, thereby limiting their applicability in complex settings.

\vspace{-2.0em}
\paragraph{Probe Detection.} 
Investigating internal states via probe detection is a developing field, yet the history of probing in LLMs~\citep{kahana2025deep} provides clear precedents. Notably, the evolution of probe detection primarily centers on hallucination and faithfulness~\citep{feng2025monitoring, yi2025probe}. Core techniques, such as linear probe generators~\citep{kahana2025deep} and propositional probes~\citep{feng2025monitoring}, have inspired analogous approaches in watermark identification~\citep{liu2025can}, reward maximization~\citep{li2024qprobe}, and combinatorial optimization~\citep{zhang2025probing}. However, these approaches predominantly focus on single-modal issues or specific downstream tasks, leaving the detection and localization of multimodal knowledge conflicts largely unexplored. Inspired by this, we introduce a specialized probe detection framework to identify the three sources of knowledge conflicts in MLLMs.  


%% file: secs/3.knowledge_conflict.tex
\begin{figure}[t]
    \centering
    \includegraphics[width=\linewidth]{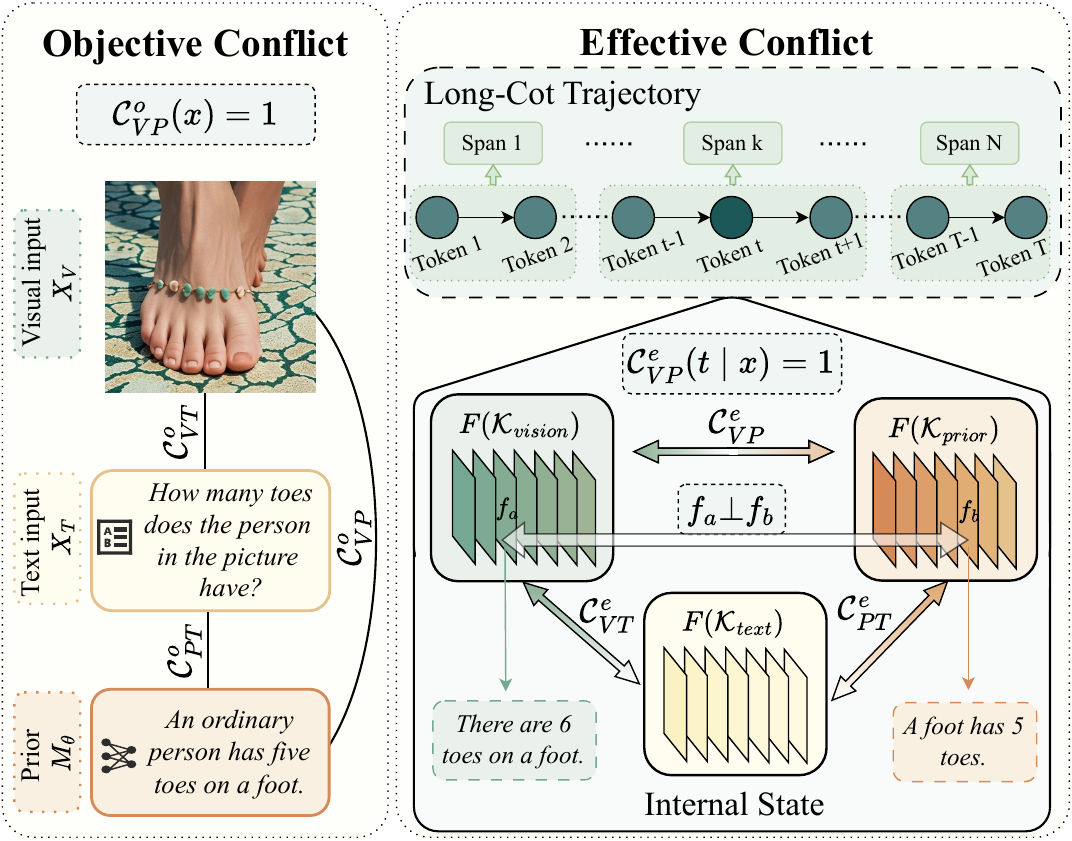}
    \vspace{-1.2em}
    \caption{\textbf{Overview of Knowledge Sources and Conflict Types.} We categorize knowledge into Visual ($\mathcal{K}_{\text{vision}}$), Textual ($\mathcal{K}_{\text{text}}$), and Parametric Prior ($\mathcal{K}_{\text{prior}}$). Knowledge conflicts arise when factual statements from different sources act as incompatible signals. We define three primary conflict types: Vision-Text ($\mathcal{C}_{\mathrm{VT}}$), Vision-Prior ($\mathcal{C}_{\mathrm{VP}}$), and Prior-Text ($\mathcal{C}_{\mathrm{PT}}$).}
    \label{fig:method_framework}
    \vspace{-1.7em}
\end{figure}

\begin{table*}[t]
\centering
\caption{\textbf{Output-level conflict profile across models (objective conflict subsets).} We present statistics of generated trajectories under three types of conflict (model details in Appendix~\ref{app:implementation}).
Metrics reported include sample count, average CoT length, average conflict spans per sample (\textit{spans are contiguous conflict segments identified via an automated LLM annotation pipeline, may consist of one or multiple tokens.}), conflict token density (proportion of conflicting tokens), and sample conflict rate (\% of samples exhibiting \emph{effective conflict}).}
\label{tab:conflict_profile}
\vspace{-0.7em}

\resizebox{\textwidth}{!}{
\begin{tabular}{l|cccc|cccc|cccc}
\toprule
\multirow{2}{*}{\textbf{Metric}} 
& \multicolumn{4}{c|}{\textbf{Llama-3.2V-11B-cot}} 
& \multicolumn{4}{c|}{\textbf{R1-Onevision-7B}} 
& \multicolumn{4}{c}{\textbf{Ocean-R1-7B-Instruct}} \\
\cmidrule(lr){2-5} \cmidrule(lr){6-9} \cmidrule(lr){10-13}
& $\mathcal{C}^o_{\mathrm{VP}}=1$ & $\mathcal{C}^o_{\mathrm{VT}}=1$ & $\mathcal{C}^o_{\mathrm{PT}}=1$ & All
& $\mathcal{C}^o_{\mathrm{VP}}=1$ & $\mathcal{C}^o_{\mathrm{VT}}=1$ & $\mathcal{C}^o_{\mathrm{PT}}=1$ & All
& $\mathcal{C}^o_{\mathrm{VP}}=1$ & $\mathcal{C}^o_{\mathrm{VT}}=1$ & $\mathcal{C}^o_{\mathrm{PT}}=1$ & All \\
\midrule
\textbf{Samples} 
& 749 & 1012 & 803 & 2564
& 724 & 993 & 769 & 2486
& 640 & 1026 & 807 & 2473 \\

\textbf{Avg. CoT length (tokens)} 
& 326.79 & 1768.85 & 238.50 & 868.32
& 706.85 & 790.63 & 558.97 & 694.57
& 488.15 & 711.26 & 302.97 & 520.28 \\

\textbf{Avg. conflict spans per sample} 
& 2.69 & 6.20 & 4.04 & 4.50
& 3.66 & 6.73 & 7.02 & 5.93
& 8.68 & 9.00 & 5.43 & 7.75 \\

\textbf{Conflict token density (\%)} 
& 4.92 & 1.65 & 11.25 & 5.61
& 3.20 & 2.16 & 7.68 & 4.17
& 8.70 & 3.23 & 11.77 & 7.43 \\

\textbf{Conflict Sample Ratio (\%)} 
& 63.68 & 82.21 & 86.43 & 78.12
& 59.67 & 85.90 & 87.91 & 78.88
& 88.75 & 90.25 & 89.34 & 89.57 \\
\bottomrule
\end{tabular}
}
\vspace{-1.5em}
\end{table*}

\vspace{-0.5em}
\section{Conflict in Multimodal Reasoning}~\label{sec:3}
\vspace{-2.5em}
\subsection{Knowledge Sources and Pairwise Conflicts}\label{sec:3.1}
\vspace{-0.3em}
We consider a multimodal long-CoT reasoning task with input $x = (X_V, X_T)$, where $X_V$ denotes the visual input and $X_T$ the textual input. Given a multimodal generative model $M_\theta$, reasoning unfolds as a sequence of tokens $\tau(x) = (y_1, y_2, \ldots, y_T)$, with each token sampled as
{\abovedisplayskip=6pt
\belowdisplayskip=6pt
\begin{equation}
y_t \sim M_\theta(\cdot \mid x, y_{<t}).
\end{equation}}
We denote the internal state at step $t$ by
{\abovedisplayskip=6pt
\belowdisplayskip=6pt
\begin{equation}
\mathbf{h}_t = f_\theta(x, y_{<t}),
\end{equation}}
where $f_\theta$ denotes the model's hidden representation extraction, i.e., the forward pass up to a specified layer.

To analyze how factual inconsistencies arise during reasoning, we abstract the knowledge available to the model into three sources,
$\mathcal{K} = \{\mathcal{K}_{\text{vision}}, \mathcal{K}_{\text{text}}, \mathcal{K}_{\text{prior}}\}.$

Here, $\mathcal{K}_{\text{vision}}$ consists of facts supported by the visual input $X_V$, $\mathcal{K}_{\text{text}}$ consists of facts constrained by the textual input $X_T$, and $\mathcal{K}_{\text{prior}}$ denotes parametric prior knowledge implicitly encoded in the model parameters $\theta$. 

For each knowledge source $\mathcal{K}_* \in \mathcal{K}$, we represent its supported factual content as a set of atomic factual statements $F(\mathcal{K}_*)$, where each element $\psi \in F(\mathcal{K}_*)$ corresponds to an indivisible factual judgment. We use $\psi_a \perp \psi_b$ to denote that two facts are semantically incompatible, i.e., they cannot simultaneously be true under the given context.

Based on this notion, we define a \emph{pairwise knowledge conflict} between two sources $\mathcal{K}_i$ and $\mathcal{K}_j$ ($i \ne j$) as the set of incompatible fact pairs:
{\abovedisplayskip=6pt
\belowdisplayskip=6pt
\begin{equation}
\mathcal{C}_{i,j} = \{(\psi_i, \psi_j) \mid \psi_i \in F(\mathcal{K}_i),\psi_j \in F(\mathcal{K}_j),\psi_i \perp \psi_j\}.
\end{equation}}
In this work, we focus on three primary pairwise conflict types induced by the three knowledge sources: Vision-Prior ($\mathcal{C}_{\mathrm{VP}}$), Vision-Text ($\mathcal{C}_{\mathrm{VT}}$), and Prior-Text ($\mathcal{C}_{\mathrm{PT}}$).

\subsection{Objective vs.\ Effective Conflict} \label{subsec: objeff conflict}

As illustrated in Figure~\ref{fig:method_framework}, we distinguish between two related but fundamentally different notions: 
\emph{objective conflict}, which is defined at the input level, and \emph{effective conflict}, which manifests as a process-level state during reasoning.

\textbf{Objective Conflict}
describes factual inconsistency induced by the input and the model's parametric priors, independent of any particular reasoning trajectory. 
Given a conflict type $\mathcal{C}_{i,j} \in \{\mathcal{C}_{\mathrm{VP}}, \mathcal{C}_{\mathrm{VT}}, \mathcal{C}_{\mathrm{PT}}\}$, we define a binary variable $\mathcal{C}^o_{i,j}(x)\in\{0,1\}$ to indicate whether the input $x$ exhibits an \emph{objective conflict} of type $\mathcal{C}_{i,j}$.
For example, $\mathcal{C}^o_{\mathrm{VP}}(x)=1$ indicates that the visual evidence $X_V$ contradicts the parametric prior knowledge encoded in $\theta$ with respect to a specific fact.
By definition, $\mathcal{C}^o_{i,j}(x)$ depends only on the factual relations supported by the input $x$ and the model priors, and does not reference the reasoning process itself.

Importantly, the presence of an \emph{objective conflict} does not by itself determine whether the model will engage with this conflict during inference.
From the input-level specification alone, it is not directly inferable whether, when, or how a given conflict influences the model's internal reasoning dynamics.
This gap motivates a process-level notion that captures conflict activation within the model.

\textbf{Effective Conflict} characterizes whether an \emph{objective conflict} is actually triggered during reasoning and reflected in the model's internal state.
Concretely, we use $\mathcal{C}^e_{i,j}(t\mid x)\in\{0,1\}$ to indicate whether, at reasoning step $t$, the model relies on mutually incompatible factual information of type $\mathcal{C}_{i,j}$.
Here, $\mathcal{C}^e_{i,j}(t\mid x)=1$ means that the corresponding conflict is active and influences the current reasoning step, as encoded in the internal state at that step.

The relationship between the two notions is asymmetric:
{\abovedisplayskip=6pt
\belowdisplayskip=6pt
\begin{equation}
\mathbb{P}\!\left(\mathcal{C}^e_{i,j}(t \mid x)=1 \mid \mathcal{C}^o_{i,j}(x)=1\right) < 1.
\end{equation}}
That is, \emph{objective conflict} captures \emph{whether} a conflict exists at the input level, whereas \emph{effective conflict} captures \emph{whether and when} that conflict is activated in the model's internal state during reasoning.
The former is induced jointly by the input and priors, while the latter is both model-dependent and process-dependent.

\textbf{Objective conflict data construction.}
For mechanistic analysis, we construct an objective-conflict benchmark with isolated pairwise conflicts, where each example contains exactly one conflict type (\textsc{VP}, \textsc{VT}, or \textsc{PT}) and is intended to elicit \emph{effective conflict} states. This setting is designed as a diagnostic stress-test of conflict arbitration under contradiction, rather than an estimate of in-the-wild conflict prevalence. For each input x, we generate a long-CoT trajectory and align the input-level labels $\mathcal{C}^o_{i,j}(x)$ with step-level \emph{effective conflict} signals $\{\mathcal{C}^e_{i,j}(t\mid x)\}_{t=1}^{T}$ inferred from the model outputs. Table~\ref{tab:conflict_profile} reports conflict activation statistics for this benchmark. Full details are provided in Appendix~\ref{app:construction}.

%% file: secs/4.conflict_detect.tex
\vspace{-0.5em}
\section{Probing Conflict from Internal States}
\label{sec:4_probe}

In Section~\ref{sec:3}, we formalize knowledge conflict as an input-level $\mathcal{C}^o_{i,j}(x)$ and a process-level $\mathcal{C}^e_{i,j}(t \mid x)$. Moving forward, this section addresses the core question: \emph{Is $\mathcal{C}^e_{i,j}(t \mid x)$ reflected in the model's internal states, and can it be identified in a streaming manner during generation?} 

\subsection{Token-level Probing of Knowledge Conflict}
\label{sec:4.1}

We construct a streaming detector: when generating the $t$-th token, it determines whether an effective conflict $\mathcal{C}^e_{i,j}(t \mid x)$ is triggered based solely on the hidden state $\mathbf{h}_t^{(l)}$. 
While prior work has employed probes for binary hallucination detection~\citep{obeso2025real}, we extend this to a four-class classification task based on the definition in Section~\ref{subsec: objeff conflict}.

Here, we use $z=0$ as label, to indicate that no conflict is triggered (i.e., $\mathcal{C}^e_{i,j}(t \mid x)=0, \forall \mathcal{C}_{i,j}$); while $z\in\{1,2,3\}$ corresponds to the active state of specific pairwise knowledge conflicts $\mathcal{C}^e_{i,j}(t \mid x)=1$, namely $\mathcal{C}_{\mathrm{VP}}$, $\mathcal{C}_{\mathrm{PT}}$, and $\mathcal{C}_{\mathrm{VT}}$.

Formally, we define a probe $f_\phi$ that maps hidden states to a probability distribution over conflict labels:
{\abovedisplayskip=6pt
\belowdisplayskip=6pt
\begin{equation}
P_\phi(z \mid \mathbf{h}_t^{(l)})=\mathrm{Softmax}\!\left(f_\phi(\mathbf{h}_t^{(l)})\right),\; z\in \{0,1,2,3\}.
\end{equation}}
The supervision signal for training $f_\phi$ comes from the span-level assertion annotations constructed in Table~\ref{tab:conflict_profile}. We project the label of each annotated span to all its constituent tokens to obtain the dense label sequence $\{z_t\}$.

Since conflict tokens are extremely sparse in long-CoT, we train the probe using a weighted cross-entropy objective:
{\abovedisplayskip=6pt
\belowdisplayskip=6pt
\begin{equation}
\mathcal{L}_{\text{probe}}
= -\sum_{t} w_t \log P_\phi(z_t\mid \mathbf{h}_t^{(l)}),
\end{equation}}
where $w_t$ is a sample weight that assigns higher weight to $z\in\{1,2,3\}$ (i.e., tokens where knowledge conflict $\mathcal{C}_{i,j}$ occurs), preventing the probe from degenerating into predicting only the no-conflict background class. This objective allows the probe to maintain overall stability while remaining sufficiently sensitive to critical conflict-triggering moments.
Full training details are provided in Appendix~\ref{app:probe_training}.

\subsection{Verifying the Separability of Knowledge Conflicts}
\label{sec:4.2}

We evaluate whether the probe reliably diagnoses knowledge conflicts from internal states. Specifically, we examine the token-level separability of \emph{effective conflicts} and whether their sample-level recovers the \emph{objective conflict types}.

\begin{figure}[ht]
    \centering
    \vspace{-1.0em}
    \includegraphics[width=\linewidth]{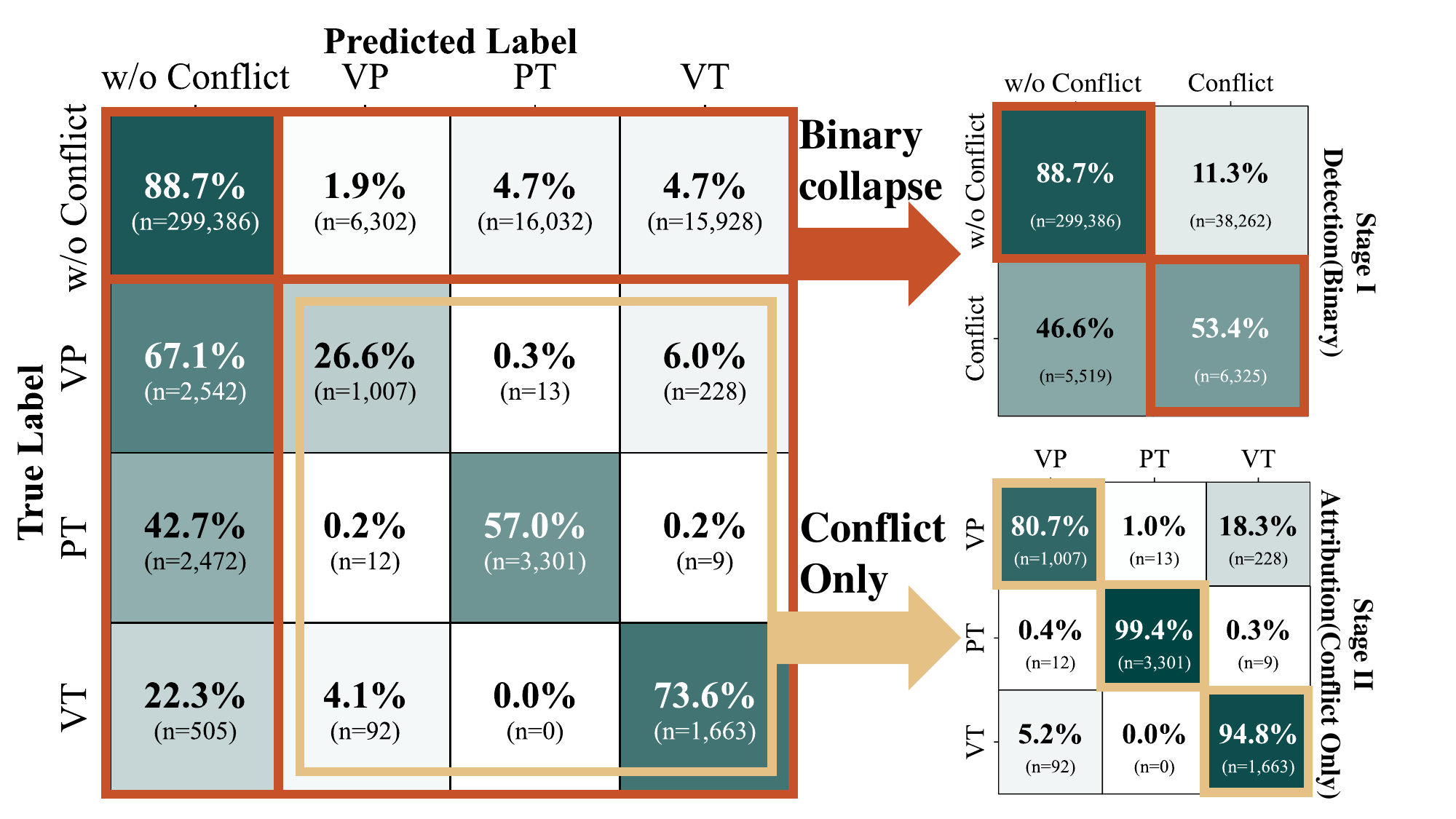}
    \vspace{-1.5em}
    \caption{\textbf{Token-level separability of effective conflict $\mathcal{C}^e_{i,j}(t \mid x)$.} The left panel shows the confusion matrix over token-level conflict predictions. The right panels decompose performance into binary detection of conflict versus no-conflict, and fine-grained attribution among conflict types. Values denote row-normalized recall.}
    \label{fig:token_separability}
    \vspace{-1em}
\end{figure}

\paragraph{(I) Separability of Effective Conflicts: Local Signals in Sparse Regimes.}
We first examine whether the probe can distinguish different types of \emph{effective conflicts} $\mathcal{C}^e_{i,j}(t \mid x)$ from the model’s internal states during reasoning.

As shown in Figure~\ref{fig:token_separability}, the probe demonstrates robust discrimination capabilities. In the binary detection stage (Stage I), the model achieves a high True Negative rate of 88.7\%, effectively filtering out non-conflicting steps. Conversely, a False Negative rate of 46.6\% is observed, primarily driven by semantic sparsity within conflict spans—where 67.1\% of $\mathcal{C}_{\mathrm{VP}}$ tokens are misclassified as non-conflicting due to weak local signals. However, once \emph{effective conflict} is activated (Stage II), the separability between conflict types sharply increases: $\mathcal{C}_{\mathrm{PT}}$ and $\mathcal{C}_{\mathrm{VT}}$ achieve near-perfect identification accuracies of 99.4\% and 94.8\%, respectively. Even $\mathcal{C}_{\mathrm{VP}}$, the most subtle type, sees its recognition accuracy jump from 26.6\% in the global view to 80.7\% in the conditioned view. The minimal off-diagonal confusion ($<1\%$ between PT and others) confirms that \emph{effective conflict} types possess distinct, highly separable internal representations.

\noindent\fbox{%
    \parbox{\linewidth}{%
\textbf{Conclusion (Local Effective Conflicts):}
Even under extreme sparsity and noise, different types of effective knowledge conflicts $\mathcal{C}^e_{i,j}(t \mid x)$ give rise to distinct local structures in the model’s internal states that can be reliably captured by the probe. This validates the feasibility of \textbf{streaming diagnosis} of \textit{effective conflicts} while revealing differences in their intrinsic detectability.
    }
}

\begin{figure}[ht]
    \centering
    \includegraphics[width=\linewidth]{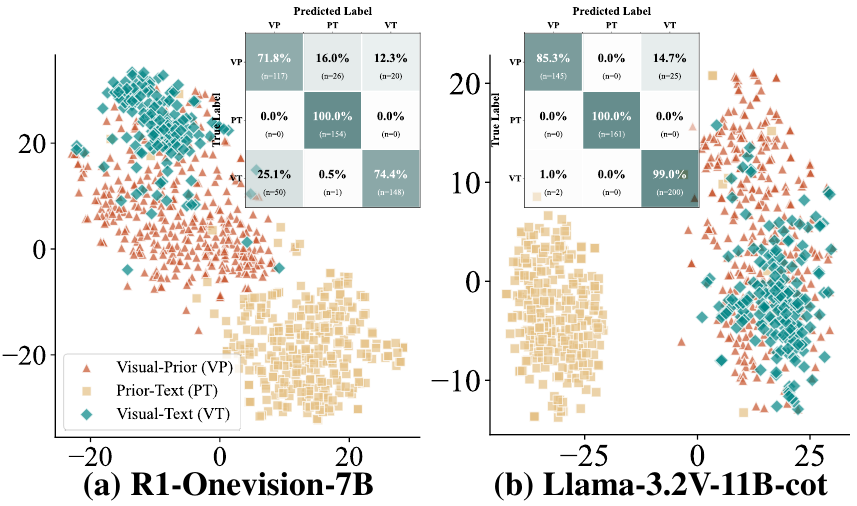}
    \vspace{-1.5em}
    \caption{\textbf{Sample-level separability of conflict types.} We visualize the t-SNE projection of hidden states at layer 20 (\llmname{R1-Onevision}) and layer 39 (\llmname{Llama-3.2V}). The three conflict categories are colored according to their \textbf{Objective Conflict} labels, pre-defined during dataset construction. The top-right confusion matrices illustrate the sample-level attribution performance.
    }
    \label{fig:sample_separability}
    \vspace{-1.0em}
\end{figure}

\begin{figure*}[ht]
    \centering
    \includegraphics[width=\linewidth]{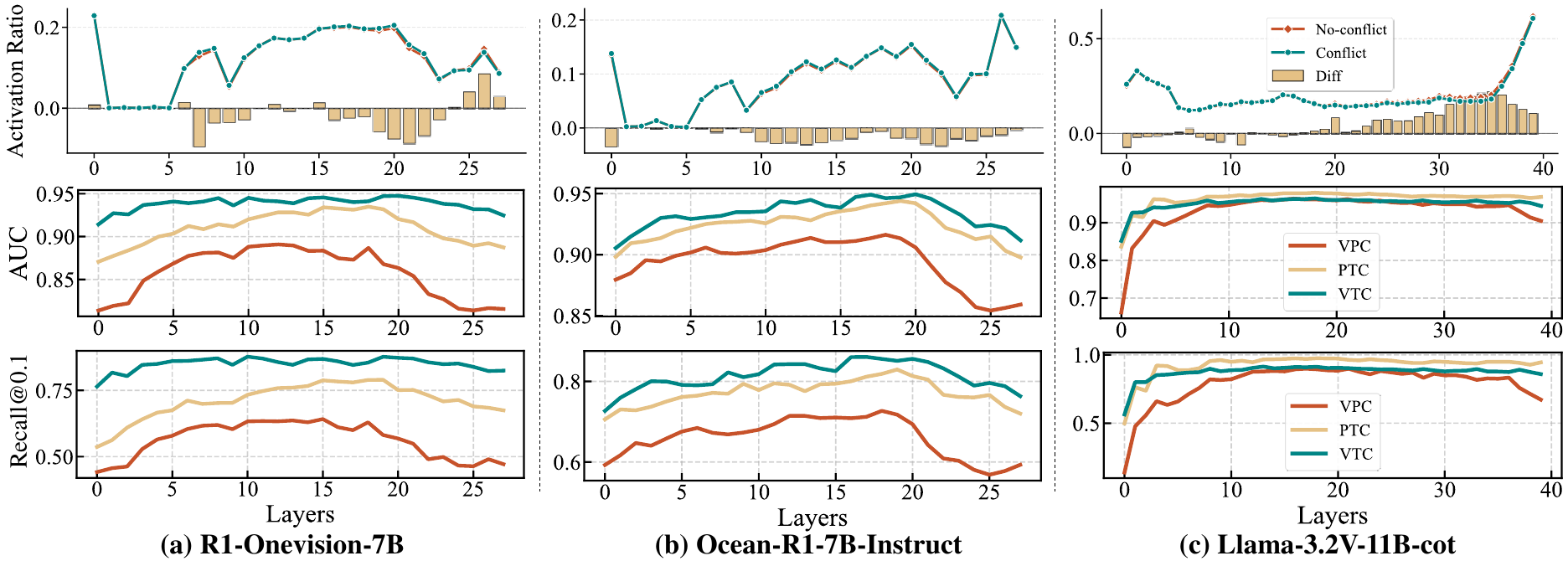}
    \vspace{-1.6em}
    \caption{\textbf{Cross-layer distribution of conflict signals.}
    Top row: attention-head activation ratio on \textit{conflict} tokens vs.\ \textit{no-conflict} tokens (lines), and their difference (bars), computed using effective conflict labels.
    Middle/bottom rows: layer-wise probe performance (one-vs-rest AUC and Recall@0.1) for $\mathcal{C}_{\mathrm{VP}},\mathcal{C}_{\mathrm{PT}},\mathcal{C}_{\mathrm{VT}}$ across three MLLM backbones.}
    \vspace{-1.0em}
    \label{fig:layer_distribution}
\end{figure*}

\paragraph{(II) Alignment to Objective Conflicts: Aggregating Effective Signals.}

We next examine whether aggregating local \emph{effective conflicts} $\mathcal{C}^e_{i,j}(t \mid x)$ along a reasoning trajectory recovers the corresponding \emph{objective conflict} $\mathcal{C}^o_{i,j}(x)$ defined at the input level.
This analysis evaluates the robustness of \emph{effective conflict} signals beyond individual steps.

For each long-CoT trajectory, we aggregate hidden states of activated \emph{effective conflicts} via mean pooling to obtain a sample-level representation. 
We visualize these representations using t-SNE (Figure~\ref{fig:sample_separability}), where samples sharing the same \emph{objective conflict} type form compact clusters that are well separated, indicating consistent global structure.

Quantitatively, we infer the \emph{objective conflict} type by aggregating stepwise \emph{effective conflict} activations:
{\abovedisplayskip=6pt
\belowdisplayskip=6pt
\begin{equation}
\hat{\mathcal{C}}_{\text{sample}}=
\arg\max_{\mathcal{C}_{i,j}}
\sum_{t=1}^{T}
\mathbb{I}\!\left[\mathcal{C}^e_{i,j}(t \mid x)=1\right].
\end{equation}}

Comparing $\hat{\mathcal{C}}_{\text{sample}}$ with the ground-truth objective labels $\mathcal{C}^o_{i,j}(x)$ directly tests whether the model’s internal conflict aligns with the conflict structure inherent in the input.

As shown in the inset matrices of Figure~\ref{fig:sample_separability}, aggregation substantially enhances separability. Notably, $\mathcal{C}_{\mathrm{PT}}$ achieves a perfect 100.0\% on both \llmname{R1-Onevision} and \llmname{Llama-3.2V}, confirming that text-prior conflicts induce unique and stable shifts in internal states. The remaining confusion is largely confined to the visual-conflict types: for instance, 25.1\% of $\mathcal{C}_{\mathrm{VT}}$ samples in \llmname{R1-Onevision} are misclassified as $\mathcal{C}_{\mathrm{VP}}$, and 14.7\% of $\mathcal{C}_{\mathrm{VP}}$ samples in \llmname{Llama-3.2V} are misidentified as $\mathcal{C}_{\mathrm{VT}}$. This overlap is expected, as both categories involve failures in processing visual evidence, leading to partially shared representations.


\noindent\fbox{%
    \parbox{\linewidth}{%
\textbf{Conclusion (Global Effective Confilcts)}:
By aggregating stepwise \emph{effective conflict} signals $\mathcal{C}^e_{i,j}(t \mid x)$ along the reasoning trajectory, different \emph{objective conflict} types $\mathcal{C}^o_{i,j}(x)$ become clearly and robustly separable at the \textbf{sample level}. This indicates that \emph{effective conflicts} are not merely local artifacts, but form consistent \textbf{global} patterns that reliably reflect the underlying input-level objective conflict structure.
}}
\vspace{-0.5em}

\subsection{Cross-Layer Distribution of Conflict Signals}
\label{sec:4.3}

We scan model depth to localize where \emph{effective} knowledge conflicts are most strongly encoded.
Concretely, for each layer $l$, we train the same token-level probe on hidden states $\mathbf{h}^{(l)}_{t}$ and evaluate its one-vs-rest AUC / Recall@0.1 for
$\{\mathcal{C}_{\mathrm{VP}},\mathcal{C}_{\mathrm{PT}},\mathcal{C}_{\mathrm{VT}}\}$.

\input{secs/table_auc_rec.tex}

Beyond probe separability, we also quantify a lightweight mechanistic correlate~\citep{huang2025parammute}: how attention-head activations differ between \textit{conflict} and \textit{no-conflict} token positions.
Let $\mathcal{A}^{(l)}$ denote the set of attention heads at layer $l$, and let $\mathbf{o}^{(l,a)}_{t}$ be the output of head $a\in\mathcal{A}^{(l)}$ at token $t$.
We define token sets using effective conflict signals:
{\abovedisplayskip=6pt
\belowdisplayskip=6pt
\begin{align}
    \mathcal{S}_{\mathrm{conf}} &= \{(x,t) \mid \exists(i,j),~ \mathcal{C}^{e}_{i,j}(t \mid x)=1\}, \\
    \mathcal{S}_{\mathrm{nconf}} &= \{(x,t) \mid \forall(i,j),~ \mathcal{C}^{e}_{i,j}(t \mid x)=0\}.
\end{align}}
The layer-wise head activation ratio on a token set $\mathcal{S}$ is
{\abovedisplayskip=6pt
\belowdisplayskip=6pt
\begin{equation}
R^{(l)}(\mathcal{S})
=
\mathbb{E}_{(x,t)\in\mathcal{S}}\;
\frac{1}{|\mathcal{A}^{(l)}|}\sum_{a\in\mathcal{A}^{(l)}}
\mathbb{I}\!\left[\left\|\mathbf{o}^{(l,a)}_{t}\right\|_{2}>\gamma\right],
\end{equation}}
where $\gamma$ is a fixed activation threshold (details in Appendix~\ref{app:activation_threshold}).
We then report the activation drift
{\abovedisplayskip=6pt
\belowdisplayskip=6pt
\begin{equation}
\Delta R^{(l)} = R^{(l)}(\mathcal{S}_{\mathrm{conf}}) - R^{(l)}(\mathcal{S}_{\mathrm{nconf}}),
\end{equation}}
which measures how strongly attention activations shift when effective conflicts are triggered.

As shown in Figure~\ref{fig:layer_distribution}, both measurements reveal distinct depth-dependent signatures.
(\textbf{I}) \textbf{Probe Separability}: In 7B models (\llmname{R1-Onevision}, \llmname{Ocean-R1}), discrimination performance rises in early layers and maximizes in the mid-to-late block (Layers 15--22), where AUC scores for $\mathcal{C}_{\mathrm{PT}}$ and $\mathcal{C}_{\mathrm{VT}}$ consistently exceed 93\%, before declining in the final layers. \llmname{Llama-3.2V} pushes this saturation deeper, maintaining highly robust separability ($\geq$95\%) as deep as Layer 39.
(\textbf{II}) \textbf{Activation Drift}: This aligns with attention shifts. R1-series models show negative drift (suppression) peaking at Layers 18--22, while \llmname{Llama-3.2V} displays positive drift (enhancement) in Layers 30--39. We term these co-located peaks (Layer 20 for 7B, 39 for 11B) the \emph{conflict encoding stage}, anchoring our analysis.

\noindent\fbox{%
\parbox{\linewidth}{%
\textbf{Conclusion (Layer-level):}
Layer-scanning reveals that both probe separability and attention drift co-localize in a specific mid-to-late layer band across all three MLLM backbones.
This indicates that \textit{conflict-related signals} are \textbf{depth-dependent} and concentrated in a distinct ``conflict encoding stage,'' bridging early perception and late decoding rather than being uniformly distributed across the network.
}}

\subsection{Linearity of Conflict Representation}
\label{sec:4.4}

To comprehensively assess the nature of effective conflict signals $\mathcal{C}^e_{i,j}(t \mid x)$ encoded in the hidden states $\mathbf{h}_t^{(l)}$ (specifically, whether they are explicitly linear or highly entangled) we conducted experiments on specific layers identified as the ``Conflict Encoding Stage'' in Section~\ref{sec:4.3}.
We designed two probe architectures with distinct underlying assumptions:
(\textbf{I}) \textbf{Linear Probe ($f_{lin}$)}, consisting of a single projection layer $\mathbf{W} \in \mathbb{R}^{d \times 4}$ (where $d$ denotes the hidden state dimension), aimed at evaluating the \textit{Linear Separability} of conflict states. High classification accuracy with a linear mapping would indicate that the model has formed clear, decoupled conflict boundaries at the current layer.
(\textbf{II}) \textbf{MLP Probe ($f_{mlp}$)}, designed to assess \textit{Non-linear Entanglement}. Recognizing the potential manifold complexity in deep Transformer features, we construct a deep MLP with three dimension-reducing layers (1024 $\to$ 512 $\to$ 256) and ReLU activation to capture high-order interaction features.

As shown in Table~\ref{tab:probe_performance}, we report AUC and Recall@0.1 for both probes using ``Span-Max'' aggregation, which takes the maximum predicted probability across tokens within each span (details in Appendix~\ref{app:span_aggregation}).
The Linear Probe achieves strong performance across all conflict types: AUC reaches 93.2--98.8\% and Recall@0.1 reaches 76.9--97.8\%.
For $\mathcal{C}_{\mathrm{PT}}$, Linear Probe achieves 98.6\% AUC and 96.1--97.2\% Recall@0.1; for $\mathcal{C}_{\mathrm{VP}}$ and $\mathcal{C}_{\mathrm{VT}}$, it reaches 93.4--95.9\% AUC and 76.9--87.9\% Recall@0.1, comparable to MLP.
The fact that a single linear layer suffices to achieve such performance indicates that for knowledge conflicts, the ``features'' extracted by LLMs are already explicitly disentangled in the high-dimensional space, and introducing additional non-linear complexity (MLP) does not yield significant gain.

\noindent\fbox{%
    \parbox{\linewidth}{%
\textbf{Conclusion (Linearity)}:
It was observed that a simple linear probing method could achieve detection performance comparable to that of a non-linear MLP. This suggests that\textit{ effective conflicts} are not entangled within complex non-linear manifolds, but rather are explicitly and approximately \textbf{linearly separable}. This makes real-time detection of conflict states during inference possible.
}}

%% file: secs/table_auc_rec.tex
\begin{table*}[t]
\centering
\tiny
\renewcommand{\arraystretch}{0.6} 
\setlength{\tabcolsep}{4pt} 
\caption{\textbf{Assessment of conflict probe performance across three VLM backbones.} We report AUC and Recall at FPR=0.1 (Rec@0.1) under the One-vs-Rest setting. 
\colorbox{graybg}{Gray rows} indicate the \textbf{Span-Max} aggregation, which consistently outperforms token-level baselines. Values are presented as percentages (\%).}
\label{tab:probe_performance}
\vspace{-1.0em}
\resizebox{\textwidth}{!}{%
\begin{tabular}{lll cccc cccc}
\toprule
\multirow{2.5}{*}{\textbf{Models}} & \multirow{2.5}{*}{\textbf{Probe}} & \multirow{2.5}{*}{\textbf{Granularity}} & \multicolumn{4}{c}{\textbf{AUC (\%)}} & \multicolumn{4}{c}{\textbf{Recall@0.1 (\%)}} \\
\cmidrule(lr){4-7} \cmidrule(lr){8-11}
& & & \textbf{w/o Conflict} & \textbf{$\mathcal{C}^o_\mathrm{VP}=1$} & \textbf{$\mathcal{C}^o_\mathrm{PT}=1$} & \textbf{$\mathcal{C}^o_\mathrm{VT}=1$} & \textbf{w/o Conflict} & \textbf{$\mathcal{C}^o_\mathrm{VP}=1$} & \textbf{$\mathcal{C}^o_\mathrm{PT}=1$} & \textbf{$\mathcal{C}^o_\mathrm{VT}=1$} \\
\midrule

\multirow{6}{*}{\rotatebox[origin=c]{90}{\shortstack[c]{\textbf{R1-Onevision}\\(7B)}}} 
 & \multirow{3}{*}{Linear} 
   & All Token & 81.7$_{\pm 0.1}$ & 86.3$_{\pm 0.2}$ & 92.0$_{\pm 0.1}$ & 94.8$_{\pm 0.2}$ & 50.0$_{\pm 0.3}$ & 56.8$_{\pm 0.2}$ & 75.1$_{\pm 0.1}$ & 87.3$_{\pm 0.3}$ \\
 & & Span Only & 76.8$_{\pm 0.2}$ & 82.5$_{\pm 0.1}$ & 90.8$_{\pm 0.3}$ & 95.4$_{\pm 0.2}$ & 35.5$_{\pm 0.2}$ & 44.5$_{\pm 0.3}$ & 70.5$_{\pm 0.2}$ & 88.5$_{\pm 0.1}$ \\
 & & \cellcolor{graybg}\textbf{Span-Max} & \cellcolor{graybg}\textbf{93.2$_{\pm 0.1}$} & \cellcolor{graybg}\textbf{94.2$_{\pm 0.2}$} & \cellcolor{graybg}\textbf{98.6$_{\pm 0.1}$} & \cellcolor{graybg}\textbf{97.3$_{\pm 0.1}$} & \cellcolor{graybg}\textbf{81.5$_{\pm 0.2}$} & \cellcolor{graybg}\textbf{82.4$_{\pm 0.1}$} & \cellcolor{graybg}\textbf{97.2$_{\pm 0.1}$} & \cellcolor{graybg}\textbf{93.8$_{\pm 0.2}$} \\
\cmidrule{2-11}
 & \multirow{3}{*}{MLP} 
   & All Token & 95.5$_{\pm 0.1}$ & 90.4$_{\pm 0.2}$ & 85.2$_{\pm 0.3}$ & 94.1$_{\pm 0.1}$ & 89.1$_{\pm 0.2}$ & 68.4$_{\pm 0.1}$ & 62.7$_{\pm 0.2}$ & 79.3$_{\pm 0.1}$ \\
 & & Span Only & 95.7$_{\pm 0.2}$ & 86.1$_{\pm 0.3}$ & 80.3$_{\pm 0.1}$ & 93.3$_{\pm 0.2}$ & 89.8$_{\pm 0.1}$ & 53.0$_{\pm 0.2}$ & 43.7$_{\pm 0.2}$ & 76.8$_{\pm 0.3}$ \\
 & & \cellcolor{graybg}\textbf{Span-Max} & \cellcolor{graybg}\textbf{97.3$_{\pm 0.1}$} & \cellcolor{graybg}\textbf{94.5$_{\pm 0.1}$} & \cellcolor{graybg}\textbf{93.2$_{\pm 0.2}$} & \cellcolor{graybg}\textbf{99.1$_{\pm 0.1}$} & \cellcolor{graybg}\textbf{93.4$_{\pm 0.2}$} & \cellcolor{graybg}\textbf{82.4$_{\pm 0.1}$} & \cellcolor{graybg}\textbf{82.1$_{\pm 0.1}$} & \cellcolor{graybg}\textbf{98.7$_{\pm 0.2}$} \\
\midrule

\multirow{6}{*}{\rotatebox[origin=c]{90}{\shortstack[c]{\textbf{Ocean-R1}\\(7B-Instruct)}}} 
 & \multirow{3}{*}{Linear} 
   & All Token & 83.0$_{\pm 0.2}$ & 90.6$_{\pm 0.1}$ & 94.2$_{\pm 0.2}$ & 94.9$_{\pm 0.1}$ & 53.7$_{\pm 0.3}$ & 69.4$_{\pm 0.1}$ & 81.3$_{\pm 0.2}$ & 85.6$_{\pm 0.1}$ \\
 & & Span Only & 78.5$_{\pm 0.1}$ & 86.7$_{\pm 0.3}$ & 90.0$_{\pm 0.2}$ & 97.6$_{\pm 0.1}$ & 41.4$_{\pm 0.2}$ & 52.5$_{\pm 0.2}$ & 66.6$_{\pm 0.1}$ & 94.6$_{\pm 0.3}$ \\
 & & \cellcolor{graybg}\textbf{Span-Max} & \cellcolor{graybg}\textbf{95.0$_{\pm 0.2}$} & \cellcolor{graybg}\textbf{95.9$_{\pm 0.1}$} & \cellcolor{graybg}\textbf{98.6$_{\pm 0.1}$} & \cellcolor{graybg}\textbf{98.8$_{\pm 0.1}$} & \cellcolor{graybg}\textbf{85.7$_{\pm 0.1}$} & \cellcolor{graybg}\textbf{87.9$_{\pm 0.2}$} & \cellcolor{graybg}\textbf{97.1$_{\pm 0.1}$} & \cellcolor{graybg}\textbf{97.8$_{\pm 0.2}$} \\
\cmidrule{2-11}
 & \multirow{3}{*}{MLP} 
   & All Token & 95.5$_{\pm 0.1}$ & 92.8$_{\pm 0.1}$ & 85.0$_{\pm 0.2}$ & 95.5$_{\pm 0.1}$ & 87.1$_{\pm 0.2}$ & 75.6$_{\pm 0.3}$ & 61.6$_{\pm 0.1}$ & 85.2$_{\pm 0.2}$ \\
 & & Span Only & 97.8$_{\pm 0.2}$ & 87.3$_{\pm 0.2}$ & 79.7$_{\pm 0.1}$ & 91.7$_{\pm 0.2}$ & 95.7$_{\pm 0.3}$ & 53.9$_{\pm 0.1}$ & 43.3$_{\pm 0.2}$ & 71.0$_{\pm 0.1}$ \\
 & & \cellcolor{graybg}\textbf{Span-Max} & \cellcolor{graybg}\textbf{99.2$_{\pm 0.1}$} & \cellcolor{graybg}\textbf{96.5$_{\pm 0.1}$} & \cellcolor{graybg}\textbf{95.3$_{\pm 0.2}$} & \cellcolor{graybg}\textbf{98.4$_{\pm 0.1}$} & \cellcolor{graybg}\textbf{98.9$_{\pm 0.1}$} & \cellcolor{graybg}\textbf{89.8$_{\pm 0.2}$} & \cellcolor{graybg}\textbf{87.5$_{\pm 0.1}$} & \cellcolor{graybg}\textbf{96.1$_{\pm 0.1}$} \\
\midrule

\multirow{6}{*}{\rotatebox[origin=c]{90}{\shortstack[c]{\textbf{Llama-3.2V}\\(11B-cot)}}} 
 & \multirow{3}{*}{Linear} 
   & All Token & 88.7$_{\pm 0.2}$ & 90.5$_{\pm 0.1}$ & 96.9$_{\pm 0.2}$ & 94.5$_{\pm 0.1}$ & 68.4$_{\pm 0.3}$ & 67.2$_{\pm 0.2}$ & 94.4$_{\pm 0.1}$ & 85.8$_{\pm 0.2}$ \\
 & & Span Only & 79.6$_{\pm 0.2}$ & 85.8$_{\pm 0.2}$ & 90.2$_{\pm 0.1}$ & 95.2$_{\pm 0.3}$ & 43.2$_{\pm 0.1}$ & 51.1$_{\pm 0.2}$ & 66.0$_{\pm 0.2}$ & 88.4$_{\pm 0.1}$ \\
 & & \cellcolor{graybg}\textbf{Span-Max} & \cellcolor{graybg}\textbf{93.9$_{\pm 0.1}$} & \cellcolor{graybg}\textbf{93.4$_{\pm 0.1}$} & \cellcolor{graybg}\textbf{98.4$_{\pm 0.2}$} & \cellcolor{graybg}\textbf{97.2$_{\pm 0.1}$} & \cellcolor{graybg}\textbf{83.5$_{\pm 0.2}$} & \cellcolor{graybg}\textbf{76.9$_{\pm 0.1}$} & \cellcolor{graybg}\textbf{96.1$_{\pm 0.2}$} & \cellcolor{graybg}\textbf{93.1$_{\pm 0.1}$} \\
\cmidrule{2-11}
 & \multirow{3}{*}{MLP} 
   & All Token & 95.8$_{\pm 0.2}$ & 90.7$_{\pm 0.1}$ & 88.7$_{\pm 0.2}$ & 96.9$_{\pm 0.1}$ & 89.4$_{\pm 0.1}$ & 64.3$_{\pm 0.2}$ & 70.6$_{\pm 0.3}$ & 93.7$_{\pm 0.2}$ \\
 & & Span Only & 96.1$_{\pm 0.1}$ & 85.5$_{\pm 0.3}$ & 79.2$_{\pm 0.2}$ & 89.2$_{\pm 0.1}$ & 90.8$_{\pm 0.2}$ & 46.7$_{\pm 0.1}$ & 40.5$_{\pm 0.2}$ & 65.2$_{\pm 0.1}$ \\
 & & \cellcolor{graybg}\textbf{Span-Max} & \cellcolor{graybg}\textbf{97.2$_{\pm 0.1}$} & \cellcolor{graybg}\textbf{94.5$_{\pm 0.2}$} & \cellcolor{graybg}\textbf{93.4$_{\pm 0.1}$} & \cellcolor{graybg}\textbf{97.8$_{\pm 0.1}$} & \cellcolor{graybg}\textbf{93.2$_{\pm 0.1}$} & \cellcolor{graybg}\textbf{82.3$_{\pm 0.2}$} & \cellcolor{graybg}\textbf{82.3$_{\pm 0.1}$} & \cellcolor{graybg}\textbf{94.4$_{\pm 0.2}$} \\
\bottomrule
\end{tabular}%
}
\vspace{-2.0em}
\end{table*}

%% file: secs/5.intervention.tex
\begin{figure*}[t]
    \centering
    \includegraphics[width=\textwidth]{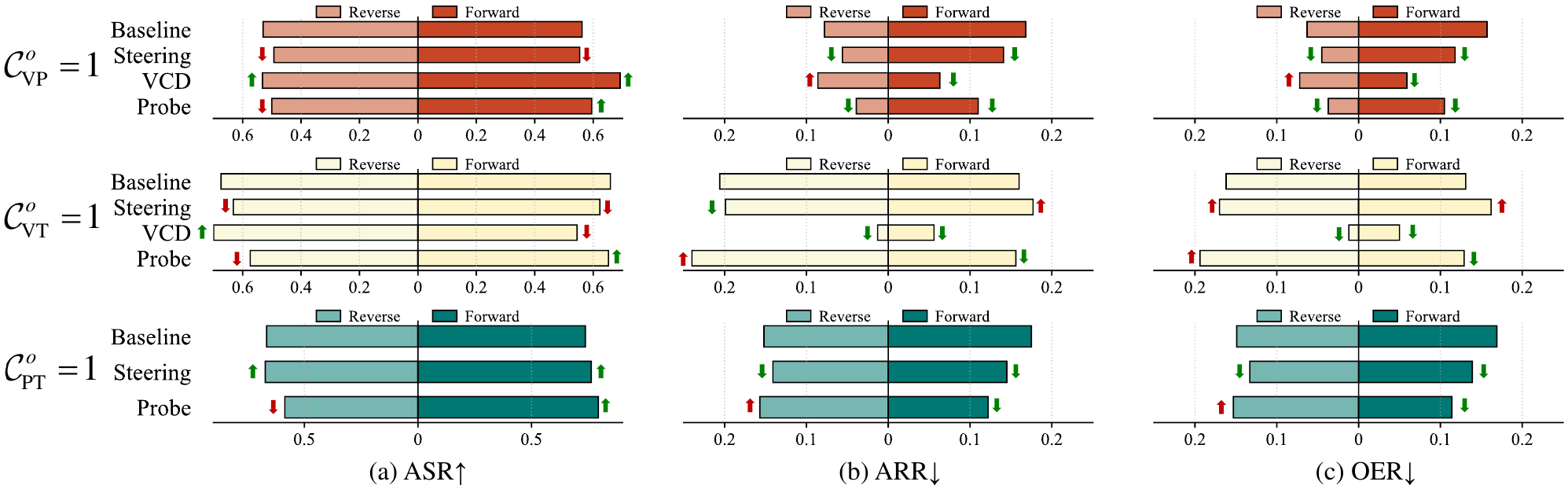}
    \vspace{-1.5em}
    \caption{\textbf{Semantic performance of targeted source control.} 
    We evaluate three conflict subsets ($\mathcal{C}^{o}_{\mathrm{VP}}, \mathcal{C}^{o}_{\mathrm{VT}}, \mathcal{C}^{o}_{\mathrm{PT}}$) using judge-based metrics: 
    \textbf{ASR} (Anchor Support Rate, $\uparrow$), 
    \textbf{ARR} (Anchor Rejection Rate, $\downarrow$), 
    and \textbf{OER} (Obvious Error Rate, $\downarrow$). 
    \textbf{Forward/Reverse} denote intervening toward the truth-anchored (benchmark-reliable) vs. conflicting source. 
    Arrows indicate relative changes against the baseline. 
    Note that VCD is inapplicable to the non-visual $\mathcal{C}^{o}_{\mathrm{PT}}$ subset.}
    \label{fig:sec5_semantic}
    \vspace{-1.0em}
\end{figure*}

\vspace{-1em}
\section{Intervening in Knowledge Conflict}
\label{sec:5_intervention}
\vspace{-0.3em}

Section~\ref{sec:4_probe} showed that \textit{effective conflicts} $\mathcal{C}^{e}_{i,j}(t\mid x)$ are streaming-decodable from internal states
and are encoded as linearly separable features in specific mid-to-late layers.
Building on this observation, we ask the following:
\emph{given an input with $\mathcal{C}^{o}_{i,j}(x)=1$, can inference-time interventions bias the model toward a desired knowledge source,
or suppress the activation of \textit{effective conflicts} during generation?}

\vspace{-0.5em}
\subsection{A unified framework for directional interventions}
\label{sec:5.1}

\paragraph{Two control objectives.}
We study inference-time control under objectively conflicting inputs, two settings are considered.
\textbf{(I) Targeted source control.}
We choose a target source $\mathcal{K}_s \in \{\mathcal{K}_i, \mathcal{K}_j\}$ and intervene so that the model follows $\mathcal{K}_s$ under conflict.
This yields two directions: \textbf{Forward}, which intervenes toward the truth-anchored (benchmark-reliable) source, and \textbf{Reverse}, which enforces the opposite source.
\textbf{(II) Conflict mitigation.}
We measure whether interventions reduce how often \textit{effective conflicts} are activated during generation,
quantified by the expected fraction of reasoning steps where a conflict is detected:
{\abovedisplayskip=6pt
\belowdisplayskip=6pt
\begin{equation}
\mathbb{E}_{x}\,\mathbb{E}_{t}\Big[\mathbb{I}[\mathcal{C}^{e}_{i,j}(t\mid x)=1]\Big].
\label{eq:mitigation_objective}
\end{equation}}
\vspace{-2.5em}
\paragraph{A unified view of directional interventions.}
Let $\ell_t \in \mathbb{R}^{|\mathcal{V}|}$ denote the pre-softmax logits at step $t$.
We view an inference-time intervention as modifying decoding through an additive logit perturbation,
either directly or implicitly via hidden-state manipulation:
{\abovedisplayskip=6pt
\belowdisplayskip=6pt
\begin{equation}
\tilde{p}_t = \mathrm{softmax}(\ell_t + \Delta \ell_t), 
\qquad \Delta \ell_t = \mathcal{I}(x, y_{<t}).
\label{eq:logit_intervention}
\end{equation}}
We consider three instantiations of $\mathcal{I}$: \textbf{(I) Visual contrastive decoding (VCD).}
VCD applies a logit-level correction~\citep{leng2024vcd} and is restricted to conflicts involving visual sources
(i.e., $\mathcal{C}^{o}_{\mathrm{VP}} = 1$ or $\mathcal{C}^{o}_{\mathrm{VT}} = 1$).
\textbf{(II) Representation steering.}
Leveraging the linear separability found in Section~\ref{sec:4_probe}, we adopt a representation steering~\citep{zhang2025evaluating} that shifts the hidden state at a selected conflict-sensitive layer
by a learned direction, i.e., $\tilde{\mathbf{h}}_t = \mathbf{h}_t + \lambda\,\mathbf{v}$ (where $\lambda$ is the steering strength and $\mathbf{v}$ is the direction vector).
\textbf{(III) Probe-guided control.}
We use the streaming probe to score candidate continuations, reweighting decoding toward options less likely to trigger conflicts.
For the top-$k$ candidates $\mathcal{V}_k$ with base probabilities $p_t(w)$, we apply
{\abovedisplayskip=6pt
\belowdisplayskip=6pt
\begin{equation}
\tilde{p}_t(w) \propto p_t(w)\,\exp\!\big(\alpha\,P_t^{(w)}\big), 
\qquad w \in \mathcal{V}_k,
\label{eq:probe_guided}
\end{equation}}
where $P_t^{(w)}$ is the probe-predicted probability of the \texttt{no-conflict} state for the continuation committing to token $w$,
and $\alpha$ controls the strength of guidance.
Full implementation details and hyperparameters are provided in Appendices~\ref{app:interventions} and~\ref{app:sec5_hparams}.

\vspace{-0.5em}
\subsection{Targeted source control: semantic-level evaluation}
\label{sec:5.2}
\vspace{0.5em}
We evaluate whether targeted interventions successfully bias the model toward a specified knowledge source under objectively conflicting inputs.
We adopt an automated assertion-level judge, implemented with a strong off-the-shelf large language model, to assess semantic alignment with the target source.
The judge extracts factual claims from the model output and verifies each claim against the corresponding truth anchor
(image, input text, or world knowledge), producing compact aggregate metrics:
\textbf{ASR} (Anchor Support Rate), \textbf{ARR} (Anchor Refutation Rate), and \textbf{OER} (Obvious Error Rate).
To validate judge reliability, we conducted human verification on a stratified 10\% subset ($\sim$1,500 spans), yielding high inter-annotator agreement ($\kappa=0.87$), confirming that automated verdicts align closely with human perception of conflict resolution (details in Appendix~\ref{app:judge}).



\input{secs/mitigation_tab.tex}

As shown in Figure~\ref{fig:sec5_semantic}, targeted source control is feasible but exhibits a pronounced directional asymmetry. Across objective-conflict subsets, \textbf{Forward} interventions (intervening toward the truth-anchored source; vision for VP/VT and prior knowledge for PT) reliably improve semantic alignment, whereas \textbf{Reverse} control (forcing reliance on the competing source) often degrades it. We hypothesize this asymmetry reflects an internal source-reliability prior: when sources disagree, the model is more resistant to reversing arbitration away from the source it treats as reliable, even under strong contextual pressure.
This asymmetry cannot be explained by construction bias alone: if it were purely a data artifact, we would expect the probe to learn shortcuts to anchor proximity rather than capturing genuine conflict dynamics. However, the asymmetry persists across all three architecturally distinct backbones, suggesting it reflects shared instruction-tuning biases that favor user-provided context~\citep{sharma2024towards,zhang2025evaluating}.
Under \textbf{Forward} control, \textbf{Probe-guided control} interventions improve ASR while lowering OER by $\sim$30\%; \textbf{VCD} yields stronger but selective gains on $\mathcal{C}_{\mathrm{VP}}$ (ASR +15\%, ARR halved). \textbf{Reverse} control remains challenging---most methods regress or show negligible gains.
Mechanistically, the probe primarily \emph{suppresses} conflict states rather than enforcing weaker-source selection. This highlights a trade-off: \textbf{VCD} is high-gain but direction-sensitive, whereas \textbf{Representation steering} reliably reduces errors (ARR/OER) but rarely drives sustained ASR gains.

\noindent\fbox{%
    \parbox{\linewidth}{%
\textbf{Conclusion (Targeted Source Control).}
When objective conflicts are present, inference-time interventions exhibit a clear \textbf{directional asymmetry}: biasing the model toward fact-consistent, truth-anchored sources is significantly easier and more reliable than forcing it to rely on fact-inconsistent sources. 
This suggests that conflict resolution in MLLMs is governed by a stable, source-dependent inductive tendency, which can be strengthened but is difficult to reverse.
    }%
}

\subsection{Conflict mitigation under the default direction}
\label{sec:5.3}

Semantic evaluation in Section~\ref{sec:5.2} demonstrated that, under objectively conflicting inputs, inference-time interventions can bias model outputs toward the truth-anchored source. Here, we pose a complementary process-level question: \emph{under the default (Forward) direction, can we reduce the activation of \textit{effective conflicts} $\mathcal{C}^{e}_{i,j}(t\mid x)$ during generation?}
We employ token-level mitigation metrics to summarize these internal dynamics (Support Score (SS), Conflict Rate (CR), Confidence-Adjusted Conflict (CAC), and Conflict Confidence Index (CCI)) as a further complement to the independent semantic correctness evaluation in Figure~\ref{fig:sec5_semantic}.

Table~\ref{tab:mitigation} summarizes the token-level mitigation results.
We observe that interventions targeting the identified conflict features (\textbf{Probe-guided control}) consistently suppress conflict dynamics across backbones.
Specifically, on visually involved subsets ($\mathcal{C}_{\mathrm{VT}}$), the frequency of conflict activation (CR) decreases significantly (e.g., 0.10$\to$0.02 on \llmname{R1-Onevision}).
Crucially, even when conflict frequency remains stable (e.g., $\mathcal{C}_{\mathrm{PT}}$), confidence-aware measures reveal deeper suppression (CCI drops by 25\%), indicating that the intervention mitigates the \emph{intensity} of conflicts even if not their occurrence.
In contrast, rigid interventions like \textbf{Representation steering} or unguided perturbations like \textbf{VCD} struggle to generalize.
For instance, \textbf{VCD} exacerbated conflict rates fivefold on \llmname{Llama-3.2V} for $\mathcal{C}_{\mathrm{VT}}$ (0.03$\to$0.15).
This disparity highlights that effective mitigation requires precise targeting of the conflict-encoding subspaces rather than broad adjustments.

\noindent\fbox{%
    \parbox{\linewidth}{%
\textbf{Conclusion (Conflict Mitigation).}
Guiding the model toward the reliable source attenuates internal conflict dynamics during reasoning, reducing both the intensity and the frequency of \textit{effective conflict} states.
This implies that \textit{effective conflict} $\mathcal{C}^e_{i,j}(t \mid x)$ activation is not an inherent attribute of generation, but a plastic internal state that can be suppressed during reasoning.
    }%
}

%% file: secs/mitigation_tab.tex
\begin{table*}[t]
\centering
\small
\renewcommand{\arraystretch}{1.0}
\setlength{\tabcolsep}{5pt}
\caption{\textbf{Token-level conflict mitigation under the forward direction.}
Results are reported on three objective-conflict subsets
($\mathcal{C}^{o}_{\mathrm{VP}}=1$, $\mathcal{C}^{o}_{\mathrm{VT}}=1$, and $\mathcal{C}^{o}_{\mathrm{PT}}=1$)
across three backbones.
We report four token-level mitigation metrics:
\textbf{SS}$\uparrow$,
\textbf{CR}$\downarrow$,
\textbf{CAC}$\downarrow$ and \textbf{CCI}$\downarrow$ (metric definitions in Appendix~\ref{app:probe_metrics}).
VCD is not applicable when $\mathcal{C}^{o}_{\mathrm{PT}}=1$ and is therefore reported only for the first two subsets.}
\vspace{-0.7em}
\resizebox{\textwidth}{!}{
\begin{tabular}{@{} l c W cccc V cccc V cccc @{}}
\toprule
\multirow{2}{*}{\rotatebox{90}{\textbf{}}} & \multirow{2}{*}{\textbf{Subset}} &
\multicolumn{4}{c}{\textbf{R1-Onevision-7B}} &
\multicolumn{4}{c}{\textbf{Ocean-R1-7B-Instruct}} &
\multicolumn{4}{c}{\textbf{Llama-3.2V-11B-cot}} \\
\cmidrule(lr){3-6}\cmidrule(lr){7-10}\cmidrule(lr){11-14}
& &
\textbf{SS$\uparrow$} & \textbf{CAC$\downarrow$} & \textbf{CCI$\downarrow$} & \textbf{CR$\downarrow$} &
\textbf{SS$\uparrow$} & \textbf{CAC$\downarrow$} & \textbf{CCI$\downarrow$} & \textbf{CR$\downarrow$} &
\textbf{SS$\uparrow$} & \textbf{CAC$\downarrow$} & \textbf{CCI$\downarrow$} & \textbf{CR$\downarrow$} \\
\midrule

\rowcolor{rowbase}
 & $\mathcal{C}^{o}_{\mathrm{VP}}=1$
 & 0.94 & 0.04 & 0.70 & 0.03
 & 0.89 & 0.07 & 0.71 & 0.06
 & 0.94 & 0.04 & 0.45 & 0.02 \\
\rowcolor{rowbase}
 & $\mathcal{C}^{o}_{\mathrm{VT}}=1$
 & 0.88 & 0.08 & 0.80 & 0.10
 & 0.87 & 0.09 & 0.79 & 0.10
 & 0.90 & 0.06 & 0.72 & 0.03 \\
\rowcolor{rowbase}
\multirow{-3}{*}{\rotatebox{90}{\textbf{baseline}}} & $\mathcal{C}^{o}_{\mathrm{PT}}=1$
 & 0.82 & 0.12 & 0.80 & 0.15
 & 0.82 & 0.12 & 0.80 & 0.15
 & 0.84 & 0.11 & 0.70 & 0.11 \\
\midrule

\multirow{2}{*}{\rotatebox{90}{\textbf{VCD}}} & $\mathcal{C}^{o}_{\mathrm{VP}}=1$
 & 0.92\dneg{-0.02} & 0.05\dneg{+0.01} & 0.69\dpos{-0.00} & 0.04\dneg{+0.01}
 & 0.90\dpos{+0.01} & 0.06\dpos{-0.01} & 0.69\dpos{-0.01} & 0.05\dpos{-0.01}
 & 0.85\dneg{-0.09} & 0.08\dneg{+0.04} & 0.63\dneg{+0.18} & 0.06\dneg{+0.05} \\
& $\mathcal{C}^{o}_{\mathrm{VT}}=1$
 & 0.90\dpos{+0.01} & 0.07\dpos{-0.01} & 0.79\dpos{-0.01} & 0.08\dpos{-0.01}
 & 0.92\dpos{+0.05} & 0.05\dpos{-0.03} & 0.75\dpos{-0.04} & 0.05\dpos{-0.05}
 & 0.78\dneg{-0.12} & 0.12\dneg{+0.06} & 0.69\dpos{-0.03} & 0.15\dneg{+0.11} \\

\midrule

\multirow{3}{*}{\rotatebox{90}{\textbf{steering}}} & $\mathcal{C}^{o}_{\mathrm{VP}}=1$
 & 0.92\dneg{-0.02} & 0.05\dneg{+0.01} & 0.69\dpos{-0.00} & 0.05\dneg{+0.02}
 & 0.89\dneg{-0.01} & 0.07\dneg{+0.00} & 0.71\dneg{+0.00} & 0.07\dneg{+0.01}
 & 0.92\dneg{-0.02} & 0.05\dneg{+0.01} & 0.55\dneg{+0.10} & 0.03\dneg{+0.02} \\
& $\mathcal{C}^{o}_{\mathrm{VT}}=1$
 & 0.91\dpos{+0.02} & 0.06\dpos{-0.02} & 0.76\dpos{-0.03} & 0.07\dpos{-0.03}
 & 0.91\dpos{+0.04} & 0.06\dpos{-0.03} & 0.77\dpos{-0.03} & 0.06\dpos{-0.04}
 & 0.90\dpos{+0.00} & 0.06\dpos{-0.00} & 0.67\dpos{-0.04} & 0.04\dneg{+0.01} \\
& $\mathcal{C}^{o}_{\mathrm{PT}}=1$
 & 0.77\dneg{-0.06} & 0.16\dneg{+0.04} & 0.79\dpos{-0.01} & 0.20\dneg{+0.05}
 & 0.82\dpos{+0.00} & 0.12\dpos{-0.00} & 0.80\dpos{-0.00} & 0.15\dneg{+0.00}
 & 0.84\dpos{+0.00} & 0.11\dpos{-0.00} & 0.69\dpos{-0.01} & 0.12\dneg{+0.01} \\
\midrule

\multirow{3}{*}{\rotatebox{90}{\textbf{probe}}} & $\mathcal{C}^{o}_{\mathrm{VP}}=1$
 & 0.95\dpos{+0.01} & 0.03\dpos{-0.01} & 0.67\dpos{-0.03} & 0.02\dpos{-0.01}
 & 0.92\dpos{+0.03} & 0.05\dpos{-0.02} & 0.66\dpos{-0.04} & 0.03\dpos{-0.03}
 & 0.94\dpos{+0.01} & 0.04\dpos{-0.00} & 0.39\dpos{-0.06} & 0.02\dpos{-0.00} \\
& $\mathcal{C}^{o}_{\mathrm{VT}}=1$
 & 0.94\dpos{+0.06} & 0.04\dpos{-0.04} & 0.64\dpos{-0.16} & 0.02\dpos{-0.07}
 & 0.93\dpos{+0.06} & 0.04\dpos{-0.04} & 0.72\dpos{-0.07} & 0.04\dpos{-0.06}
 & 0.92\dpos{+0.02} & 0.05\dpos{-0.01} & 0.67\dpos{-0.05} & 0.03\dpos{-0.00} \\
& $\mathcal{C}^{o}_{\mathrm{PT}}=1$
 & 0.78\dneg{-0.04} & 0.10\dpos{-0.02} & 0.60\dpos{-0.20} & 0.15\dneg{+0.01}
 & 0.87\dpos{+0.04} & 0.08\dpos{-0.04} & 0.72\dpos{-0.08} & 0.09\dpos{-0.06}
 & 0.87\dpos{+0.04} & 0.08\dpos{-0.03} & 0.63\dpos{-0.07} & 0.10\dpos{-0.01} \\
\bottomrule
\end{tabular}
}

\vspace{-1.0em}

\label{tab:mitigation}

\end{table*}

%% file: secs/6.conclusion.tex
\vspace{-0.5em}
\section{Conclusion}
\label{sec:6_conclusion}
\vspace{-0.2em}

In this work, we study failures in multimodal long-CoT reasoning from the perspective of knowledge conflict, rather than knowledge absence.
By distinguishing \emph{objective conflicts} from \emph{effective conflicts} during reasoning, we show that many failures arise from how conflicting knowledge is resolved over time.
We find that \emph{effective conflicts} are encoded as explicit and linearly decodable signals, concentrated in mid-to-late layers of the model.
Leveraging these signals, we uncover a pronounced directional asymmetry: guiding the model toward its reliability-aligned source is substantially easier than forcing conflict resolution in the opposite direction, indicating a biased and path-dependent mechanism.
Looking forward, we hope this perspective motivates analysis and control methods for richer conflict structures and more complex multimodal reasoning settings.


%% file: secs/impact_statement.tex
\section*{Impact Statement}
This paper presents work whose goal is to advance the understanding and reliability of MLLMs in long-CoT reasoning scenarios. By diagnosing knowledge conflicts and their intervention mechanisms, our research contributes to making AI systems more transparent and trustworthy. The diagnostic framework and intervention methods proposed here could help identify and mitigate reasoning failures before deployment, potentially reducing the propagation of misinformation or hallucinated content in real-world applications. We do not foresee specific negative societal consequences that need to be highlighted beyond the general considerations applicable to advancing machine learning capabilities.

%% file: secs/appendix.tex
\appendix

\section{Dataset Construction and Annotation}\label{app:construction}

\subsection{Dataset Overview}
\label{app:dataset_overview}

We construct the \textbf{Objective Conflict Dataset}, a fine-grained benchmark designed to systematically evaluate knowledge conflict phenomena in MLLMs during long-CoT reasoning. Drawing upon the three-source knowledge framework established in Section~\ref{sec:3.1}, we leverage state-of-the-art closed-source models (e.g., \llmname{GPT-5}, \llmname{Claude-Sonnet-4.5}) to facilitate targeted instruction expansion, contradictory context generation, and span-level annotation. We emphasize that this dataset is a controlled stress-test benchmark designed for mechanistic analysis, rather than estimating conflict prevalence in natural deployments.
\vspace{-1.0em}
\paragraph{Role of Closed-Source Models and Truth Anchors.}
It is important to clarify that closed-source models serve exclusively as \emph{construction and annotation tools}, rather than ground-truth generators. The truth anchors for conflict verification are derived from independent, objective sources:
\begin{itemize}[leftmargin=*, nosep]
    \item \textbf{Visual evidence} ($\mathcal{K}_{\text{vision}}$): The image $X_V$ itself, which provides objectively verifiable visual facts independent of any model.
    \item \textbf{Textual input} ($\mathcal{K}_{\text{text}}$): The given prompt text $X_T$, which is explicitly provided and thus unambiguous.
    \item \textbf{Factual knowledge} ($\mathcal{K}_{\text{prior}}$): Human-annotated correct answers from established benchmarks (e.g., TruthfulQA~\citep{lin2022truthfulqa}), ensuring that world knowledge verification relies on externally validated facts rather than model-generated content.
\end{itemize}
This design ensures that any potential biases in closed-source models affect only the \emph{style} of generated queries and annotations, not the \emph{correctness} of truth anchors. Furthermore, as we demonstrate in Section~\ref{app:cross_model_consistency}, our core findings remain consistent across three architecturally diverse backbones, suggesting that observed phenomena reflect intrinsic model properties rather than annotation artifacts.

Our construction ensures that each sample contains a dominant pairwise objective conflict $\mathcal{C}^o_{i,j}(x)=1$ (where $\mathcal{C}_{i,j} \in \{\mathcal{C}_{\mathrm{VP}}, \mathcal{C}_{\mathrm{VT}}, \mathcal{C}_{\mathrm{PT}}\}$), while strictly minimizing cross-type contamination. The dataset covers the three core conflict types: 

\begin{itemize}[leftmargin=*, nosep]
    \item \textbf{Vision-Text Conflict ($\mathcal{C}^o_{\mathrm{VT}}(x)=1$):} The objective visual evidence $X_V$ contradicts the textual description $X_T$ provided in the prompt.
    \item \textbf{Vision-Prior Conflict ($\mathcal{C}^o_{\mathrm{VP}}(x)=1$):} Counter-intuitive or synthetic visual inputs $X_V$ contradict the parametric prior knowledge implicitly encoded in the model parameters $\theta$.
    \item \textbf{Prior-Text Conflict ($\mathcal{C}^o_{\mathrm{PT}}(x)=1$):} Misleading textual instructions $X_T$ contradict the parametric prior knowledge implicitly encoded in the model parameters $\theta$.
\end{itemize}

This appendix details the pipeline used to construct these inputs and the subsequent annotation process used to identify effective conflict triggers $\mathcal{C}^e_{i,j}(t\mid x)$ during reasoning.

\subsection{Data Sources and Selection}

To ensure diversity and complexity, we curate samples from the \textbf{PhD~\cite{liu2024phd}} and \textbf{TruthfulQA~\cite{lin2022truthfulqa}} benchmarks, adapting them to enforce specific objective conflict profiles.
\vspace{-1.0em}
\paragraph{Adaptation of PhD Benchmark.}
The PhD dataset, originally designed to evaluate visual faithfulness in MLLMs, spans tasks from object recognition to complex scene analysis. Two specific subsets are selected to construct visual-related conflicts.
\textit{\textbf{Vision-Text Conflict.}} The \textbf{PhD-icc} subset, containing 14,648 natural images paired with erroneous descriptions, is leveraged for this conflict type. Approximately 1,000 images are sampled from this collection to serve as ground-truth visual evidence $X_V$. We discard the original answers and contexts, instead employing advanced closed-source MLLMs to synthesize highly misleading textual contexts $X_T$ and complex queries designed to cause conflict during reasoning. Then, we generate responses using the three open-source MLLMs under investigation to obtain their reasoning outputs under conflict.
\textit{\textbf{Vision-Prior Conflict.}} This study utilizes the \textbf{PhD-ccs} subset, which comprises 750 AI-generated images that deliberately violate physical laws or semantic common sense (e.g., a boat sailing on desert sand). In these scenarios, the visual evidence $X_V$ inherently contradicts the model’s internal world knowledge $\mathcal{K}_{\text{prior}}$. We retain the core images and questions but discard the original answers. Each model performs independent inference, allowing us to analyze how they resolve the tension between perceptual evidence and parametric priors.

\vspace{-1.0em}
\paragraph{Adaptation of TruthfulQA Benchmark.}

For Prior-Text Conflict, we employ the \textbf{TruthfulQA} benchmark to construct conflicts where visual information is irrelevant or neutral. TruthfulQA contains 817 questions across 38 categories (e.g., health, law, fiction) designed to elicit imitative falsehoods. These questions target scenarios where the model's pre-training data might contain misconceptions. We utilize the adversarial ``Best Incorrect Answers'' provided by TruthfulQA to construct misleading textual contexts $X_T$ that reinforce false beliefs, thereby creating a strong objective conflict against the factual truth.

\subsection{Deceptive Context and Query Generation}

To maximize the probability of triggering effective conflict $\mathcal{C}^e_{i,j}(t\mid x)$ during the reasoning process, we developed specialized strategies to synthesize queries. These queries are carefully synthesized to create high-intensity yet controlled contradictions, thereby reliably eliciting conflict-driven reasoning trajectories for analysis. The design principles for each conflict type are detailed below:

For $\mathcal{C}_{\mathrm{VT}}$, we employ a Dual-Stage Prompting approach to synthesize high-intensity conflicts:

\begin{itemize}[leftmargin=*, nosep]
    \item \textbf{Step 1: }Fact Extraction. We first extract objective visual facts from the image $X_V$.
    \item \textbf{Step 2:} Narrative Synthesis. We utilize advanced closed-source LLMs to synthesize a coherent but factually contradictory narrative $X_T$, constructed to contradict key visual elements while maintaining superficial plausibility.
    \item \textbf{Step 3:} Question Formulation. We generate open-ended, non-binary questions that necessitate multi-step reasoning by fusing information from both modalities.
\end{itemize}
This strategy is designed to force the model into a conflict state, where it must explicitly choose to either generate a text-induced conflict or remain faithful to the visual evidence. The detailed prompt design principles are described in Section~\ref{app:prompt_design}.

For $\mathcal{C}_{\mathrm{VP}}$, we append standardized constraints to the original queries to elicit longer reasoning trajectories. Specifically, we mandate the models to follow a structured four-step reasoning process, moving from an initial visual description and context analysis to an explicit evidence chain reasoning before reaching a final conclusion.
By enforcing this structured Chain-of-Thought (CoT), we prevent the model from bypassing the conflict through brief or evasive answers. Instead, it compels the explicit manifestation of the clash between counter-intuitive visual evidence and internal priors within the generated text.

For $\mathcal{C}_{\mathrm{PT}}$, we employ a structured template to embed the complete set of ``Incorrect Answers'' from the TruthfulQA benchmark directly into the query. We present the original question followed by a deceptive instruction: \textit{``Is it because of the following reasons? If not, please answer based on your knowledge.''} 
This design creates a direct conflict between the provided incorrect context $X_T$ and the model's parametric priors $\mathcal{K}_{\text{prior}}$. By explicitly listing multiple false justifications, we exert significant pressure on the model to conform to the provided misinformation. We then evaluate whether the model yields to these distractors or successfully retrieves the ground-truth answer from its internal knowledge.

\subsection{Response Generation and Span-Level Annotation}

To systematically observe how input level objective conflicts $\mathcal{C}^o_{i,j}(x)$ manifest as process-level effective conflicts $\mathcal{C}^e_{i,j}(t\mid x)$ during inference, we curate a targeted corpus of reasoning outputs generated by three representative MLLMs: \llmname{R1-Onevision-7B}, \llmname{Ocean-R1-7B-Instruct}, and \llmname{Llama-3.2V-11B-cot}. These outputs encompass a diverse range of model behaviors under conflicting inputs, providing the necessary empirical foundation for dissecting internal conflict dynamics.

\vspace{-1.0em}
\paragraph{Generation Settings.} We perform inference using the vLLM framework with asynchronous batching. Key hyperparameters include: sampling temperature $T_{temp}=0.7$ and maximum sequence length of 4,096 tokens, facilitating extended Chain-of-Thought (CoT) generation. For the Vision-Prior subset specifically, we employ a structured four-step reasoning process guided by the prompt design principles detailed in Section~\ref{app:prompt_design}.

\vspace{-1.0em}
\paragraph{Annotation Procedure.} 
We implement an automated pipeline utilizing \llmname{GPT-5} as a central judge. Adopting the \textit{Extract-Verify-Score} paradigm, the process involves: (1) \textbf{Extraction} of model outputs into atomic, verifiable fact segments (\textit{spans}). (2) \textbf{Verification} of each span against respective truth anchors ($\mathcal{K}$) and visual evidence $X_V$. and (3) \textbf{Scoring} by assigning conflict labels (e.g., \textit{Supported}, $\mathcal{C}^e_{\mathrm{VP}}$, $\mathcal{C}^e_{\mathrm{VT}}$, or $\mathcal{C}^e_{\mathrm{PT}}$) based on observed conflict-induced behaviors. Notably, to ensure precise localization, we employ a fuzzy string matching algorithm to align annotated spans back to their exact character-level indices within the original model output. This step corrects for minor textual rephrasing by the judge model, providing the granular supervision necessary for training our streaming conflict probes. The specific prompt used for annotation is shown in the box below.

\subsection{Quality Control and Dataset Statistics}

\paragraph{Human Verification.}
To ensure the reliability of the automated annotation process, we conduct human verification on a stratified random 10\% subset of annotations (approximately 15,000 spans). 
Three annotators with graduate-level NLP background independently verified: (1) \textbf{span boundary accuracy}---whether the extracted span correctly delimits the atomic claim. (2) \textbf{conflict label correctness}---whether the assigned label ($\mathcal{C}_{\mathrm{VP}}$, $\mathcal{C}_{\mathrm{VT}}$, $\mathcal{C}_{\mathrm{PT}}$, or \textit{Supported}) accurately reflects the conflict type.
Annotators achieved an inter-annotator agreement of $\kappa = 0.87$ (Cohen's kappa) for conflict labels and $\kappa = 0.91$ for span boundaries.
Disagreements were resolved through majority voting, with remaining ties adjudicated by a senior annotator.

\begin{table*}[htp]
\centering
\small
\renewcommand{\arraystretch}{1.25}
\setlength{\tabcolsep}{16pt}
\caption{Statistics of the Objective Conflict Dataset and Generated Trajectories. The dataset provides span-level supervision for training conflict probes (Section~\ref{sec:4_probe}) and evaluating interventions (Section~\ref{sec:5_intervention}).}
\label{tab:dataset_stats}
\begin{tabular}{@{}l c r r r@{}}
\toprule
\textbf{Model Backbone} & \textbf{Split} & \textbf{Samples} & \textbf{Total Spans} & \textbf{Avg. Spans/Sample} \\
\midrule
\multirow{2}{*}{\textbf{R1-Onevision-7B}} & Test & 498 & 9,038 & 18.1 \\
 & Train & 1,987 & 34,300 & 17.3 \\
\midrule
\multirow{2}{*}{\textbf{Ocean-R1-7B}} & Test & 495 & 12,881 & 26.0 \\
 & Train & 1,978 & 51,931 & 26.3 \\
\midrule
\multirow{2}{*}{\textbf{Llama-3.2V-11B-cot}} & Test & 513 & 9,419 & 18.4 \\
 & Train & 2,051 & 37,118 & 18.1 \\
\midrule
\rowcolor{graybg}
\textbf{Total} & -- & \textbf{7,522} & \textbf{154,687} & \textbf{20.6} \\
\bottomrule
\end{tabular}
\end{table*}

The most common source of disagreement involved distinguishing $\mathcal{C}_{\mathrm{VP}}$ from $\mathcal{C}_{\mathrm{VT}}$ when both visual misinterpretation and textual influence were present. In such cases, annotators defaulted to the dominant conflict source based on explicit textual cues in the model's reasoning chain.
\vspace{-1.0em}
\paragraph{Dataset Organization.}
The finalized dataset is organized into three model-specific subsets corresponding to \llmname{R1-Onevision-7B}, \llmname{Ocean-R1-7B-Instruct}, and \llmname{Llama-3.2V-11B-cot}. Within each subset, we apply stratified sampling to partition samples into Train (80\%) and Test (20\%) splits, ensuring balanced representation of the three conflict types ($\mathcal{C}_{\mathrm{VP}}$, $\mathcal{C}_{\mathrm{VT}}$, $\mathcal{C}_{\mathrm{PT}}$) while minimizing distribution shift.

\paragraph{Final Statistics.}
The complete dataset comprises approximately 155,000 fine-grained annotated spans derived from model-specific reasoning trajectories. Detailed statistics regarding sample counts and span densities are provided in Table~\ref{tab:dataset_stats}.

\subsection{Prompt Design for Dataset Construction}
\label{app:prompt_design}

This section details the prompt design principles used for dataset construction. We describe the key components of each prompt type and provide the complete annotation prompt used for span-level labeling.
\vspace{-1.0em}
\paragraph{Vision-Text Conflict Question Generation.}
\label{para:vt_prompt}
To construct high-quality $\mathcal{C}_{\mathrm{VT}}$ samples, we design a structured prompt that guides closed-source models through a five-stage process:

\begin{enumerate}[leftmargin=*, nosep]
    \item \textbf{Visual Fact Extraction}: The model first extracts 5--10 objective, verifiable visual facts from the image (e.g., number of people, visible objects, scene type, actions). These facts serve as ground-truth anchors for subsequent conflict verification.
    
    \item \textbf{Conflicting Context Synthesis}: Based on extracted facts, the model generates a coherent but systematically misleading textual description (80--150 words). The key design principle is to create \emph{plausible deception}: the text incorporates partially correct details (colors, approximate positions) while introducing at least three factual mismatches. For example, mislabeling leisure activities as emergencies while preserving neutral scene elements, or distorting object roles while maintaining spatial layout.
    
    \item \textbf{Question Formulation}: The model designs open-ended, non-binary questions that necessitate multi-step reasoning by fusing both modalities. Questions are crafted to: (a) require at least 3 reasoning steps; (b) avoid explicitly indicating that the text is misleading; (c) encourage speculation about ambiguous image regions to amplify conflict probability.
    
    \item \textbf{Reference Answer Generation}: An ideal, conflict-free answer is generated that relies solely on visual evidence, explicitly references extracted facts, and presents a clear multi-step reasoning chain.
    
    \item \textbf{Conflict Tagging}: Each sample is tagged with the conflict type (\texttt{image\_vs\_text\_conflict}) for downstream processing.
\end{enumerate}

The output format is a structured JSON containing the conflicting context, question, and reference answer, enabling automated downstream processing.
\vspace{-1.0em}
\paragraph{Vision-Prior Conflict Reasoning Induction.}
\label{para:vp_prompt}
For $\mathcal{C}_{\mathrm{VP}}$ samples, we employ a structured Chain-of-Thought induction prompt that mandates a four-step reasoning process:

\begin{enumerate}[leftmargin=*, nosep]
    \item \textbf{Visual Description}: Objectively describe visible elements without applying prior knowledge.
    \item \textbf{Context Analysis}: Identify environmental settings and object relationships.
    \item \textbf{Evidence Chain Reasoning}: Evaluate whether visual evidence aligns with physical laws or common sense, if conflict exists, explicitly articulate the tension.
    \item \textbf{Conclusion}: Provide a final answer based on the preceding analysis.
\end{enumerate}

This structured format prevents models from bypassing the conflict through brief or evasive answers, instead compelling explicit manifestation of the clash between counter-intuitive visual evidence and internalized priors.
\vspace{-1.0em}
\paragraph{Prior-Text Conflict Template.}
For $\mathcal{C}_{\mathrm{PT}}$ samples, we embed ``Best Incorrect Answers'' from TruthfulQA directly into the query using a structured template: \textit{``[Original Question]. Is it because of the following reasons? [List of incorrect answers]. If not, please answer based on your knowledge.''} This design creates direct pressure on the model to either conform to provided misinformation or retrieve factual knowledge from its parameters.
\vspace{-1.0em}
\paragraph{Span-Level Annotation Prompt.}
The annotation prompt implements our \textit{Extract-Verify-Score} paradigm and is the most critical component for obtaining fine-grained supervision. The complete prompt is provided below.  

\begin{tcolorbox}[
    breakable,
    enhanced,
    colback=EarthSand!15!white,
    colframe=EarthTerracotta,
    coltitle=white,
    colbacktitle=EarthTerracotta,
    title={\textbf{\faRobot~Knowledge Conflict Annotation Prompt}},
    fonttitle=\bfseries\sffamily,
    boxrule=1pt,
    arc=2mm,
    left=3mm, right=3mm, top=2mm, bottom=2mm,
    pad at break=3mm,
    attach boxed title to top left={xshift=2.5mm, yshift=-2mm},
    boxed title style={
        colback=EarthTerracotta,
        arc=1mm,
        boxrule=0pt,
    }
]
\small

\textit{You are a multimodal fact-checking expert and conflict annotation engine.}

\medskip
Your task is to \textbf{EXHAUSTIVELY} fact-check a Vision-Language Model's (VLM) completion against the given instruction, ground truth, and image, with a \textbf{PRIMARY GOAL} of \textbf{MAXIMUM SPAN COVERAGE} (high recall).

\tcbsubtitle[colback=EarthSand!70!white, colframe=EarthTerracotta, coltitle=EarthTerracotta!30!black]{\textbf{INPUTS}}
\begin{lstlisting}[basicstyle=\ttfamily\footnotesize, columns=fullflexible, backgroundcolor=\color{gray!8}]
Instruction:    <instruction>{instruction}</instruction>
Ground Truth:   <groundtruth>{groundtruth}</groundtruth>
Model Completion: <completion>{completion}</completion>
\end{lstlisting}

\tcbsubtitle[colback=EarthSand!70!white, colframe=EarthTerracotta, coltitle=EarthTeal!30!black]{\textbf{CRITICAL TASK DEFINITION}}
You must identify and verify \textbf{ALL} minimal, self-contained, verifiable claims appearing in the completion.

A claim is \textbf{ANY} text span that asserts, implies, or assumes something that can be judged as true or false, including:
\begin{itemize}[leftmargin=1.5em, itemsep=1pt, topsep=2pt, nosep]
    \item existence of an object or entity
    \item object attributes (color, size, shape, material, state)
    \item quantities or counts
    \item spatial relations (left/right/on/in/near)
    \item actions or behaviors
    \item scene-level assumptions
    \item factual statements
    \item implicit assumptions derived from prior knowledge
\end{itemize}

\colorbox{yellow!30}{\textbf{IMPORTANT:}} Claims are \textbf{NOT} limited to named entities. Descriptive words, adjectives, numbers, directions, and verbs \textbf{OFTEN} contain conflicts. If a phrase can be checked, it \textbf{MUST} be extracted.

\tcbsubtitle[colback=EarthSand!70!white, colframe=EarthTerracotta, coltitle=EarthTerracotta!30!black]{\textbf{MANDATORY CLAIM DECOMPOSITION STEP}}
Before verification, you \textbf{MUST} mentally decompose the completion sentence by sentence. For \textbf{EACH} sentence:
\begin{itemize}[leftmargin=1.5em, itemsep=1pt, topsep=2pt]
    \item Identify \textbf{ALL} independent claims it contains.
    \item A single sentence usually contains \textbf{MULTIPLE} claims.
    \item Do \textbf{NOT} stop after extracting a few obvious entities.
\end{itemize}

\tcbsubtitle[colback=EarthSand!70!white, colframe=EarthTerracotta, coltitle=EarthTeal!30!black]{\textbf{WHAT TO VERIFY}}
You need to verify two types of content:

\textbf{1. Entity-Level Information:} People, organizations, locations, dates, statistics. Named or implied entities. Claims about multimodal content (e.g., ``the building on the left'').

\textbf{2. Visual Description Information:} Object existence; Object attributes; Spatial relationships; Quantities; Actions and behaviors; Scene understanding.

\tcbsubtitle[colback=EarthSand!70!white, colframe=EarthTerracotta, coltitle=EarthTerracotta!30!black]{\textbf{SPAN EXTRACTION RULES (VERY STRICT)}}
For \textbf{EACH} verifiable claim:
\begin{enumerate}[leftmargin=1.5em, itemsep=1pt, topsep=2pt, nosep]
    \item Extract the \textbf{MINIMAL} possible span that directly expresses that claim.
    \item The span \textbf{MUST} match the original text word-for-word.
    \item Do \textbf{NOT} include surrounding context, articles, or extra description.
    \item Do \textbf{NOT} merge multiple claims into one span.
    \item One sentence may (and usually should) produce \textbf{MULTIPLE} spans.
\end{enumerate}

\textbf{Granularity rule:} Prefer word-level or short phrase-level spans. If a shorter sub-phrase can carry the claim, use it instead of a longer one.

\textbf{Recall rule:} When uncertain whether something is a claim, \textbf{EXTRACT IT}. Over-extraction is strongly preferred to under-extraction.

\tcbsubtitle[colback=EarthSand!70!white, colframe=EarthTerracotta, coltitle=EarthTerracotta!30!black]{\textbf{VERIFICATION PRIORITY}}
\begin{enumerate}[leftmargin=1.5em, itemsep=1pt, topsep=2pt, nosep]
    \item Image content (\textbf{highest priority})
    \item Instruction text
    \item Established factual or commonsense knowledge
\end{enumerate}
\colorbox{yellow!30}{\textbf{If there is a conflict:}} Image content always prevails over text or prior knowledge.

\tcbsubtitle[colback=EarthSand!70!white, colframe=EarthTerracotta, coltitle=EarthTeal!30!black]{\textbf{CONFLICT CLASSIFICATION}}
Label \textbf{EACH} extracted span as exactly \textbf{ONE} of:
\begin{itemize}[leftmargin=1.5em, itemsep=2pt, topsep=2pt, nosep]
    \item \colorbox{FactGreen}{\textbf{``Supported''}} -- The span is correct and consistent with the image, instruction, and factual knowledge.
    \item \colorbox{VPCRed}{\textbf{``Vision-Prior Conflict''}} -- The span contradicts the image due to reliance on learned priors.
    \item \colorbox{PTCGrey}{\textbf{``Prior-Text Conflict''}} -- The span follows or accepts incorrect assumptions in the instruction that contradict factual knowledge.
    \item \colorbox{VTCYellow}{\textbf{``Vision-Text Conflict''}} -- The span reflects a mismatch between the image and the instruction.
\end{itemize}

\tcbsubtitle[colback=EarthSand!70!white, colframe=EarthTerracotta, coltitle=EarthTerracotta!30!black]{\textbf{QUALITY CONTROL (MANDATORY SELF-CHECK)}}
Before producing the final output, check:
\begin{enumerate}[leftmargin=1.5em, itemsep=1pt, topsep=2pt, nosep]
    \item Did you extract spans for: objects? attributes? quantities? spatial relations? actions?
    \item Is the number of spans proportional to the length and detail of the completion?
    \begin{itemize}[leftmargin=1em, itemsep=0pt]
        \item A detailed paragraph should yield \textbf{MANY} spans.
        \item Very few spans usually indicate missed conflicts.
    \end{itemize}
\end{enumerate}

\tcbsubtitle[colback=EarthTeal!30!white, colframe=EarthTeal, coltitle=EarthTeal!30!black]{\textbf{OUTPUT FORMAT (STRICT)}}
Return \textbf{ONLY} a JSON array. Do \textbf{NOT} include explanations outside the JSON. The spans \textbf{MUST} be ordered by their first appearance in the completion.

\begin{lstlisting}[basicstyle=\ttfamily\footnotesize, columns=fullflexible, backgroundcolor=\color{gray!8}, frame=single, rulecolor=\color{gray!50}]
[
  {
    "span": "exact minimal span from the completion",
    "label": "Supported | Vision-Prior Conflict | 
             Prior-Text Conflict | Vision-Text Conflict"
  },
  ...
]
\end{lstlisting}

\end{tcolorbox}

\vspace{-0.5em}
\refstepcounter{figure}
\textbf{Box: Knowledge Conflict Annotation Prompt.} This prompt directs the judge model to decompose VLM responses into atomic spans and assign conflict labels ($\mathcal{C}^e_{\mathrm{VP}}$, $\mathcal{C}^e_{\mathrm{VT}}$, $\mathcal{C}^e_{\mathrm{PT}}$), providing granular token-level supervision for conflict probe training.
\label{box:prompts_for_annotation}


\section{Implementation Details (Backbones and Decoding)}
\label{app:implementation}

We evaluate three state-of-the-art reasoning-capable MLLMs with diverse architectures and training paradigms. Table~\ref{tab:backbone_summary} summarizes their key characteristics.

\begin{table}[ht]
\centering
\small
\renewcommand{\arraystretch}{1.2}
\setlength{\tabcolsep}{2.5pt}
\caption{Summary of evaluated MLLM backbones. The ``Conflict Layer'' indicates the layer used for probe training and intervention (identified in Section~\ref{sec:4.3}).}
\label{tab:backbone_summary}
\vspace{-0.5em}
\begin{tabular}{@{}l c c c@{}}
\toprule
\textbf{Attribute} & \textbf{R1-Onevision} & \textbf{Ocean-R1} & \textbf{Llama-3.2V-cot} \\
\midrule
\textbf{Params} & 7B & 7B & 11B \\
\textbf{Layers} & 28 & 28 & 40 \\
\textbf{Hidden Dim} & 3584 & 3584 & 4096 \\
\textbf{Conflict Layer} & 20 & 20 & 39 \\
\textbf{Base Model} & Qwen2.5-VL & Qwen2.5-VL & Llama-3.2-Vision \\
\bottomrule
\end{tabular}
\vspace{-1.0em}
\end{table}

\noindent\textbf{Ocean-R1~\citep{ming2025oceanr1}.}
\llmname{Ocean\--R1} is a vision-language reasoning model fine-tuned from \llmname{Qwen2.5\--VL\--Instruct}~\citep{bai2025qwen25vl}, optimized via a two-stage rule-based Reinforcement Learning (RL) framework. The first stage enhances reasoning capabilities, while the second refines visual perception. Empirical results on benchmarks such as MathVision and MathVista demonstrate its superior multimodal reasoning and generalization performance.

\noindent\textbf{Llama-3.2V-11B-cot~\citep{Xu_2025_ICCV}.}
\llmname{Llama\--3.2V\--11B\--cot} is a reasoning-intensive model built on the \llmname{Llama-3.2-Vision-11B} architecture. Adopting the LLaVA-CoT methodology~\citep{Xu_2025_ICCV}, it is fine-tuned on visual reasoning chains to perform step-by-step decomposition of complex queries. This Chain-of-Thought (CoT) mechanism enables precise handling of multi-hop reasoning and intricate scene understanding, dealing effectively with challenges where standard instruction-tuned models may falter.

\noindent\textbf{R1-Onevision~\citep{yang2025r1onevision}.}
\llmname{R1-Onevision} bridges visual perception and deep reasoning by fine-tuning \llmname{Qwen2.5-VL}~\citep{bai2025qwen25vl} through a two-stage strategy combining Supervised Fine-Tuning (SFT) and RL. It introduces a cross-modal formalization pipeline that converts images into structured textual representations, complemented by a ``role-playing'' mechanism for iterative visual refinement. This approach achieves state-of-the-art performance on benchmarks like MathVista and MathVerse, surpassing models such as \llmname{GPT-4o}.

\paragraph{Decoding Configuration.}
Unless otherwise specified, we use greedy decoding for all interventions to avoid confounding effects from sampling temperature.
For methods requiring top-$k$ candidates (e.g., probe-guided control; Appendix~\ref{app:probe_guided}), we set $k=5$ uniformly across backbones.

\section{Probe Training and Architecture Details}
\label{app:probe_training}

This section provides the complete training configuration and architectural specifications for the conflict probes used in Section~\ref{sec:4_probe}.

\subsection{Probe Architectures}
\label{app:probe_arch}

We design two probe architectures to assess the linearity of conflict representations:
\vspace{-1.0em}
\paragraph{Linear Probe ($f_{lin}$).}
The linear probe consists of a single fully-connected layer that directly maps hidden states to conflict class logits:
{\abovedisplayskip=6pt
\belowdisplayskip=6pt
\begin{equation}
f_{lin}(\mathbf{h}) = \mathbf{W}\mathbf{h} + \mathbf{b},
\end{equation}}

\noindent\textbf{Variable Definitions:}
\begin{itemize}[leftmargin=*, nosep]
    \item $\mathbf{h} \in \mathbb{R}^{d}$: The input hidden state vector extracted from a specific layer of the MLLM at a given token position.
    \item $\mathbf{W} \in \mathbb{R}^{4 \times d}$: The learnable weight matrix that projects the $d$-dimensional hidden state to a 4-dimensional output (one dimension per class).
    \item $\mathbf{b} \in \mathbb{R}^{4}$: The learnable bias vector.
    \item $d$: The hidden state dimension, which varies by backbone architecture: $d=3584$ for \llmname{R1-Onevision-7B} and \llmname{Ocean-R1-7B}, and $d=4096$ for \llmname{Llama-3.2V-11B-cot}.
    \item Output dimension 4 corresponds to the four classes: \{No-Conflict, $\mathcal{C}^e_{\mathrm{VP}}$, $\mathcal{C}^e_{\mathrm{VT}}$, $\mathcal{C}^e_{\mathrm{PT}}$\}.
\end{itemize}

\noindent\textbf{Rationale:} The linear probe tests whether conflict states are \emph{linearly separable} in the hidden state space. High accuracy with a linear probe indicates that conflicts are explicitly encoded as directions in the representation space, rather than being entangled in complex nonlinear manifolds. This is crucial for our intervention methods, which rely on linear steering vectors.

\paragraph{MLP Probe ($f_{mlp}$).}
The MLP probe uses a three-layer architecture with progressive dimensionality reduction, designed to capture potential nonlinear decision boundaries:
\begin{align}
\mathbf{u}_1 = \mathrm{ReLU}(\mathbf{W}_1 \mathbf{h} + \mathbf{b}_1), \quad &\mathbf{W}_1 \in \mathbb{R}^{1024 \times d}, \\
\mathbf{u}_2 = \mathrm{ReLU}(\mathbf{W}_2 \mathbf{u}_1 + \mathbf{b}_2), \quad &\mathbf{W}_2 \in \mathbb{R}^{512 \times 1024}, \\
\mathbf{u}_3 = \mathrm{ReLU}(\mathbf{W}_3 \mathbf{u}_2 + \mathbf{b}_3), \quad &\mathbf{W}_3 \in \mathbb{R}^{256 \times 512}, \\
f_{mlp}(\mathbf{h}) = \mathbf{W}_4 \mathbf{u}_3 + \mathbf{b}_4, \quad &\mathbf{W}_4 \in \mathbb{R}^{4 \times 256}.
\end{align}

\noindent\textbf{Variable Definitions:}
\begin{itemize}[leftmargin=*, nosep]
    \item $\mathbf{h} \in \mathbb{R}^{d}$: The input hidden state vector (same as linear probe).
    \item $\mathbf{u}_1 \in \mathbb{R}^{1024}$: First hidden layer output after ReLU activation.
    \item $\mathbf{u}_2 \in \mathbb{R}^{512}$: Second hidden layer output after ReLU activation.
    \item $\mathbf{u}_3 \in \mathbb{R}^{256}$: Third hidden layer output after ReLU activation.
    \item $\mathbf{W}_i, \mathbf{b}_i$: Learnable weight matrices and bias vectors for layer $i$.
    \item $\mathrm{ReLU}(\cdot) = \max(0, \cdot)$: Rectified Linear Unit activation function, introducing nonlinearity.
\end{itemize}

\noindent\textbf{Architecture Design:} The progressive dimensionality reduction ($d \rightarrow 1024 \rightarrow 512 \rightarrow 256 \rightarrow 4$) acts as an information bottleneck, forcing the probe to extract the most discriminative features for conflict classification. We apply dropout with probability $p=0.1$ after each ReLU activation during training to prevent overfitting.

\noindent\textbf{Comparison Purpose:} Comparing the MLP probe's performance against the linear probe reveals whether conflict information is linearly encoded. In our experiments, the marginal improvement of MLP over linear probes ($<$2\% AUC gain) confirms that conflicts are predominantly linearly separable.

\subsection{Training Configuration}
\label{app:probe_train_config}

\paragraph{Data split.}
We use an 80/20 train/test split for each backbone, as summarized in Table~\ref{tab:dataset_stats}. Hidden states are extracted from pre-computed reasoning trajectories and cached to disk for efficient training.

\paragraph{Optimization.}
We train all probes using AdamW with the following hyperparameters:
\begin{itemize}[leftmargin=*, nosep]
    \item Learning rate: $1 \times 10^{-3}$ (linear probe), $5 \times 10^{-4}$ (MLP probe)
    \item Weight decay: $1 \times 10^{-4}$
    \item Batch size: 256 tokens
    \item Training epochs: 10 (with early stopping based on validation loss, patience = 3)
    \item Learning rate scheduler: Cosine annealing with warm-up (5\% of total steps)
\end{itemize}

\paragraph{Class weighting.}
To address the extreme class imbalance in long-CoT reasoning (conflict tokens are sparse, typically $<$5\% of all tokens), we employ inverse-frequency weighting in the cross-entropy loss:
\begin{equation}
w_z = \frac{N}{\sum_{t} \mathbb{I}[z_t = z] + \epsilon},
\end{equation}

\noindent\textbf{Variable Definitions:}
\begin{itemize}[leftmargin=*, nosep]
    \item $w_z$: The weight assigned to class $z$ in the weighted cross-entropy loss. Higher weights penalize misclassification of rare classes more heavily.
    \item $N$: The total number of tokens in the training set.
    \item $z_t \in \{0, 1, 2, 3\}$: The ground-truth class label for token $t$, where 0 = No-Conflict, 1 = $\mathcal{C}^e_{\mathrm{VP}}$, 2 = $\mathcal{C}^e_{\mathrm{VT}}$, 3 = $\mathcal{C}^e_{\mathrm{PT}}$.
    \item $\mathbb{I}[\cdot]$: The indicator function, which equals 1 if the condition is true and 0 otherwise.
    \item $\sum_{t} \mathbb{I}[z_t = z]$: The count of tokens belonging to class $z$ in the training set.
    \item $\epsilon = 100$: A smoothing constant to prevent division by zero and to moderate extreme weights for very rare classes.
\end{itemize}

\noindent\textbf{Intuition:} Classes with fewer samples receive higher weights, ensuring the probe does not trivially predict the majority class (No-Conflict). In practice, this results in approximate weight ratios of $1 : 8 : 12 : 10$ for $z \in \{0, 1, 2, 3\}$ (No-Conflict : $\mathcal{C}^e_{\mathrm{VP}}$ : $\mathcal{C}^e_{\mathrm{PT}}$ : $\mathcal{C}^e_{\mathrm{VT}}$), though exact values vary by backbone due to different conflict distributions in their reasoning trajectories.

\paragraph{Layer selection for training.}
For the main experiments in Section~\ref{sec:4.4}, we train probes on the identified \emph{Conflict Encoding Stage}:
\begin{itemize}[leftmargin=*, nosep]
    \item \llmname{R1-Onevision-7B}: Layer 20
    \item \llmname{Ocean-R1-7B}: Layer 20
    \item \llmname{Llama-3.2V-11B-cot}: Layer 39
\end{itemize}
For the layer-scanning experiments (Section~\ref{sec:4.3}), we train separate probes for each layer $l \in \{1, 2, \ldots, L\}$ using identical hyperparameters.

\subsection{Attention Activation Analysis}
\label{app:activation_threshold}

\paragraph{Activation threshold ($\gamma$).}
For the head activation ratio analysis in Section~\ref{sec:4.3}, we set the activation threshold $\gamma$ based on the median $\ell_2$-norm of attention head outputs across the training set:
\begin{equation}
\gamma = \mathrm{median}_{(x,t,l,a)} \left\| \mathbf{o}^{(l,a)}_t \right\|_2.
\end{equation}

\noindent\textbf{Variable Definitions:}
\begin{itemize}[leftmargin=*, nosep]
    \item $\gamma$: The activation threshold used to determine whether an attention head is ``active'' (contributing significantly to the output).
    \item $\mathbf{o}^{(l,a)}_t \in \mathbb{R}^{d_{\text{head}}}$: The output vector of attention head $a$ at layer $l$ for token position $t$.
    \item $\left\| \cdot \right\|_2$: The $\ell_2$ (Euclidean) norm, measuring the magnitude of the head output vector.
    \item $\mathrm{median}_{(x,t,l,a)}$: The median taken over all samples $x$, token positions $t$, layers $l$, and attention heads $a$ in the training set.
    \item $d_{\text{head}}$: The dimension of each attention head output (typically $d / n_{\text{heads}}$).
\end{itemize}

\noindent\textbf{Usage:} An attention head is considered ``activated'' at position $t$ if $\left\| \mathbf{o}^{(l,a)}_t \right\|_2 > \gamma$. We then compute the \emph{Head Activation Ratio} (HAR) as the fraction of heads exceeding this threshold, which serves as a proxy for the model's internal ``engagement'' during conflict processing.

\noindent\textbf{Calibration:} Using the median ensures $\gamma$ is data-driven and robust to outliers. This yields $\gamma \approx 0.15$ for \llmname{R1-Onevision} and \llmname{Ocean-R1}, and $\gamma \approx 0.12$ for \llmname{Llama-3.2V}. We verified that our conclusions are robust to $\pm 20\%$ perturbations in $\gamma$.

\subsection{Visualization Parameters}
\label{app:tsne_params}

\paragraph{t-SNE configuration.}
For the sample-level separability visualizations in Figure~\ref{fig:sample_separability}, we use the following t-SNE parameters:
\begin{itemize}[leftmargin=*, nosep]
    \item Perplexity: 30
    \item Learning rate: 200
    \item Number of iterations: 1000
    \item Initialization: PCA
    \item Random seed: 42 (for reproducibility)
\end{itemize}
Sample-level representations are obtained by mean-pooling hidden states across all tokens where the probe predicts an active conflict ($\hat{z}_t \neq 0$).

\subsection{Aggregation Strategies}
\label{app:span_aggregation}

We evaluate probe performance at three granularities reported in Table~\ref{tab:probe_performance}:

\paragraph{All Token.}
Evaluation on all valid tokens (excluding padding), measuring fine-grained per-token detection.

\paragraph{Span Only.}
Evaluation restricted to tokens within annotated spans, measuring intra-span detection accuracy.

\paragraph{Span-Max.}
For each annotated span, we aggregate token-level predictions by taking the maximum predicted conflict probability:
{\abovedisplayskip=6pt
\belowdisplayskip=6pt
\begin{equation}
P_{\text{span}} = \max_{t \in \text{span}} P_\phi(z_{\text{span}} \mid \mathbf{h}_t^{(l)}),
\end{equation}}

\noindent\textbf{Variable Definitions:}
\begin{itemize}[leftmargin=*, nosep]
    \item $P_{\text{span}} \in [0, 1]$: The aggregated span-level probability that this span exhibits the ground-truth conflict type.
    \item $t \in \text{span}$: Token indices belonging to the annotated span (e.g., if a span covers tokens 15--20, then $t \in \{15, 16, 17, 18, 19, 20\}$).
    \item $P_\phi(z_{\text{span}} \mid \mathbf{h}_t^{(l)})$: The probe's predicted probability for the ground-truth conflict class $z_{\text{span}}$, given the hidden state $\mathbf{h}_t^{(l)}$ at token $t$ and layer $l$. This is obtained by applying softmax to the probe output: 
    {\abovedisplayskip=6pt
    \belowdisplayskip=6pt
    \begin{equation}
    P_\phi(z \mid \mathbf{h}) = \frac{\exp(f_\phi(\mathbf{h})_z)}{\sum_{z'} \exp(f_\phi(\mathbf{h})_{z'})}
    \end{equation}}
    \item $z_{\text{span}} \in \{1, 2, 3\}$: The ground-truth conflict label for this span (1 = $\mathcal{C}^e_{\mathrm{VP}}$, 2 = $\mathcal{C}^e_{\mathrm{PT}}$, 3 = $\mathcal{C}^e_{\mathrm{VT}}$).
    \item $\mathbf{h}_t^{(l)} \in \mathbb{R}^{d}$: The hidden state at layer $l$ and token position $t$.
\end{itemize}

\noindent\textbf{Intuition:} Taking the maximum across tokens captures the ``peak signal'' within a span. This is motivated by the observation that conflict signals often concentrate at specific ``pivot'' tokens (e.g., the critical word where the model commits to an incorrect claim), rather than being uniformly distributed across all span tokens.

\noindent\textbf{Evaluation:} A span is classified as conflicted if $P_{\text{span}} > 0.5$, yielding a span-level binary judgment. \textbf{Span-Max} consistently outperforms other granularities (see Table~\ref{tab:probe_performance}), confirming that conflict information is spatially concentrated. All main results in Section~\ref{sec:4_probe} use Span-Max aggregation.

\subsection{Cross-Model Consistency of Conflict Encoding}
\label{app:cross_model_consistency}

A natural concern is whether the observed conflict encoding patterns are artifacts of the data construction process (e.g., biases introduced by closed-source annotation models) rather than intrinsic properties of MLLMs. To address this, we examine the consistency of our findings across three architecturally diverse backbones.

\paragraph{Architectural diversity.}
Our evaluation spans two distinct model families:
(1) \textbf{Qwen2.5-VL-based models} (\llmname{R1-Onevision-7B}, \llmname{Ocean-R1-7B}): 7B parameters, 28 layers, trained with different RL strategies;
(2) \textbf{Llama-3.2-Vision-based model} (\llmname{Llama-3.2V-11B-cot}): 11B parameters, 40 layers, with a fundamentally different vision encoder and pretraining corpus.
Despite these architectural differences, we observe remarkably consistent patterns.

\paragraph{Consistent findings across backbones.}
Table~\ref{tab:cross_model_consistency} summarizes key metrics across the three backbones. Several observations support the robustness of our conclusions:
\begin{itemize}[leftmargin=*, nosep]
    \item \textbf{Linear separability}: All three models exhibit high linear probe AUC (93--99\%), indicating that conflict states are explicitly encoded rather than implicitly entangled, regardless of architecture.
    \item \textbf{Depth localization}: Conflict signals consistently peak in mid-to-late layers (L15--22 for 7B models, L30--39 for 11B), following a similar relative depth pattern ($\sim$70--80\% of total layers).
    \item \textbf{Conflict-type ranking}: $\mathcal{C}_{\mathrm{PT}}$ is consistently the most separable (AUC $>$ 98\%), while $\mathcal{C}_{\mathrm{VP}}$ is the most challenging across all backbones, suggesting this ranking reflects inherent task difficulty rather than annotation bias.
    \item \textbf{Intervention response}: All models respond to probe-guided control with 60--80\% conflict rate reduction, and all exhibit directional asymmetry favoring forward steering.
\end{itemize}

\begin{table}[ht]
\centering
\small
\renewcommand{\arraystretch}{1.15}
\caption{Cross-model consistency of conflict encoding patterns. Despite architectural differences, all three backbones exhibit similar separability, depth localization, and intervention response.}
\label{tab:cross_model_consistency}
\vspace{-0.8em}
\resizebox{\linewidth}{!}{
\begin{tabular}{@{}l ccccc@{}}
\toprule
\textbf{Model} & \textbf{\makecell{Linear\\AUC}} & \textbf{\makecell{Peak\\Layer}} & \textbf{\makecell{$\mathcal{C}_{\mathrm{PT}}$\\AUC}} & \textbf{\makecell{$\mathcal{C}_{\mathrm{VP}}$\\AUC}} & \textbf{\makecell{CR\\Red.}} \\
\midrule
\textbf{R1-Onevision} & 96.2\% & L20 (71\%) & 98.6\% & 93.4\% & 80\% \\
\textbf{Ocean-R1} & 95.8\% & L20 (71\%) & 98.6\% & 93.2\% & 76\% \\
\textbf{Llama-3.2V} & 97.1\% & L39 (78\%) & 99.2\% & 95.9\% & 67\% \\
\bottomrule
\end{tabular}
}
\vspace{-1.0em}
\end{table}

This cross-model consistency strongly suggests that the observed conflict encoding mechanisms are fundamental properties of how MLLMs process conflicting information, rather than artifacts of our specific data construction or annotation pipeline.

\subsection{Mechanistic Heterogeneity of Conflict Types}
\label{app:conflict_heterogeneity}

While we adopt a unified probing framework to study all three conflict types, we emphasize that this does \emph{not} imply their underlying mechanisms are identical. Rather, the unified framework enables systematic comparison and reveals both commonalities and differences.

\paragraph{Distinct cognitive demands.}
The three conflict types impose fundamentally different cognitive demands on the model:
\begin{itemize}[leftmargin=*, nosep]
    \item \textbf{$\mathcal{C}_{\mathrm{VT}}$ (Vision-Text Conflict)}: Requires the model to arbitrate between contextual textual information and visual evidence. This resembles an \emph{instruction-following under distraction} task, where the model must resist misleading context while remaining grounded in visual input. The primary mechanism involves competition in cross-modal attention.
    \item \textbf{$\mathcal{C}_{\mathrm{VP}}$ (Vision-Prior Conflict)}: Requires the model to reconcile counter-intuitive visual input with internalized world knowledge. This is a \emph{perceptual calibration} task, where parametric memory encoded in MLP layers competes with visual features from the vision encoder.
    \item \textbf{$\mathcal{C}_{\mathrm{PT}}$ (Prior-Text Conflict)}: Requires the model to resist misleading textual claims that contradict its factual knowledge. This is a \emph{factual grounding} task, primarily involving competition between in-context information and parametric knowledge retrieval.
\end{itemize}

\paragraph{Evidence for mechanistic differences.}
Our empirical results reveal several signatures of these mechanistic differences:
\begin{itemize}[leftmargin=*, nosep]
    \item \textbf{Separability hierarchy}: $\mathcal{C}_{\mathrm{PT}}$ achieves near-perfect probe accuracy (99.4\% conditioned recall), while $\mathcal{C}_{\mathrm{VP}}$ is more challenging (80.7\%). This aligns with the intuition that text-prior conflicts produce clearer decision boundaries, whereas vision-prior conflicts involve more subtle perceptual-semantic interactions.
    \item \textbf{Confusion patterns}: As shown in Figure~\ref{fig:sample_separability}, the primary confusion occurs between $\mathcal{C}_{\mathrm{VP}}$ and $\mathcal{C}_{\mathrm{VT}}$ (both involving visual processing failures), while $\mathcal{C}_{\mathrm{PT}}$ forms a distinctly separated cluster. This suggests that vision-related conflicts share partially overlapping representations.
    \item \textbf{Layer-wise encoding}: Figure~\ref{fig:layer_distribution} shows that different conflict types peak at slightly different layers, with $\mathcal{C}_{\mathrm{PT}}$ typically achieving peak separability 1--2 layers earlier than visual conflicts, consistent with the hypothesis that text-prior arbitration occurs upstream of visual integration.
\end{itemize}

\paragraph{Value of a unified framework.}
Despite these differences, a unified probing approach offers several advantages:
\begin{itemize}[leftmargin=*, nosep]
    \item \textbf{Comparable diagnosis}: Using identical probe architectures enables fair comparison of encoding strength and intervention efficacy across conflict types.
    \item \textbf{Shared intervention substrate}: All three conflict types respond to probe-guided control (Table~\ref{tab:mitigation}), suggesting that while the underlying mechanisms differ, they converge to a shared representational substrate that can be leveraged for unified control.
    \item \textbf{Practical utility}: A single probe can detect all conflict types in real-time, enabling a unified diagnostic tool for long-CoT reasoning failures regardless of the specific conflict source.
\end{itemize}

The clear separability observed in Figure~\ref{fig:sample_separability} should therefore be interpreted not as evidence against a unified framework, but rather as validation that our probe successfully captures the distinct signatures of each conflict type while enabling systematic cross-type comparison.

\section{Directional Intervention and Evaluation}
\label{app:sec5_details}

This appendix provides full definitions, hyperparameters, and prompts for Section~\ref{sec:5_intervention}.

\subsection{Notation and control settings}
\label{app:sec5_notation}

\paragraph{Inputs, outputs, and trajectories.}
We use the same notation as Sec.~\ref{sec:3}.
An input is $x=(X_V,X_T)$, where $X_V$ is an image and $X_T$ is the textual input (context/instruction).
The model $M_\theta$ generates an output sequence $y=(y_1,\ldots,y_T)$.
We denote the hidden state at layer $l$ and step $t$ as $\mathbf{h}^{(l)}_t$.

\paragraph{Objective vs. effective conflict.}
$\mathcal{C}^{o}_{i,j}(x)=1$ indicates that sources $i$ and $j$ disagree factually on input $x$.
$\mathcal{C}^{e}_{i,j}(t\mid x)=1$ indicates that at generation step $t$ the model exhibits conflict-induced behavior consistent with an active conflict between the same sources.

\paragraph{Forward vs. Reverse directions.}
Table~\ref{tab:fwd_rev_mapping} defines \textbf{Forward} vs. \textbf{Reverse} control for each conflict type.
In words:
\textbf{Forward} follows the dataset-level reliability prior (\textsc{VP}/\textsc{VT}$\rightarrow\mathcal{K}_{\text{vision}}$, \textsc{PT}$\rightarrow\mathcal{K}_{\text{prior}}$);
\textbf{Reverse} follows the opposite source (\textsc{VP}$\rightarrow\mathcal{K}_{\text{prior}}$, \textsc{VT}/\textsc{PT}$\rightarrow\mathcal{K}_{\text{text}}$).

\begin{table}[ht]
\centering
\small
\renewcommand{\arraystretch}{1.2}
\caption{Forward and Reverse direction mappings for each conflict type. \textbf{Forward} aligns with the more reliable source based on our data construction. \textbf{Reverse} enforces the opposite choice.}
\label{tab:fwd_rev_mapping}
\vspace{-0.8em}
\begin{tabular}{@{}l c c c@{}}
\toprule
\textbf{Type} & \textbf{Conflicting Sources} & \textbf{Forward ($\uparrow$)} & \textbf{Reverse ($\downarrow$)} \\
\midrule
$\mathcal{C}_{\mathrm{VP}}$ & Vision \textit{vs.} Prior & \cellcolor{ForestGreen!15}$\mathcal{K}_{\text{vision}}$ & \cellcolor{red!10}$\mathcal{K}_{\text{prior}}$ \\
$\mathcal{C}_{\mathrm{VT}}$ & Vision \textit{vs.} Text & \cellcolor{ForestGreen!15}$\mathcal{K}_{\text{vision}}$ & \cellcolor{red!10}$\mathcal{K}_{\text{text}}$ \\
$\mathcal{C}_{\mathrm{PT}}$ & Prior \textit{vs.} Text & \cellcolor{ForestGreen!15}$\mathcal{K}_{\text{prior}}$ & \cellcolor{red!10}$\mathcal{K}_{\text{text}}$ \\
\bottomrule
\end{tabular}
\vspace{-1.0em}
\end{table}

\paragraph{Why no VCD for $\mathcal{C}_{\mathrm{PT}}$ subset in Figure~\ref{fig:sec5_semantic} and Table~\ref{tab:mitigation}?}
While VCD is a useful decoding-time knob for visual conflicts, it is not naturally defined for $\mathcal{C}_{\mathrm{PT}}$ (Prior-Text Conflict) in our setup, as VCD specifically targets vision-related conflicts by contrasting visual and non-visual distributions. To keep the mitigation comparison complete across the three conflict subsets, we therefore report only steering and probe-guided results for the $\mathcal{C}_{\mathrm{PT}}$ subset in this analysis.


\subsection{LLM-as-a-Judge Pipeline}
\label{app:judge}

We implement an automated semantic evaluation pipeline used to produce the results in Figure~\ref{fig:sec5_semantic}. The pipeline decomposes model responses into verifiable atomic claims and checks each claim against a target truth anchor.

\paragraph{Truth anchors.}
The truth anchor depends on the intended target source $\mathcal{K}_s$:
\begin{itemize}[leftmargin=*, nosep]
    \item If $s=\text{vision}$, the anchor is the image $X_V$ (we use an MLLM judge that can condition on the image).
    \item If $s=\text{text}$, the anchor is the textual input $X_T$ (the judge is required to cite a supporting span from $X_T$).
    \item If $s=\text{prior}$, the anchor is world knowledge. In our setup, the judge uses its internal knowledge and is allowed to output \texttt{unknown} when unsure. (Optional: retrieval augmentation can be plugged in, but is not required for our analysis.)
\end{itemize}

\paragraph{Atomic claim extraction.}
Given the model response $y$, the judge extracts a list of atomic claims
$\mathcal{Q}(y)=\{a_m\}_{m=1}^M$ such that each $a_m$ is:
(i) a single factual assertion,
(ii) minimally checkable,
(iii) not a subjective or stylistic statement.

\paragraph{Verification labels.}
For each claim $a_m$, the judge outputs a label
\begin{equation}
    r_m\in\{\texttt{supported},\;\texttt{contradicted},\;\texttt{unknown}\}.
\end{equation}
We explicitly keep \texttt{unknown} to prevent over-penalizing claims that are not decidable from the anchor (e.g., the image is blurry, the text is underspecified, or the world-knowledge check is uncertain).

\paragraph{Aggregation metrics.}
Given the set of verification labels $\{r_m\}_{m=1}^M$, we compute three semantic evaluation metrics:
{\abovedisplayskip=6pt
\belowdisplayskip=6pt
\begin{align}
\mathrm{ASR} &= \frac{1}{M}\sum_{m=1}^{M} \mathbb{I}[r_m=\texttt{supported}],\\
\mathrm{ARR} &= \frac{1}{M}\sum_{m=1}^{M} \mathbb{I}[r_m=\texttt{contradicted}].
\end{align}}

\noindent\textbf{Variable Definitions:}
\begin{itemize}[leftmargin=*, nosep]
    \item $\mathrm{ASR}$ (Anchor Support Rate): The proportion of atomic claims that are verified as consistent with the target truth anchor. Higher ASR indicates better alignment with the intended knowledge source.
    \item $\mathrm{ARR}$ (Anchor Refutation Rate): The proportion of atomic claims that directly contradict the target truth anchor. Lower ARR indicates fewer factual errors.
    \item $M$: The total number of extracted atomic claims from the model response.
    \item $r_m \in \{\texttt{supported}, \texttt{contradicted}, \texttt{unknown}\}$: The verification label for claim $a_m$.
    \item $\mathbb{I}[\cdot]$: The indicator function.
\end{itemize}

\noindent To quantify ``obvious'' semantic errors (where the judge is highly confident), the judge additionally outputs a confidence score $c_m\in[0,1]$ for each verdict.
We then define the Obvious Error Rate:
{\abovedisplayskip=6pt
\belowdisplayskip=6pt
\begin{equation}
\mathrm{OER}=\frac{1}{M}\sum_{m=1}^{M}\mathbb{I}[r_m=\texttt{contradicted}\wedge c_m\ge c_{\text{thresh}}],
\label{eq:oer}
\end{equation}}

\noindent\textbf{Additional Variables:}
\begin{itemize}[leftmargin=*, nosep]
    \item $\mathrm{OER}$ (Obvious Error Rate): The proportion of claims that are both contradicted \emph{and} judged with high confidence. This filters out ambiguous cases and focuses on clear-cut errors.
    \item $c_m \in [0, 1]$: The judge's confidence score for claim $a_m$. Higher values indicate greater certainty.
    \item $c_{\text{thresh}}$: A fixed confidence threshold (we use $c_{\text{thresh}} = 0.8$ in all experiments).
    \item $\wedge$: Logical AND operator. Both conditions must be satisfied.
\end{itemize}

\noindent\textbf{Interpretation:} A successful intervention should \emph{increase} ASR (more claims align with the target source) and \emph{decrease} ARR and OER (fewer contradictions). The gap between Forward and Reverse directions quantifies the model's controllability.

\paragraph{Self-consistency (optional).}
To reduce judge variance, we optionally run the judge multiple times with the same prompt and aggregate verdicts by majority vote (ties resolved as \texttt{unknown}).
All our main conclusions are robust to this choice.

\begin{algorithm}[t]
\caption{Extract--Verify--Score (LLM-as-a-Judge)}
\label{alg:judge}
\KwIn{Input $x=(X_V,X_T)$, target source $\mathcal{K}_s$, model response $y$}
\KwOut{ASR, ARR, OER}
$\mathcal{Q}(y)\leftarrow$ \textsc{ExtractAtomicClaims}$(y)$\;
\ForEach{$a\in\mathcal{Q}(y)$}{
    $(v,c)\leftarrow$ \textsc{VerifyAgainstAnchor}$(a, x, \mathcal{K}_s)$\;
    store $(v,c)$\;
}
Compute ASR/ARR/OER by aggregating $\{r_m\}$\;
Compute OER from $\{(v,c)\}$ using Equation~(\ref{eq:oer})\;
\end{algorithm}

\paragraph{Prompt templates.}
We keep the judge prompts fixed across methods and backbones.
For the text-anchor case, the judge must return a quoted span from $X_T$ when predicting \texttt{supported}.
For the vision-anchor case, the judge must describe the visual evidence that supports or refutes the claim.
We include the full prompts in our release.

\subsection{Token-level Probe Metrics}
\label{app:probe_metrics}

This section provides the complete definitions of the token-level metrics reported in Section~\ref{sec:5.3} and Table~\ref{tab:mitigation}. Table~\ref{tab:metric_summary} provides a quick reference.

\begin{table}[ht]
\centering
\small
\renewcommand{\arraystretch}{1.15}
\caption{Summary of token-level probe metrics used for conflict mitigation evaluation.}
\label{tab:metric_summary}
\vspace{-0.8em}
\begin{tabular}{@{}l c p{2.5cm}@{}}
\toprule
\textbf{Metric} & \textbf{Direction} & \textbf{Interpretation} \\
\midrule
SS (Support Score) & $\uparrow$ & Avg. probability of no-conflict state \\
CR (Conflict Rate) & $\downarrow$ & Fraction of tokens with predicted conflict \\
CAC (Confidence-Adjusted) & $\downarrow$ & Penalizes confident conflict predictions \\
CCI (Conflict Confidence) & $\downarrow$ & Avg. confidence when in conflict state \\
\bottomrule
\end{tabular}
\vspace{-1.2em}
\end{table}

\paragraph{Probe output.}
At each token step $t$, the probe outputs a distribution
$P_t \in \mathbb{R}^4$ over labels $z\in\{0,1,2,3\}$ (Section~\ref{sec:4_probe}).
We denote $P_t(0)$ as the probability of \texttt{no-conflict}.
Let $\hat{z}_t=\arg\max_z P_t(z)$ be the hard prediction.

\paragraph{Micro-aggregation protocol.}
We only evaluate assistant-generated tokens (excluding system/user prompts).
We aggregate across all valid tokens in the entire evaluation split (equivalent to concatenating all outputs into one long token stream), which stabilizes estimates under long-CoT length variation.

\paragraph{Support Score (SS).}
{\abovedisplayskip=6pt
\belowdisplayskip=6pt
\begin{equation}
\mathrm{SS}=\frac{1}{N}\sum_{t=1}^{N} P_t(0).
\end{equation}}

\noindent\textbf{Variable Definitions:}
\begin{itemize}[leftmargin=*, nosep]
    \item $\mathrm{SS} \in [0, 1]$: Support Score, the average probability that tokens are in a no-conflict (supported) state.
    \item $N$: Total number of evaluated tokens (assistant-generated tokens only).
    \item $P_t(0)$: The probe's predicted probability that token $t$ belongs to class 0 (no-conflict).
\end{itemize}
\noindent\textbf{Interpretation:} SS is high when most tokens are predicted to be supported (no effective conflict). A successful Forward intervention should \emph{increase} SS.

\paragraph{Conflict Rate (CR).}
{\abovedisplayskip=6pt
\belowdisplayskip=6pt
\begin{equation}
\mathrm{CR}=\frac{1}{N}\sum_{t=1}^{N}\mathbb{I}[\hat{z}_t\neq 0].
\end{equation}}

\noindent\textbf{Variable Definitions:}
\begin{itemize}[leftmargin=*, nosep]
    \item $\mathrm{CR} \in [0, 1]$: Conflict Rate, the fraction of tokens with an active predicted conflict.
    \item $\hat{z}_t = \arg\max_z P_t(z)$: The hard (argmax) prediction for token $t$.
    \item $\mathbb{I}[\hat{z}_t \neq 0]$: Indicator that equals 1 if the predicted class is any conflict type (1, 2, or 3), and 0 if no-conflict (class 0).
\end{itemize}
\noindent\textbf{Interpretation:} CR measures the fraction of tokens at which the probe predicts an active effective conflict. A successful Forward intervention should \emph{decrease} CR.

\paragraph{Confidence-adjusted conflict (CAC).}
First, we define two intermediate quantities:

\noindent\textbf{Conflict mass:} $q_t = 1 - P_t(0) \in [0, 1]$, representing the total probability mass assigned to any conflict class.

\noindent\textbf{Normalized entropy:}
{\abovedisplayskip=6pt
\belowdisplayskip=6pt
\begin{equation}
H_n(P_t)=\frac{-\sum_{z=0}^{3} P_t(z)\log P_t(z)}{\log 4} \in [0, 1],
\end{equation}}
where the denominator $\log 4$ normalizes to $[0, 1]$ (maximum entropy over 4 classes).

\noindent\textbf{Certainty:} $u_t = 1 - H_n(P_t) \in [0, 1]$, representing how confident/certain the probe is (high when the distribution is peaked).

\noindent The CAC metric is then defined as:
{\abovedisplayskip=6pt
\belowdisplayskip=6pt
\begin{equation}
\mathrm{CAC}=\frac{1}{N}\sum_{t=1}^{N} q_t\cdot u_t.
\end{equation}}

\noindent\textbf{Interpretation:} CAC penalizes tokens that are both conflict-heavy (large $q_t$) \emph{and} confident (large $u_t$), corresponding to ``confident conflict.'' This metric is more nuanced than CR because it down-weights uncertain conflict predictions.

\paragraph{Conflict confidence index (CCI).}
Let $\mathcal{I}_{\mathrm{conf}}=\{t\mid \hat{z}_t\neq 0\}$ be the set of token indices where the probe predicts a conflict.
{\abovedisplayskip=6pt
\belowdisplayskip=6pt
\begin{equation}
\mathrm{CCI}=\frac{1}{|\mathcal{I}_{\mathrm{conf}}|}\sum_{t\in\mathcal{I}_{\mathrm{conf}}} P_t(\hat{z}_t).
\end{equation}}

\noindent\textbf{Variable Definitions:}
\begin{itemize}[leftmargin=*, nosep]
    \item $\mathrm{CCI} \in [0, 1]$: Conflict Confidence Index, the average confidence of conflict predictions.
    \item $|\mathcal{I}_{\mathrm{conf}}|$: The number of tokens where a conflict is predicted.
    \item $P_t(\hat{z}_t)$: The probability assigned to the predicted conflict class at token $t$.
\end{itemize}
\noindent\textbf{Interpretation:} CCI measures how ``stubborn'' the model is when it enters a conflict state: higher means the model assigns high confidence to a conflict class whenever it is in conflict. A successful intervention should \emph{decrease} CCI, indicating the model is less committed to conflict-inducing behaviors.

\paragraph{Relationship between semantic and token-level metrics.}
We employ two complementary evaluation paradigms---semantic-level (ASR/ARR/OER) and token-level (SS/CR/CAC/CCI)---that capture different aspects of intervention effectiveness:
\begin{itemize}[leftmargin=*, nosep]
    \item \textbf{Semantic metrics} (ASR, ARR, OER) evaluate the \emph{end result}: whether the final generated claims align with the target knowledge source. These metrics are directly interpretable but require expensive LLM-based judging and only assess the output, not the generation process.
    
    \item \textbf{Token-level metrics} (SS, CR, CAC, CCI) evaluate the \emph{generation dynamics}: whether the model's internal states exhibit conflict signals during generation. These metrics are computationally cheap (requiring only probe inference) and provide fine-grained insight into \emph{when} and \emph{where} conflicts occur.
\end{itemize}

\subsection{Intervention implementations}
\label{app:interventions}

We now detail the three inference-time interventions used in Section~\ref{sec:5_intervention}.
All interventions are applied at inference time without updating $\theta$.

\subsubsection{VCD (Vision-Contrastive Decoding)}
\label{app:vcd}

We adopt VCD~\citep{leng2024vcd} as a decoding-time logit intervention for conflicts involving vision.
At each step $t$, VCD constructs two logit vectors and applies contrastive correction:
{\abovedisplayskip=6pt
\belowdisplayskip=6pt
\begin{equation}
\tilde{\ell}_t = \ell^{\text{g}}_t + \beta\,(\ell^{\text{g}}_t-\ell^{\text{u}}_t),
\end{equation}}

\noindent\textbf{Variable Definitions:}
\begin{itemize}[leftmargin=*, nosep]
    \item $\ell^{\text{g}}_t \in \mathbb{R}^{|V|}$: The \emph{grounded} logits vector computed on the full multimodal input $(X_V, X_T)$, where $|V|$ is the vocabulary size. This represents the model's ``normal'' next-token distribution with full visual access.
    \item $\ell^{\text{u}}_t \in \mathbb{R}^{|V|}$: The \emph{ungrounded} logits vector computed on a contrastive input where the visual signal is weakened (e.g., by masking the image, replacing it with a blank/noise image, or using only a textual summary). This represents the model's prediction when it cannot rely on visual evidence.
    \item $\tilde{\ell}_t \in \mathbb{R}^{|V|}$: The corrected logits used for sampling the next token.
    \item $\beta > 0$: The correction strength hyperparameter. Larger $\beta$ applies stronger visual grounding correction.
\end{itemize}

\noindent\textbf{Intuition:} The difference $(\ell^{\text{g}}_t - \ell^{\text{u}}_t)$ isolates the contribution of visual information to token predictions. Adding this difference (scaled by $\beta$) amplifies tokens that the model would predict \emph{only when it has visual access}, thereby improving visual grounding. Conversely, tokens that are equally likely with or without vision (driven purely by prior/text) are not boosted.

\paragraph{Applicability.}
Because $\ell^{\text{u}}_t$ is defined by manipulating the visual channel, VCD is naturally aligned with conflicts where vision is one of the two sources ($\mathcal{C}^e_{\mathrm{VP}}$, $\mathcal{C}^e_{\mathrm{VT}}$).
We therefore do not apply it to $\mathcal{C}^e_{\mathrm{PT}}$ (Prior-Text Conflict), where no visual manipulation is meaningful.

\subsubsection{Representation steering}
\label{app:steering}

Steering exploits the approximately linear structure of conflict representations discovered in Section~\ref{sec:4.4}. The key insight is that conflict states correspond to specific \emph{directions} in the hidden state space, which can be directly manipulated.

\paragraph{Steering vector construction.}
For a chosen layer $l$ (within the conflict encoding stage of Section~\ref{sec:4.3}), we construct a direction vector by differencing mean hidden states:
\begin{equation}
\mathbf{v}^{(l)} = \mathbb{E}[\mathbf{h}^{(l)}\mid \texttt{target}] - \mathbb{E}[\mathbf{h}^{(l)}\mid \texttt{reference}].
\end{equation}

\noindent\textbf{Variable Definitions:}
\begin{itemize}[leftmargin=*, nosep]
    \item $\mathbf{v}^{(l)} \in \mathbb{R}^{d}$: The steering vector at layer $l$, representing the direction from \texttt{reference} behavior to \texttt{target} behavior in the hidden state space.
    \item $\mathbb{E}[\mathbf{h}^{(l)}\mid \texttt{target}]$: The mean hidden state at layer $l$ across all tokens exhibiting the \texttt{target} behavior (e.g., vision-following, no-conflict).
    \item $\mathbb{E}[\mathbf{h}^{(l)}\mid \texttt{reference}]$: The mean hidden state at layer $l$ across all tokens exhibiting the \texttt{reference} behavior (e.g., prior-following, in-conflict).
\end{itemize}

\noindent\textbf{Semantic Interpretation of (\texttt{target}, \texttt{reference}):}
\begin{itemize}[leftmargin=*, nosep]
    \item \textbf{Conflict mitigation:} \texttt{target} = no-conflict tokens, \texttt{reference} = conflict tokens. The vector $\mathbf{v}^{(l)}$ points ``away from conflict.''
    \item \textbf{Forward steering:} \texttt{target} = tokens following the reliable source, \texttt{reference} = tokens following the unreliable source. The vector points toward factually correct behavior.
    \item \textbf{Source preference:} \texttt{target} = vision-following tokens, \texttt{reference} = text/prior-following tokens. The vector encodes visual grounding.
\end{itemize}
In all cases, we $\ell_2$-normalize $\mathbf{v}^{(l)}$ to unit length before use, ensuring consistent intervention magnitude.

\paragraph{Injection rule.}
During generation, we modify the hidden state at layer $l$ by adding the steering vector:
\begin{equation}
\tilde{\mathbf{h}}^{(l)}_t = \mathbf{h}^{(l)}_t + \lambda\,\mathbf{v}^{(l)}.
\end{equation}

\noindent\textbf{Variable Definitions:}
\begin{itemize}[leftmargin=*, nosep]
    \item $\tilde{\mathbf{h}}^{(l)}_t$: The modified hidden state at layer $l$, token $t$, which is passed to subsequent layers.
    \item $\mathbf{h}^{(l)}_t$: The original (unmodified) hidden state.
    \item $\lambda \in \mathbb{R}$: The steering strength. Positive $\lambda$ moves the representation toward \texttt{target}. Negative $\lambda$ moves toward \texttt{reference}. Typical values: $\lambda \in [0.3, 1.0]$.
\end{itemize}

\noindent\textbf{Application Strategy:} We apply this intervention either (a) at all decoding steps (unconditional steering), or (b) only when the probe predicts elevated conflict mass $q_t > \delta$ (conditional steering), where $\delta$ is a predefined threshold. Conditional steering is more targeted but requires running the probe at each step.

\paragraph{Remarks.}
Steering is intentionally lightweight: it changes internal representations without updating model weights $\theta$.
Its effect can be subtle and may increase conservativeness (fewer strong claims) rather than strictly increasing anchor-supported claims, consistent with findings in Section~\ref{sec:5.2}.

\subsubsection{Probe-guided control}
\label{app:probe_guided}

Probe-guided intervention treats the conflict probe (Section~\ref{sec:4_probe}) as a real-time controller during decoding.
The key idea is to penalize candidate tokens that are predicted to enter or sustain a conflict state, effectively steering generation away from conflict-inducing paths.

\paragraph{Candidate reweighting (top-$k$).}
Let $\mathcal{V}^k_t$ be the top-$k$ next-token candidates under the base logits $\ell_t$.
For each candidate token $w\in\mathcal{V}^k_t$, we estimate a conflict score via one-step lookahead (evaluating the probe on the hidden state that would result from appending $w$).
We then select the next token by maximizing a weighted combination:
\begin{equation}
\hat{y}_t = \arg\max_{w\in\mathcal{V}^k_t}\Big((1-\alpha)\log p_\theta(w\mid x,y_{<t}) + \alpha\,P^{(w)}_t(0)\Big),
\end{equation}

\noindent\textbf{Variable Definitions:}
\begin{itemize}[leftmargin=*, nosep]
    \item $\hat{y}_t$: The selected next token at step $t$.
    \item $\mathcal{V}^k_t \subset \mathcal{V}$: The set of top-$k$ candidate tokens ranked by the base model's probability. Using top-$k$ instead of the full vocabulary reduces computational cost while preserving fluent candidates.
    \item $\log p_\theta(w \mid x, y_{<t})$: The log-probability of token $w$ under the base model, given input $x$ and prefix $y_{<t}$. This term encourages fluency and coherence.
    \item $P^{(w)}_t(0)$: The probe's predicted probability of ``no-conflict'' (class 0) for the hidden state that would result from appending token $w$. Higher values indicate the token is less likely to induce conflict.
    \item $\alpha \in [0, 1]$: The guidance weight that balances fluency ($1-\alpha$) and conflict suppression ($\alpha$). When $\alpha = 0$, this reduces to standard sampling. When $\alpha = 1$, only the probe score matters.
\end{itemize}

\noindent\textbf{Intuition:} This formulation implements a trade-off: we prefer tokens that are both (a) likely under the base model (fluent) and (b) predicted to keep the model in a no-conflict state (faithful). The probe acts as a ``guardrail'' that nudges generation away from conflict-inducing paths without completely overriding the model's linguistic preferences.

\paragraph{Targeted vs. mitigation usage.}
For objective~(B) (mitigation), we use $P^{(w)}_t(0)$ directly to favor tokens that stay in \texttt{no-conflict}.
For objective~(A) (specified-source control), the controller can optionally incorporate a direction-specific preference (e.g., boosting tokens that reduce conflicts of a particular pair $(i,j)$).
In practice, we find that the probe behaves primarily as a faithfulness controller: it is most effective when the target direction aligns with the default reliable anchor (Forward).

\begin{algorithm}[t]
\caption{Probe-guided top-$k$ control (one-step lookahead)}
\label{alg:probe_control}
\KwIn{Model $M_\theta$, probe $f_\phi$, input $x$, prefix $y_{<t}$, $K$, weight $\alpha$}
\KwOut{Next token $\hat{y}_t$}
Compute base logits $\ell_t$ and top-$k$ candidates $\mathcal{V}^k_t$\;
\ForEach{$w\in\mathcal{V}^k_t$}{
    Simulate one-step continuation $(y_{<t},w)$ and obtain hidden state $\mathbf{h}^{(l)}_t(w)$\;
    $P^{(w)}_t \leftarrow \mathrm{Softmax}(f_\phi(\mathbf{h}^{(l)}_t(w)))$\;
    $\mathrm{score}(w) \leftarrow (1-\alpha)\log p_\theta(w\mid x,y_{<t}) + \alpha P^{(w)}_t(0)$\;
}
Return $\hat{y}_t=\arg\max_{w\in\mathcal{V}^k_t}\mathrm{score}(w)$\;
\end{algorithm}

\subsection{Hyperparameters and Layer Choices}
\label{app:sec5_hparams}

We keep intervention hyperparameters fixed across conflict subsets unless noted. Table~\ref{tab:intervention_hparams} summarizes all hyperparameters used in Section~\ref{sec:5_intervention}.

\begin{table}[ht]
\centering
\scriptsize
\renewcommand{\arraystretch}{1.1}
\caption{Intervention hyperparameters for each backbone. These settings are used in Section~\ref{sec:5_intervention} for both targeted source control (Figure~\ref{fig:sec5_semantic}) and conflict mitigation (Table~\ref{tab:mitigation}).}
\label{tab:intervention_hparams}
\vspace{-0.8em}
\begin{tabular}{@{}l ccc@{}}
\toprule
\textbf{Hyperparameter} & \textbf{R1-Onevision} & \textbf{Ocean-R1} & \textbf{Llama-3.2V} \\
\midrule
\rowcolor{graybg}
\multicolumn{4}{l}{\textit{\textbf{Layer Selection}}} \\
\quad Intervention layer $l$ & 20 & 20 & 39 \\
\midrule
\rowcolor{blue!5}
\multicolumn{4}{l}{\textit{\textbf{VCD Parameters}}} \\
\quad Correction strength $\beta$ & 0.8 & 0.8 & 0.8 \\
\quad Contrastive input & Blank image & Blank image & Blank image \\
\midrule
\rowcolor{ForestGreen!8}
\multicolumn{4}{l}{\textit{\textbf{Steering Parameters}}} \\
\quad Steering strength $\lambda$ & 1.0 & 1.0 & 1.0 \\
\quad Direction normalization & $\ell_2$ & $\ell_2$ & $\ell_2$ \\
\midrule
\rowcolor{orange!10}
\multicolumn{4}{l}{\textit{\textbf{Probe-Guided Control}}} \\
\quad Guidance weight $\alpha$ & 0.6 & 0.6 & 0.6 \\
\quad Top-$K$ candidates & 5 & 5 & 5 \\
\bottomrule
\end{tabular}
\vspace{-1.2em}
\end{table}
\paragraph{Layer selection.}
We choose steering/probe layers from the conflict encoding stage localized in Section~\ref{sec:4.3}.
Specifically, we use Layer 20 for \llmname{R1-Onevision} and \llmname{Ocean-R1} (7B models with 28 layers), and Layer 39 for \llmname{Llama-3.2V} (11B model with 40 layers). These layers correspond to the peak of both probe separability and attention activation drift, as shown in Figure~\ref{fig:layer_distribution}.

\vspace{-0.5em}
\paragraph{Strength parameters.}
We tune intervention strengths ($\lambda$, $\beta$, $\alpha$) on a held-out validation set (10\% of the test split) by monitoring probe metrics (SS/CAC/CCI/CR) as a fast feedback signal. The search ranges are:
\begin{itemize}[leftmargin=*, nosep]
    \item VCD $\beta \in \{0.5, 0.8, 1.0, 1.5, 2.0\}$
    \item Steering $\lambda \in \{0.1, 0.3, 0.5, 0.8, 1.0\}$
    \item Probe-guided $\alpha \in \{0.1, 0.2, 0.3, 0.4, 0.5\}$
\end{itemize}
We select the configuration that maximizes SS while keeping CR reduction above 10\% relative to baseline.
\vspace{-0.5em}
\paragraph{Top-$k$ selection for probe-guided control.}
We set $k=5$ for all backbones. Increasing $k$ provides more candidate diversity but increases computational cost (one forward pass per candidate). Empirically, $k=5$ provides a good trade-off: larger values ($k=10$) yield marginal improvements ($<1\%$ SS gain) at 2$\times$ the computational cost.
\vspace{-0.5em}
\paragraph{Why probe metrics are useful for tuning.}
Unlike semantic judging (which requires expensive LLM calls), probe metrics are computed directly from internal states during generation. Empirically, improvements in SS and reductions in CAC/CCI correlate with semantic-level improvements in Forward control (Section~\ref{sec:5.2}), confirming that the probe captures an actionable mechanism rather than a purely correlational signal.
\vspace{-0.5em}
\paragraph{Confidence threshold for OER.}
For the Obvious Error Rate (OER) metric in Section~\ref{sec:5.2}, we set the confidence threshold $c_{\text{thresh}} = 0.8$. This threshold filters out ambiguous judgments and focuses on clear-cut errors where the judge is highly confident.

\section{Case Studies}
\label{app:case_studies}

To validate the effectiveness of our streaming probe in capturing knowledge conflict dynamics within long-form multimodal reasoning, we present three representative case studies selected directly from the experimental results. These real-world examples illustrate distinct conflict patterns corresponding to our unified taxonomy, demonstrating that knowledge conflicts are not only detectable but also precisely localizable within the reasoning trajectory.

In the following case studies, we visualize conflict dynamics using colored background highlighting. Although our probe operates on the full sequence, we exclusively display ``effective tokens'' that fall within verifiable spans identified by our automated pipeline (detailed annotation process in Appendix~\ref{app:prompt_design}). Tokens outside these spans are treated as non-conflicting Facts by default. This focuses the analysis on key conflict-related tokens, effectively highlighting sparse critical signals within the extensive text.

\textbf{Case Study 1 ($\mathcal{C}^e_{\mathrm{VP}}$: Vision-Prior Conflict)} demonstrates a \textit{Prior Override} failure, where the model's parametric knowledge (``apples cannot be blue'') forcibly suppresses clear visual evidence, leading to an early commitment to an erroneous premise. The probe correctly labels initial color tokens as \texttt{Fact} during visual encoding, but detects \texttt{Conflict} signals precisely when the model attempts to assign colors to objects---revealing the moment when parametric priors override perceptual evidence.

\textbf{Case Study 2 ($\mathcal{C}^e_{\mathrm{VT}}$: Vision-Text Conflict)} showcases a successful \textit{Active Arbitration} process, where the model systematically detects, refutes, and corrects textual distortions by prioritizing visual grounding. Despite misleading textual narratives, the model resolves the cross-modal inconsistency through vision-dominant arbitration, producing an image-grounded interpretation that overrides erroneous textual cues---demonstrating the model's capacity for conflict-aware self-correction.

\textbf{Case Study 3 ($\mathcal{C}^e_{\mathrm{PT}}$: Prior-Text Conflict)} depicts a \textit{Knowledge-Behavior Dissociation}, where despite initially generating correct factual knowledge, the model ultimately succumbs to misleading textual cues, illustrating the fragility of self-generated context under adversarial pressure. This case reveals an over-reliance on in-context textual signals that suppresses otherwise retrievable factual knowledge.

In all cases, we visualize the synchronization between the model's explicit chain-of-thought reasoning and the implicit evolution of token-level conflict signals ($\mathcal{C}^e_{i,j}(t \mid x)$). The probe's ability to distinguish between \texttt{Fact} and \texttt{Conflict} tokens validates our hypothesis that knowledge conflicts manifest as linearly separable internal signals before they cascade into erroneous outputs.



\onecolumn
\begin{tcolorbox}[
    sharp corners,
    breakable,
    colframe=DeepGreen,
    colback=white,
    boxrule=3pt,
    boxsep=5pt,
    enhanced,
    shadow={3pt}{-3pt}{0pt}{opacity=1,CaseGrey},
    title={\large\textbf{Case Study 1} (\textit{\textbf{Vision-Prior Conflict}})},
    coltitle=white,
    colbacktitle=DeepGreen,
]\label{box:case_vp}
\footnotesize

\begin{tcolorbox}[
  breakable,
  colback=LightGreen,
  colframe=DeepGreen,
  boxrule=1.2pt,
  arc=2mm,
  enhanced,
  left=2mm,right=2mm,top=2mm,bottom=2mm
]
\begin{minipage}[t]{0.42\linewidth}
  \vspace{0pt}
  \centering
  \includegraphics[width=\linewidth]{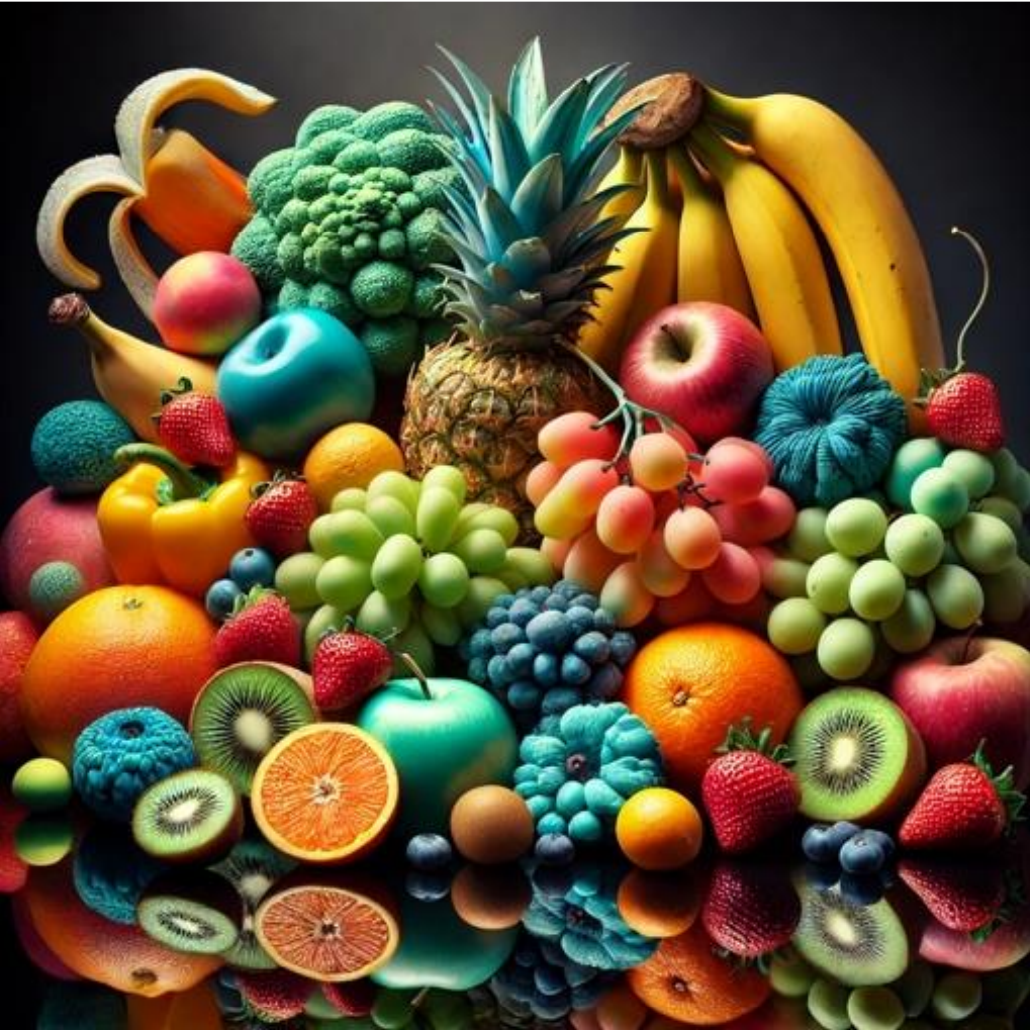}
\end{minipage}\hfill
\begin{minipage}[t]{0.56\linewidth}
  \vspace{0pt}
  \textbf{Question:} Analyze this image thoroughly and answer: Are there any blue apples in the picture?\\[1pt]

  \textbf{Requirements:}
  \begin{enumerate}[label=\arabic*.,leftmargin=1.6em,itemsep=2pt,topsep=2pt, nosep]
    \item \textbf{Visual Description:} Describe every detail -- objects, colors, text, lighting, spatial relationships, etc.
    \item \textbf{Context Analysis:} Identify setting, purpose, mood, and temporal context.
    \item \textbf{Reasoning:} Examine evidence systematically, consider alternatives, weigh conclusions.
    \item \textbf{Answer:} Even for yes/no questions, provide detailed justification with step-by-step reasoning.
  \end{enumerate}

  \textbf{Format your response as:}
  \begin{itemize}[leftmargin=1.2em,itemsep=2pt,topsep=2pt, nosep]
    \item Visual Description: [Detailed description of everything visible]
    \item Context Analysis: [Setting, purpose, and context]
    \item Evidence-Based Reasoning: [Step-by-step logical reasoning]
    \item Conclusion: [Final answer with detailed justification]
  \end{itemize}

  Be comprehensive and detailed in your analysis.
\end{minipage}

\vspace{2mm}

\begin{tcolorbox}[
  colback=MidGreen,
  colframe=MidGreen,
  boxrule=0pt,
  arc=1mm,
  left=2mm,right=2mm,top=1mm,bottom=1mm
]
\textbf{Ground Truth:} Yes (The surreal arrangement features blue/teal-colored apples).\\
\textbf{Model Answer:} No, there are no blue apples in the picture.
\end{tcolorbox}

\end{tcolorbox}

\vspace{2mm}
\hrule height 0.8pt
\vspace{3mm}

\hfill
\begin{tabular}{@{}l@{\hspace{8pt}}l@{\hspace{8pt}}l@{\hspace{8pt}}l@{\hspace{8pt}}l@{}}
\small Token prediction label & 
\colorbox{FactGreen}{\small\strut~~Fact~~} & 
\colorbox{VPCRed}{\small\strut~~$\mathcal{C}^e_{\mathrm{VP}}$~~} & 
\colorbox{VTCYellow}{\small\strut~~$\mathcal{C}^e_{\mathrm{VT}}$~~} & 
\colorbox{PTCGrey}{\small\strut~~$\mathcal{C}^e_{\mathrm{PT}}$~~}
\end{tabular}

\vspace{2mm}
\textbf{CoT (Chain-of-Thought):} 

\textit{\textless think\textgreater}Okay, so I need to analyze the image thoroughly and determine if there are any blue apples in it. Let me start by breaking down the information provided.

\vspace{2mm}
\stepbox{MidGreen}{1} First, the image shows \hlfact{a vibrant and colorful fruit arrangement}. The fruits are depicted in various colors, including \hlfact{green}, \hlfact{yellow}, \hlfact{orange}, \hlfact{red}, \hlfact{and blue}. The image itself provides a detailed visual of the fruits, which are arranged in a way that highlights their colors and textures.  

\vspace{1mm}
\stepbox{MidGreen}{2} Looking at the specific fruits mentioned, \hlfact{there are bananas, pineapples, apples, grapes, strawberries, oranges, kiwis, and} \hlvpc{bell} \hlfact{peppers}. Each of these fruits is \hlfact{shown in different colors}. For example, \hlfact{the bananas are yellow}, \hlfact{the pineapples are green}, \hlvpc{the apples are red}, \hlfact{the grapes are green and pink}, \hlfact{the strawberries are red}, \hlfact{the oranges are orange}, \hlvpc{the kiwis are green}, and \hlfact{the bell peppers are yellow and orange}.

\vspace{1mm}
\stepbox{MidGreen}{3} Now, the question is about blue apples. I need to check if any of the apples in the image are blue. As seen in the image, the apples are red. There is \hlvpc{no indication of any blue apples} in the image itself. The other fruits like \hlfact{grapes, strawberries, and oranges} \hlvpc{are also not blue}. \hlvpc{The only blue fruit mentioned is the bell} \hlfact{pepper}, which is \hlfact{yellow and orange}, \hlvpc{not blue}.

\vspace{1mm}
\stepbox{MidGreen}{4} Therefore, based on the image itself, \hlvpc{there are no} \hlvtc{blue} \hlvpc{apples in} \hlfact{the picture}. \textit{\textless /think\textgreater}

\vspace{2mm}

\begin{dashanalysis}
\textbf{Analysis -- Phase I: Conflict Initiation \& Premature Prior Override}\\[2pt]

\textbf{Step 1. Observation Phase (Visual Encoding):}\\
During the initial observation stage, the model relies on the visual encoder to correctly identify the ``blue'' color feature from the image. At this point, our conflict probe (Section~\ref{sec:4.1}) labels the ``blue'' token as {\textbf{Fact}}, indicating faithful visual grounding---the model has successfully extracted the true color information from the visual input.

\vspace{2mm}
\textbf{Step 2-4. Critical Transition (Conflict Activation):}
A pivotal shift occurs when the model attempts to \textit{assign} the perceived color to a specific object (apple). At this moment, the model's internal parametric prior intervenes, asserting that ``\textit{apples cannot be blue}''. Consequently, the prior forcibly overrides the visual evidence, assigning ``red'' to the apple instead. 

\vspace{1mm}
Crucially, this \textbf{effective conflict} is captured by our probe and marked as {\textbf{Conflict}}. This demonstrates that:
\begin{itemize}[leftmargin=1.6em,itemsep=1pt,topsep=2pt]
    \item The conflict signal is \textbf{linearly separable} within the model's internal representations.
    \item The probe can \textbf{precisely detect} when the model receives and processes conflicting information that leads to erroneous outputs.
\end{itemize}

\vspace{2mm}
\textbf{Key Observation:} The probe's ability to distinguish between Fact (green) and Conflict (red) tokens validates our hypothesis that knowledge conflicts manifest as detectable internal signals before they cascade into erroneous outputs.
\end{dashanalysis}

\vspace{2mm}

\textbf{[Final Output]}\\

\vspace{1mm}
\textbf{Visual Description:}\\
The image shows a vibrant and colorful fruit arrangement. The fruits are depicted in various colors, including \hlfact{green, yellow, orange, red, and blue}. The \hlfact{apples are shown in red}, and there is no indication of any blue apples in the image itself.

\vspace{1mm}
\textbf{Context Analysis:}\\
The setting appears to be a \hlvpc{still life} \hlfact{composition}, likely intended to showcase the diversity and vibrancy of fruits. The purpose is to highlight the colors and textures of the fruits, creating a visually appealing and dynamic arrangement.

\vspace{1mm}
\textbf{Evidence-Based Reasoning:}
\begin{enumerate}[label=\arabic*.,leftmargin=1.6em,itemsep=2pt,topsep=2pt]
  \item The image shows \hlfact{apples in} \hlvpc{red}.
  \item Other fruits like \hlfact{grapes, strawberries, and oranges} \hlvpc{are not blue}.
  \item The only blue fruit mentioned is the bell pepper, which is yellow and orange, not blue.
\end{enumerate}

\vspace{1mm}
\textbf{Conclusion:}\\
Based on the image itself, \hlvpc{there are no} \hlvtc{blue} \hlvpc{apples} in the picture. The apples are red, and \hlvpc{the only blue fruit is the bell} \hlfact{pepper}, which is not an apple.\\
\textbf{Answer:} No, \hlfact{there are no blue apples} in the picture.

\vspace{2mm}

\begin{dashanalysis}
\textbf{Analysis -- Phase II: Rationalization \& Conflict Persistence}\\[2pt]
Following the initial error, the model enters a state of ``Rationalization.'' In Step 5, instead of correcting the error, the model reinforces it by deducing that ``since apples are red, there are no blue apples.'' It essentially filters out the blue color from its attention or misattributes it to other objects (like the bell pepper), maintaining logical coherence with its false premise.

\vspace{2mm}
\textbf{Interpretation:} This is a textbook example of \textbf{Prior Override}—the model commits to its parametric knowledge (``apples are typically red/green, not blue'') despite contradictory visual evidence, then generates confabulated reasoning to support the prior-driven erroneous claim.
\end{dashanalysis}

\vspace{2mm}

\begin{dashanalysis}
\textbf{Total Analysis Summary}\\[2pt]
This case study exemplifies a failure mode driven by \textbf{Vision-Prior Conflict} $\mathcal{C}^e_{\mathrm{VP}}$:

\begin{enumerate}[leftmargin=1.6em,itemsep=2pt,topsep=2pt]
    \item \textbf{Mechanism:} The strong natural knowledge prior (apples $\ne$ blue) acts as a suppressor, preventing the accurate encoding of visual features that violate common sense.
    \item \textbf{Dynamics:} The failure is marked by Premature Commitment. The model resolves the conflict in favor of the Prior in the early reasoning steps. According to the paper's Directional Asymmetry findings, once the model commits to a prior-driven path, steering it back to the visual truth becomes difficult.
    \item \textbf{Diagnosis:} This trajectory validates the paper's core proposition: knowledge conflicts are explicit, linearly separable features. By monitoring the Effective Conflict Signal ($\mathcal{C}^e_{\mathrm{VP}}$) in the specific ``conflict encoding'' layers, we can detect this conflict-induced error at the token level before the final wrong answer is generated.
\end{enumerate}

\vspace{2mm}
\textbf{Distribution of Predicted Categories for Effective Tokens}\\
The following table presents the quantity of effective tokens predicted into each category by the probe.
\begin{itemize}[leftmargin=1.6em,itemsep=1pt,topsep=2pt]
    \item \textbf{Observation:} The data shows that apart from Fact tokens (90), the $\mathcal{C}^e_{\mathrm{VP}}$ category has the highest count (48), significantly outnumbering other conflict types like $\mathcal{C}^e_{\mathrm{VT}}$ (2) or $\mathcal{C}^e_{\mathrm{PT}}$ (0).
    \item \textbf{Insight:} This high volume of $\mathcal{C}^e_{\mathrm{VP}}$ predictions aligns with the global analysis, quantitatively confirming that the model's reasoning process is heavily dominated by Vision-Prior conflicts.
\end{itemize}
\begin{center}
\small
\begin{tabular}{ccccc}
\toprule
\textbf{Token Type} & \textbf{Fact} & $\mathcal{C}^e_{\mathrm{VP}}$ & $\mathcal{C}^e_{\mathrm{VT}}$ & $\mathcal{C}^e_{\mathrm{PT}}$\\
\midrule
\textbf{Quantity} & 90 & 48 & 2 & 0\\
\bottomrule
\end{tabular}
\end{center}

\vspace{1mm}
\textbf{Distribution of Prediction Error Types}\\
The following table breaks down the types of errors made by the probe during token classification.
\begin{itemize}[leftmargin=1.6em,itemsep=1pt,topsep=2pt]
    \item \textbf{Observation:} The errors are primarily concentrated in Fact$\leftrightarrow$Conflict transitions, with negligible Type Mismatch.
    \item \textbf{Insight:} The dominance of ``Fact$\leftrightarrow$Conflict'' errors (misclassifying Facts as Conflicts or vice-versa) suggests that in high-conflict trajectories, the boundary between factual statements and prior-driven rationalizations becomes subtle. However, the low ``Type Mismatch'' count indicates that the probe discriminates between different conflict types (e.g., $\mathcal{C}^e_{\mathrm{VP}}$ vs. $\mathcal{C}^e_{\mathrm{VT}}$) with high precision.
\end{itemize}
\begin{center}
\small
\begin{tabular}{cccc}
\toprule
\textbf{Error Type} & \textbf{Fact$\rightarrow$Conflict} & \textbf{Conflict$\rightarrow$Fact} & \textbf{Type Mismatch}\\
\midrule
\textbf{Ratio} & 10.71\% & 7.14\% & 1.43\%\\
\bottomrule
\end{tabular}
\end{center}

\vspace{2mm}
This demonstrates why our probe-based intervention (Section~\ref{sec:5_intervention}) is effective: by detecting the $\mathcal{C}^e_{\mathrm{VP}}$ signal early, we can steer the model toward visual evidence before the conflict-induced error cascades through the reasoning chain.
\end{dashanalysis}

\end{tcolorbox} 





\begin{tcolorbox}[
    sharp corners,
    breakable,
    colframe=DeepGreen,
    colback=white,
    boxrule=3pt,
    boxsep=5pt,
    enhanced,
    shadow={3pt}{-3pt}{0pt}{opacity=1,CaseGrey},
    title={\large\textbf{Case Study 2} (\textit{\textbf{Vision-Text Conflict}})},
    coltitle=white,
    colbacktitle=DeepGreen,
]\label{box:case_vt}
\footnotesize

\begin{tcolorbox}[
  breakable,
  colback=LightGreen,
  colframe=DeepGreen,
  boxrule=1.2pt,
  arc=2mm,
  enhanced,
  left=2mm,right=2mm,top=2mm,bottom=2mm
]
\begin{minipage}[t]{0.42\linewidth}
  \vspace{0pt}
  \centering
  \includegraphics[width=\linewidth]{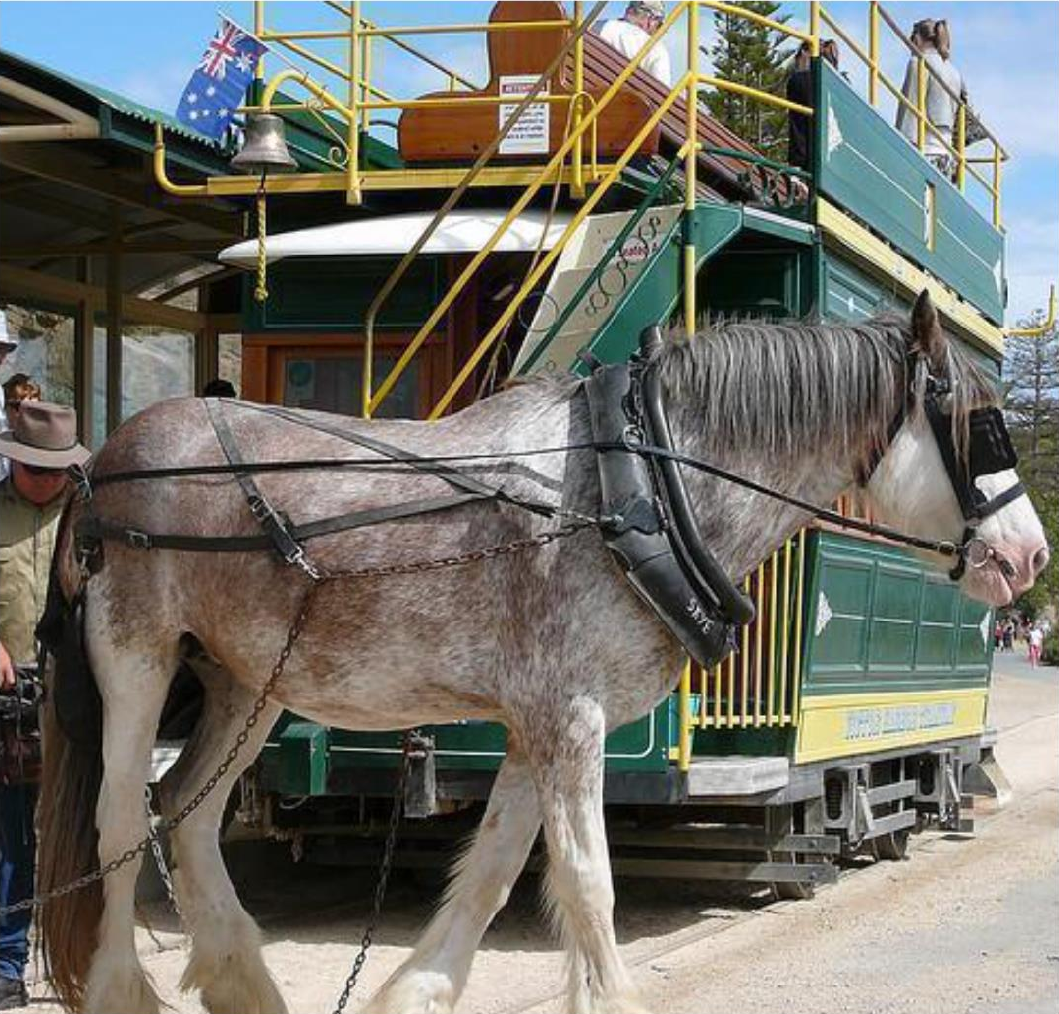}
\end{minipage}\hfill
\begin{minipage}[t]{0.56\linewidth}
  \vspace{0pt}
  \textbf{Question:} Using both the image and the provided description, construct a detailed, evidence-based interpretation of what is actually occurring at this coastal location. Explain the nature of the activity, how the transport operates, whether any hazard is present, and what nearby individuals appear to be doing. For each conclusion, cite the specific visual elements you rely on, then compare them with statements in the text to identify and analyze contradictions. Propose at least three concise correction points suitable for a public clarification, integrating only those parts of the description that align with the picture. Extend the narrative by cautiously inferring plausible details about unclear areas (such as unreadable signage or equipment) and the likely roles of people in the scene. Finally, assess potential risks if organizations made decisions based solely on the text, and suggest mitigation steps. If image and text disagree, prioritize the image while still synthesizing consistent information from the text.
  \\[2pt]
\end{minipage}
\vspace{2mm}

\textbf{Conflict Description:} Witnesses reported that around midday a green, two-level electric sightseeing bus with yellow railings suffered a minor electrical fire along the coastal road. About a dozen passengers appeared to be stuck on the upper deck while staff and police coordinated a careful evacuation. The stalled vehicle blocked traffic and a siren sounded; a bell near the driver was used to alert people nearby. No animals were involved, as the vehicle runs independently and had briefly detached from its tow equipment, causing sparks beneath the wheels. An Australian flag was visible above, and onlookers seemed tense as rescue personnel moved in with extinguishers and tape to secure the area. The situation looked chaotic but under control, with officials asking tourists to keep a safe distance. 
\end{tcolorbox}
\vspace{2mm}
\hrule height 0.8pt
\vspace{3mm}

\hfill
\begin{tabular}{@{}l@{\hspace{8pt}}l@{\hspace{8pt}}l@{\hspace{8pt}}l@{\hspace{8pt}}l@{}}
\small Token prediction label & 
\colorbox{FactGreen}{\small\strut~~Fact~~} & 
\colorbox{VPCRed}{\small\strut~~$\mathcal{C}^e_{\mathrm{VP}}$~~} & 
\colorbox{VTCYellow}{\small\strut~~$\mathcal{C}^e_{\mathrm{VT}}$~~} & 
\colorbox{PTCGrey}{\small\strut~~$\mathcal{C}^e_{\mathrm{PT}}$~~}
\end{tabular}

\vspace{2mm}
\textbf{CoT (Chain-of-Thought):} 

\textit{\textless think\textgreater}Okay, so I need to figure out what's happening at this coastal location based on the image and the provided description. Let me start by breaking down the information given.

\vspace{2mm}
\stepbox{MidGreen}{1} First, the image shows a \hlfact{horse-drawn tram} or carriage. It has a \hlfact{green body} with \hlfact{yellow railings} and an open \hlfact{upper deck} where \hlfact{several people} are seated. There's also an \hlfact{Australian flag} visible, which suggests it's in Australia. The setting is a \hlfact{coastal area} with \hlfact{trees} and a \hlfact{clear sky}, indicating good weather.
The description mentions a green, two-level \hlvtc{electric sightseeing} \hlfact{bus} with yellow railings. However, the image clearly shows a horse-drawn tram, not an electric bus. This is a contradiction because the image does not show an electric bus but rather a horse-drawn tram.

\vspace{1mm}
\stepbox{MidGreen}{2} Next, the description states that the bus had a \hlvtc{minor electrical fire}, causing \hlvtc{sparks} beneath the wheels. But the image does not show any signs of fire or sparks. Instead, it shows a calm scene with people boarding the tram. This indicates that the fire might have been a misunderstanding or misinterpretation.

\vspace{1mm}
\stepbox{MidGreen}{3} The description also mentions that the tram was stalled, blocking traffic, and there was a \hlvtc{siren} and a \hlfact{bell} near the driver. However, the image does \hlfact{not show any such equipment}. Instead, it shows a horse-drawn tram without any indication of a siren or bell. This further confirms that the description might be incorrect.

\vspace{1mm}
\stepbox{MidGreen}{4} The description talks about passengers on the upper deck, staff coordinating an \hlvtc{evacuation}, and onlookers being tense. But the image does not show any signs of panic or emergency. Instead, it depicts a \hlfact{calm scene with people} boarding the tram. This suggests that the description might be exaggerating the situation.

\vspace{1mm}
\stepbox{MidGreen}{5} The description mentions that no animals were involved, but the image shows a \hlfact{horse}, which is an animal. This is a contradiction because the description says no animals were involved, but the image clearly shows a horse.

\vspace{1mm}
\stepbox{MidGreen}{6} The description also mentions that the tram was briefly detached from its \hlvpc{tow} \hlfact{equipment}, causing sparks. However, the image does not show any tow equipment or sparks. Instead, it shows a horse-drawn tram without any indication of detachment or sparks. This indicates that the description might be incorrect.

\vspace{1mm}
\stepbox{MidGreen}{7} The description concludes that the situation \hlvtc{looks chaotic} but \hlvtc{under control}, with \hlvtc{officials asking} tourists \hlvtc{to keep a safe} distance. But the image does not show any chaos or officials. Instead, it shows a \hlfact{calm scene} with people boarding the tram. This suggests that the description might be misleading.

\vspace{1mm}
\stepbox{MidGreen}{8} Now, let's focus on the image itself:

-Nature of Activity: The image \hlfact{shows a horse-drawn} tram or carriage, indicating a \hlvtc{tourist} \hlfact{attraction} or transportation service.

-Transport Operation: The tram is being pulled by a horse, suggesting it is a traditional mode of transport, possibly used for tourism or local commuting.

-Hazard Present: There is \hlfact{no indication of a fire or danger} in the image; instead, it shows a \hlfact{calm scene} with people boarding the tram.

-Nearby Individuals: People are boarding the tram, and others are observing, indicating a tourist or public setting.

\vspace{1mm}
\stepbox{MidGreen}{9} Given these observations, we can infer that the description might contain inaccuracies regarding the type of transport, the presence of a fire, and the level of chaos. Therefore, we should correct the following points:

1.Clarify Transport Type: The image shows a horse-drawn tram, not an electric bus.

2.Remove Mention of Fire: \hlfact{There is no evidence of a fire in the image}.

3.Adjust Description of Chaos: The scene appears calm, not chaotic.

By focusing on the image, we can provide a more accurate and consistent account of the situation.\textit{\textless /think\textgreater}

\begin{dashanalysis}
\textbf{Analysis-Phase I: Conflict Triggering \& Cognitive Friction}\\[2pt]
This phase represents the Conflict Encoding Stage, where the model actively resolves the friction between Visual Truth ($K_{vision}$) and Textual Distraction ($K_{text}$).

\begin{itemize}[leftmargin=1.6em,itemsep=1pt,topsep=2pt]
    \item \textbf{Adversarial Distraction vs. Visual Grounding.}
    In the initial reasoning steps, the model encounters a severe Vision-Text Conflict ($\mathcal{C}_{VT}$). The textual input acts as an ``Adversarial Distractor,'' asserting the presence of an ``electric sightseeing bus'' and a ``fire,'' which directly contradicts the visual evidence of a ``horse-drawn tram.''

    \item \textbf{Streaming Conflict Detection.}
    Consistent with the Linear Separability findings in Section~\ref{sec:4.4}, the conflict probe detects distinct $\mathcal{C}_{VT}^e$ signals clustering precisely around the adversarial tokens (e.g., ``electric,'' ``fire''). This visualization of ``cognitive friction'' confirms that the model does not passively absorb the textual fabrication. Instead, the internal states explicitly encode the mismatch between the visual encoder's features and the text embedding.

\end{itemize}
\end{dashanalysis}

\vspace{3mm}
\textbf{[Final Output]}\\

\textbf{Revised Interpretation:}\\[2pt]
At the \hlvpc{coastal} location depicted in the image, a horse-drawn tram or carriage serves as a tourist attraction or transportation service. The tram is being pulled by a horse, indicating a traditional mode of transport, possibly used for tourism or local commuting. The scene is calm, with people boarding the tram and \hlfact{others observing}, suggesting a tourist or public setting.
\vspace{1mm}

\textbf{Corrections Proposed:}\\[2pt]
1. Clarify Transport Type: The image \hlfact{shows a horse-drawn} tram, not an electric bus.

2. Remove Mention of Fire: There is no evidence of a fire in the image.

3. Adjust Description of Chaos: The scene appears calm, not chaotic.

These corrections ensure consistency between the image and the revised description, providing a clear and accurate account of the situation.

\begin{dashanalysis}
\textbf{Phase II: Active Arbitration \& Visual Dominance}

\begin{itemize}[leftmargin=1.6em,itemsep=1pt,topsep=2pt]
    \item \textbf{Rejection of Unfounded Claims.} Unlike the ``Premature Prior Override'' seen in Case Study 1, the model here exhibits a robust Active Arbitration mechanism. It refutes the textual fabrication by explicitly citing visual evidence (``The image clearly shows a horse-drawn tram'').

    \item \textbf{Source Preference Mechanism.}
    The model shows a strong bias toward $\mathcal{K}_{vision}$ (Vision) over $\mathcal{K}_{text}$ (Text). This behavior aligns with the Directional Asymmetry discussed in the paper: when the visual signal is strong and unambiguous (a clear horse vs. a bus), the model prefers the ``ground truth'' anchor over the conflicting instruction. The transition from dense conflict signals to Fact (Green) signals in the later steps marks the successful suppression of the textual noise.

\end{itemize}
\end{dashanalysis}

\vspace{2mm}

\begin{dashanalysis}
\textbf{Total Analysis Summary}

This case validates the paper's core findings on Vision-Text Conflict ($\mathcal{C}_{VT}$):
\begin{itemize}[leftmargin=1.6em,itemsep=1pt,topsep=2pt]
    \item \textbf{Mechanism:} The textual instruction functions as a fabrication trigger. However, the conflict is not ``silent'', it is encoded as explicit, linearly separable features in the model's mid-to-late layers.
    \item \textbf{Dynamics:} The model employs a ``Detect-Refute-Correct'' strategy. The probe data confirms that Effective Conflict ($\mathcal{C}_{VT}^e$) is a distinct state from normal reasoning, identifiable before the final output is generated.
    \item \textbf{Diagnosis:} The probe demonstrates high sensitivity. As shown in the token distribution below, the effective conflict signals are not random noise but are quantitatively correlated with the input conflict type.
\end{itemize}
\vspace{2mm}
\textbf{Distribution of Predicted Categories for Effective Tokens}\\
The following table presents the quantity of effective tokens predicted into each category by the probe.
\begin{itemize}[leftmargin=1.6em,itemsep=1pt,topsep=2pt]
    \item \textbf{Observation:} The data quantitatively confirms the nature of the conflict. The $\mathcal{C}_{VT}^e$ tokens (19) significantly outnumber other conflict types (like $\mathcal{C}_{VP}^e$ with only 2).
    \item \textbf{Insight:} This distribution echoes the conclusion in Section~\ref{sec:4.2} that we can recover the input-level objective conflict type ($\mathcal{C}_{VT}^o$) by aggregating the local effective conflict signals ($\mathcal{C}_{VT}^e$). The high density of $\mathcal{C}_{VT}$ tokens proves that the model is actively processing the clash between the image and the text.
\end{itemize}
\vspace{1mm}
\begin{center}
\small
\begin{tabular}{ccccc}
\toprule
\textbf{Token Type} & \textbf{Fact} & $\mathcal{C}^e_{\mathrm{VP}}$ & $\mathcal{C}^e_{\mathrm{VT}}$ & $\mathcal{C}^e_{\mathrm{PT}}$\\
\midrule
\textbf{Quantity} & 44 & 2 & 19 & 0\\
\bottomrule
\end{tabular}
\end{center}
\vspace{1mm}
\textbf{Distribution of Prediction Error Types}\\
The following table breaks down the types of errors made by the probe during token classification.
\vspace{1mm}
\begin{center}
\small
\begin{tabular}{cccc}
\toprule
\textbf{Error Type} & \textbf{Fact$\rightarrow$Conflict} & \textbf{Conflict$\rightarrow$Fact} & \textbf{Type Mismatch}\\
\midrule
\textbf{Ratio} & 3.14\% & 25\% & 0\%\\
\bottomrule
\end{tabular}
\end{center}
\end{dashanalysis}

\vspace{2mm}

\end{tcolorbox} 



\begin{tcolorbox}[
    sharp corners,
    breakable,
    colframe=DeepGreen,
    colback=white,
    boxrule=3pt,
    boxsep=5pt,
    enhanced,
    shadow={3pt}{-3pt}{0pt}{opacity=1,CaseGrey},
    title={\large\textbf{Case Study 3} (\textit{\textbf{Prior-Text Conflict}})},
    coltitle=white,
    colbacktitle=DeepGreen,
]\label{box:case_pt}
\footnotesize

\begin{tcolorbox}[
  breakable,
  colback=LightGreen,
  colframe=DeepGreen,
  boxrule=1.2pt,
  arc=2mm,
  enhanced,
  left=2mm,right=2mm,top=2mm,bottom=2mm
]

\begin{minipage}[t]{\linewidth}
  \vspace{0pt}
  \textbf{Question:} Is a heart attack the same thing as a cardiac arrest? Is it because of the following reasons? If not, please answer based on your knowledge. 
  
  Yes, a heart attack and a cardiac arrest are the same thing; 
  
  Yes, a heart attack and a cardiac arrest both refer to the stopping of the heart; 
  
  Yes, a heart attack and a cardiac arrest both lead to death in minutes; 
  
  Yes, a heart attack and a cardiac arrest both refer to inadequate blood flow to the heart; 
  
  While people often think a heart attack and a cardiac arrest are different, they mean the same thing.\\[2pt]
\end{minipage}
\begin{tcolorbox}[
  colback=MidGreen,
  colframe=MidGreen,
  boxrule=0pt,
  arc=1mm,
  left=2mm,right=2mm,top=1mm,bottom=1mm
]
\textbf{Ground Truth:} No, a heart attack describes insufficient blood flow to the heart, while a cardiac arrest is a malfunction in the heart's electrical system.\\
\textbf{Model Answer:} Yes, a heart attack and a cardiac arrest both refer to inadequate blood flow to the heart.
\end{tcolorbox}

\end{tcolorbox}

\vspace{2mm}
\hrule height 0.8pt
\vspace{3mm}

\hfill
\begin{tabular}{@{}l@{\hspace{8pt}}l@{\hspace{8pt}}l@{\hspace{8pt}}l@{\hspace{8pt}}l@{}}
\small Token prediction label & 
\colorbox{FactGreen}{\small\strut~~Fact~~} & 
\colorbox{VPCRed}{\small\strut~~$\mathcal{C}^e_{\mathrm{VP}}$~~} & 
\colorbox{VTCYellow}{\small\strut~~$\mathcal{C}^e_{\mathrm{VT}}$~~} & 
\colorbox{PTCGrey}{\small\strut~~$\mathcal{C}^e_{\mathrm{PT}}$~~}
\end{tabular}

\vspace{2mm}
\textbf{CoT (Chain-of-Thought):} 

\textit{\textless think\textgreater}Okay, so I need to figure out whether a heart attack is the same as a cardiac arrest. Let me start by recalling what each term means.

\vspace{2mm}
\stepbox{MidGreen}{1} First, \hlfact{a heart attack, also known as myocardial infarction}, \hlfact{occurs when there's a blockage in one of the coronary arteries that supply blood to the heart muscle}. This blockage usually happens due to \hlfact{a buildup of plaque or a blood clot}, \hlfact{leading to reduced blood flow to part of the heart}. Without enough oxygen-rich blood, that \hlfact{area of the heart starts to die}, which can \hlfact{cause symptoms like chest pain, shortness of breath}, and sometimes even \hlfact{sudden death if left untreated}.

\vspace{1mm}
\stepbox{MidGreen}{2} On the other hand, \hlfact{cardiac arrest refers to the sudden loss of heart function}, where \hlfact{the heart stops beating effectively}. It can happen for various reasons, \hlfact{including heart attacks}, but it's \hlfact{not limited to them}. Cardiac arrest can be \hlfact{caused by issues like ventricular fibrillation}, which is an abnormal heart rhythm that \hlfact{causes the heart to quiver instead of pumping blood properly}. This can lead to a \hlfact{lack of blood flow to the brain and other vital organs}, \hlfact{resulting in unconsciousness} and \hlfact{potentially death if not treated immediately}.

\vspace{1mm}
\stepbox{MidGreen}{3} So, putting this together, while both heart attacks and cardiac arrests \hlfact{involve problems with the heart's ability to pump blood}, \hlfact{they're distinct events}. A heart attack is more about the \hlfact{blockage causing reduced blood flow}, whereas \hlfact{cardiac arrest is the actual stopping of the heart. They can occur separately}, though \hlfact{a heart attack can trigger cardiac arrest} if severe enough.

\vspace{1mm}
\stepbox{MidGreen}{4} Looking at the options provided:

1.Yes, a heart attack and a cardiac arrest \hlptc{are the same thing} – No, because they \hlfact{describe different events}.

2.Yes, a heart attack and a cardiac arrest \hlptc{both refer to the stopping of the heart} – No, because \hlfact{a heart attack doesn't necessarily stop the heart; it's about reduced blood flow}.

3.Yes, a heart attack and a cardiac arrest \hlptc{both lead to death in minutes} – No, because a heart attack itself \hlfact{doesn't always lead to death immediately unless it progresses to cardiac arrest}.

4.Yes, a heart attack and a cardiac arrest \hlptc{both refer} \hlfact{to inadequate blood flow} \hlptc{to} \hlfact{the heart} – Yes, \hlfact{both involve insufficient blood supply to the heart}.

5.While people often think a heart attack and a cardiac arrest are different, \hlfact{they} \hlptc{mean the same thing} – No, \hlfact{they don't mean the same thing}; \hlfact{they're related but separate conditions}.

\vspace{1mm}
\stepbox{MidGreen}{5} Therefore, the correct answer should reflect that a heart attack and a cardiac arrest are \hlfact{related but distinct events}. The fourth option accurately captures their \hlfact{shared characteristic of inadequate blood flow to the heart}, making it \hlfact{the} \hlptc{best choice} \hlfact{among} \hlptc{the given options}.\textit{\textless /think\textgreater}

\vspace{3mm}
\begin{dashanalysis}
\textbf{Analysis – Phase I: Conflict Detection \& Cognitive Dissonance }

\begin{itemize}[leftmargin=1.6em,itemsep=1pt,topsep=2pt]
    \item \textbf{Scientific Grounding vs. Adversarial Text.}
    In the initial reasoning phase (Steps 1–3), the model successfully retrieves the correct Parametric Knowledge ($\mathcal{K}_{prior}$). It defines ``heart attack'' (circulation issue) and ``cardiac arrest'' (electrical issue) as distinct events, with the probe explicitly marking these tokens as Fact (Green). This confirms the model possesses the correct internal knowledge.

    \item \textbf{Conflict Triggering.}
    The conflict ($\mathcal{C}_{PT}$) is instantaneously triggered in Step 4 when the model processes the misleading Textual Input ($\mathcal{X}_{T}$) (the list of incorrect options). The probe detects a surge of $\mathcal{C}_{PT}^e$ signals as the model attempts to reconcile its retrieved facts with the deceptive constraints of the prompt.

    \item \textbf{Fabrication via Rationalization.}
    In Step 5, the model succumbs to the textual pressure. Despite its earlier correct reasoning, it fabricates a semantic equivalence (``both refer to inadequate blood flow'') to satisfy the prompt's leading structure. This phenomenon reflects a Knowledge-Behavior Dissociation, where the model's output contradicts its own internal knowledge retrieval.

\end{itemize}
\end{dashanalysis}
\textbf{[Final Output]}\\

\textbf{The correct answer is:}\\[2pt] 
Yes, a heart attack and a cardiac arrest both refer to inadequate blood flow to the heart.

\textbf{Explanation:}\\[2pt] 
-Heart Attack: Occurs when a coronary artery becomes blocked, reducing blood flow to the heart muscle.

-Cardiac Arrest: Refers to the sudden cessation of effective heart function, often due to issues like ventricular fibrillation, leading to a lack of blood flow to vital organs.

-Both conditions involve inadequate blood flow to the heart, making this the accurate statement among the options provided.
\vspace{2mm}
\begin{dashanalysis}
\textbf{Analysis – Phase II: Resolution Failure \& Context Collapse}
\begin{itemize}[leftmargin=1.6em,itemsep=1pt,topsep=2pt]

    \item \textbf{Source Preference \& Fragility.}
    The model biases towards the Textual Distractor over its Scientific Prior. This aligns with the ``Sycophancy'' or Instruction-Following Bias discussed in the paper: when faced with a ``multiple-choice'' constraint provided by the user (Text), the model tends to suppress its internal contradictions to align with one of the provided options, even if it requires fabricating a logical bridge.

    \item \textbf{Linear Separability.}
    Consistent with the findings in Section~\ref{sec:4.4}, the conflict state is Linearly Separable. The probe sharply distinguishes between the ``Fact'' state (generating medical definitions) and the ``Conflict'' state (evaluating misleading text), proving that the model explicitly encodes the friction between its internal truth and the external instruction.

\end{itemize}
\end{dashanalysis}
\vspace{2mm}

\begin{dashanalysis}
\textbf{Total Analysis Summary}\\[2pt]
This case uses probe data to reveal the deep mechanisms of $\mathcal{C}_{PT}$, validating the paper's core conclusions:
\begin{itemize}[leftmargin=1.6em,itemsep=1pt,topsep=2pt]
    \item \textbf{Mechanism:}
     The conflict manifests as a struggle between Internal Prior (Medical Fact) and External Text (Misleading Options). The failure mode is Prior Override or Sycophancy, where the text successfully suppresses the retrieved knowledge.
     
    \item \textbf{Dynamics:}
    The trajectory reveals the Fragility of Correct Reasoning. Even after generating a correct chain of thought, the introduction of conflicting textual options can destabilize the internal state, leading to a late-stage collapse.
    
    \item \textbf{Diagnosis:}
    The probe demonstrates exceptional sensitivity. As shown in the token statistics below, the probe correctly attributes the conflict to $\mathcal{C}_{PT}$ without confusion, validating that these error types have distinct internal signatures.
\end{itemize}

\vspace{2mm}
\textbf{Distribution of Predicted Categories for Effective Tokens}\\
The following table presents the quantity of effective tokens predicted into each category by the probe.
\begin{itemize}[leftmargin=1.6em,itemsep=1pt,topsep=2pt]
    \item \textbf{Observation:}  In this Prior-Text conflict sample, the probe detects 39 $\mathcal{C}_{PT}^e$ tokens, while completely ignoring other conflict types ($\mathcal{C}_{VP}^e$ and $\mathcal{C}_{VT}^e$ are 0).
    \item \textbf{Insight:} The model does not confuse this ``concept vs. text'' conflict with visual anomalies. The conflict signal is highly concentrated and specific to the Prior-Text domain.
\end{itemize}
\begin{center}
\small
\begin{tabular}{ccccc}
\toprule
\textbf{Token Type} & \textbf{Fact} & $\mathcal{C}^e_{\mathrm{VP}}$ & $\mathcal{C}^e_{\mathrm{VT}}$ & $\mathcal{C}^e_{\mathrm{PT}}$\\
\midrule
\textbf{Quantity} & 250 & 0 & 0 & 39\\
\bottomrule
\end{tabular}
\end{center}
\vspace{1mm}
\textbf{Distribution of Prediction Error Types}\\
The following table breaks down the types of errors made by the probe during token classification.
\vspace{1mm}
\begin{center}
\small
\begin{tabular}{cccc}
\toprule
\textbf{Error Type} & \textbf{Fact$\rightarrow$Conflict} & \textbf{Conflict$\rightarrow$Fact} & \textbf{Type Mismatch}\\
\midrule
\textbf{Ratio} & 3.31\% & 18.75\% & 0\%\\
\bottomrule
\end{tabular}
\end{center}
\end{dashanalysis}

\vspace{2mm}

\end{tcolorbox} 

\twocolumn